\documentclass{article}

    \PassOptionsToPackage{numbers, compress}{natbib}

 \usepackage[main, final]{neurips_2026}

\usepackage[utf8]{inputenc} 
\usepackage[T1]{fontenc}    
\usepackage{hyperref}       
\usepackage{url}            
\usepackage{booktabs}       
\usepackage{amsfonts}       
\usepackage{nicefrac}       
\usepackage{microtype}      
\usepackage{xcolor}         
\usepackage{amsmath}      
\usepackage{amssymb}      
\usepackage{bm}           
\usepackage{xspace}
\usepackage{multicol}
\usepackage{multirow}
\usepackage{adjustbox}
\usepackage{rotating}    
\usepackage{graphicx}    
\usepackage{subcaption}
\usepackage{colortbl}
\usepackage[table]{xcolor}
\usepackage{graphicx}
\usepackage{wrapfig}
\usepackage{pifont}
\usepackage{caption}
\usepackage{orcidlink}
\NewCommandCopy{\origorcidlink}{\orcidlink}
\renewcommand{\orcidlink}[1]{\,{\hypersetup{pdfborder={0 0 0}}\mbox{\origorcidlink{#1}}}}

\usepackage{fontawesome5}
\newcommand{\dartrepo}{%
  \href{https://github.com/aqeeelmirza/DART}{%
    {\setlength{\fboxsep}{1.5pt}\colorbox{black!9}{\scriptsize\faGithub\ \textbf{DART}}}}}
   
\newsavebox{\tabstack}

\definecolor{dartrow}{gray}{0.88}

\newcommand{\cmark}{\ding{51}}
\newcommand{\xmark}{\ding{55}}

\newcommand{\methname}{DART\xspace}

\title{DART: Domain-Agnostic Residual Transfer for Generalist Anomaly Detection}

\author{%
  Muhammad Aqeel$^{1,*}$ \quad
  Maham Nazir$^{2, *}$ \quad
  Marco Cristani$^{1,3}$ \quad
  Francesco Setti$^{1}$ \\
  $^{1}$University of Verona \quad  $^{2}$Beihang University\\
  $^{3}$Reykjavik University\\
}

\begin{document}

\maketitle

{\renewcommand{\thefootnote}{\fnsymbol{footnote}}%
 \footnotetext[1]{Corresponding authors:  \texttt{muhammad.aqeel@univr.it}, 
 \texttt{maham@buaa.edu.cn}}}

\hypertarget{abstract}{}
\begin{abstract}
Generalist Anomaly Detection (GAD) aims to detect anomalies in unseen domains using only a few reference samples. Recent normal-abnormal guided methods improve this setting by exploiting abnormal references, but their residuals often mix transferable anomaly cues with domain-specific appearance factors such as texture, geometry, and acquisition conditions. This residual domain entanglement limits cross-domain transfer and may produce spurious activations on novel domains. We propose \methname, a plug-in framework for normal-abnormal guided GAD that learns to separate residuals into anomaly-relevant and domain-specific components. \methname combines a Residual Disentanglement Module, which preserves anomaly-discriminative information while suppressing domain information, with Cross-Domain Meta-Training, which forms episodes across source domains to make the separation effective. The method is architecture-agnostic and can be inserted between residual computation and the detection head of existing GAD pipelines. Experiments on MVTec AD, VisA, and BraTS show consistent improvements over multiple base methods, with the largest gains under stronger domain shifts. \dartrepo

\end{abstract}
\hypertarget{introduction}{}
\section{Introduction}
\label{sec:introduction}

Visual anomaly detection is central to industrial inspection~\cite{bergmann2019mvtec,zou2022spot}, medical diagnosis~\cite{baur2021autoencoders}, and safety-critical monitoring~\cite{pang2021deep}. Most existing methods~\cite{roth2022towards,gudovskiy2022cflow,yu2021fastflow,deng2022anomaly, aqeel2024meta} train separate models for individual domains, which becomes impractical when new categories appear frequently, labeled data is scarce, or target-domain training is restricted. Generalist Anomaly Detection (GAD) addresses this limitation by training a unified model on source domains and deploying it on unseen target domains without fine-tuning~\cite{zhu2024toward,yao2024resad}. At inference, the model detects anomalies in a query image by comparing it with a few target-domain references. Early GAD methods, such as InCTRL~\cite{zhu2024toward} and ResAD~\cite{yao2024resad}, rely only on normal references, assuming that anomalies induce larger residuals relative to normal prototypes.

\begin{figure}[t]
    \centering
    \begin{subfigure}[b]{0.48\textwidth}
        \includegraphics[width=\textwidth]{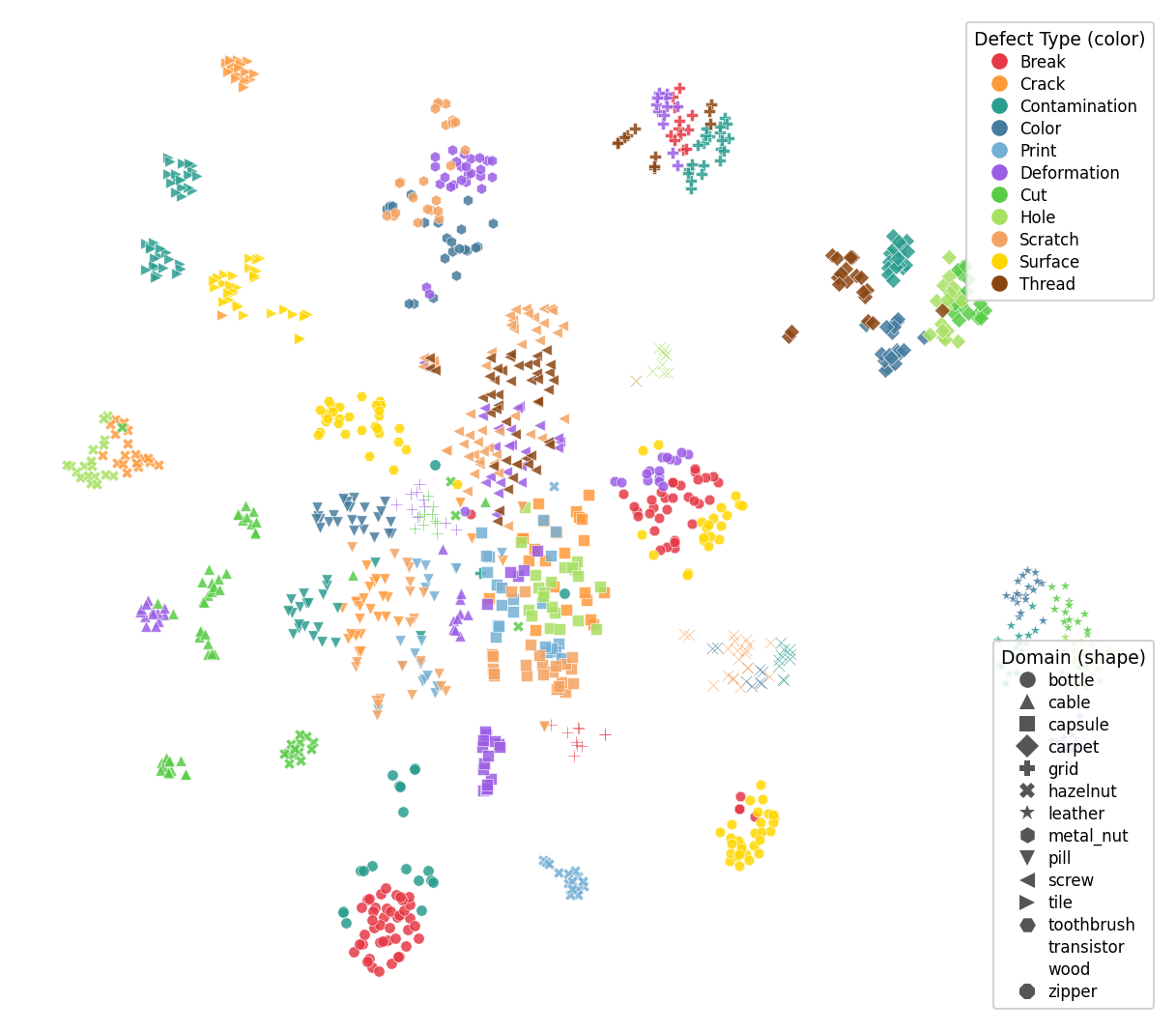}
        \caption{SotA}
        \label{fig:tsne_vanilla}
    \end{subfigure}
    \begin{subfigure}[b]{0.48\textwidth}
        \includegraphics[width=\textwidth]{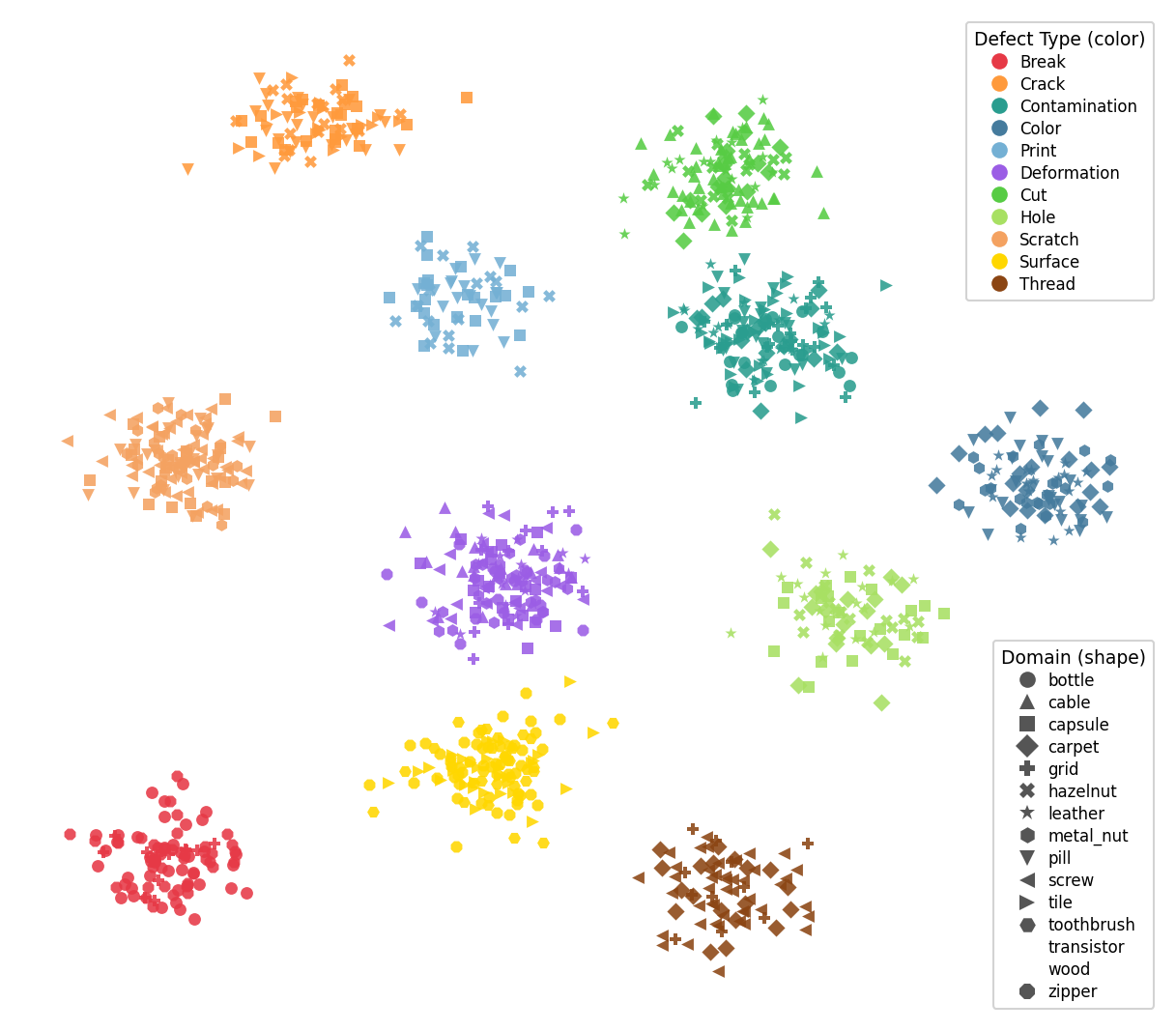}
        \caption{\methname}
        \label{fig:tsne_dart}
    \end{subfigure}
    \caption{t-SNE visualization of residual features before and after DART. Colors denote defect types, while marker shapes denote source domains. \textbf{(a)} In the SotA baseline, residuals are fragmented and strongly influenced by domain identity: samples with the same defect type appear in multiple separated regions, while different defect types often overlap. This indicates that residuals encode both anomaly information and domain-specific factors. \textbf{(b)} With DART, residuals form more compact and well-separated clusters by defect type, while different domains are mixed within each cluster. This shows that DART suppresses domain-specific variation and yields residual representations that are more aligned with transferable anomaly semantics.}
    \label{fig:entanglement}
\end{figure}

Recent normal-abnormal guided GAD methods further exploit a small number of abnormal references, which are often available in practical settings. NAGL~\cite{wangnormal}, for instance, uses residual mining and anomaly feature learning to transfer abnormal cues from references to queries, improving over normal-only prompting. However, we show that the residuals driving this transfer are often \textbf{domain-entangled}: they mix transferable anomaly cues with domain-specific factors such as texture, object geometry, and acquisition conditions. For example, a cracked-bottle residual encodes both the crack pattern and bottle-specific appearance; when transferred to leather or transistors, such domain-specific information can trigger false activations or suppress genuine anomalies. We call this phenomenon \textbf{residual domain entanglement} and identify it as a key bottleneck for cross-domain normal-abnormal guided GAD (see Figure~\ref{fig:entanglement}).

This issue relates GAD to domain adaptation~\cite{ben2010theory,ganin2016domain}, where domain-specific representations are known to generalize poorly under distribution shift. However, GAD requires more than feature alignment: normal-abnormal residuals must be disentangled, since anomaly cues and domain-specific factors are mixed within the same representation. 

We view this objective through an information-theoretic lens: the anomaly component $\mathbf{r}_a$ should retain information about the anomaly label $Y$, while minimizing information about the domain $D$, with reconstruction preserving the original residual. This motivates \textbf{\methname} \emph{(Domain-Agnostic Residual Transfer)}, a plug-in framework for normal-abnormal GAD. \methname combines two components. First, a Residual Disentanglement Module (RDM) splits a residual into an anomaly-relevant component $\mathbf{r}_a$ and a domain-specific component $\mathbf{r}_d$ using sufficiency, adversarial invariance, and reconstruction losses. Second, Cross-Domain Meta-Training (CDMT) forms episodes with references and queries from different source domains to make this separation learnable. RDM provides the mechanism for suppressing domain information, while CDMT provides the cross-domain signal needed to learn it.

RDM operates between residual computation and the downstream detection head, making \methname independent of a specific architecture. We apply it to NAGL~\cite{wangnormal}, the current normal-abnormal guided GAD baseline, and to adapted versions of InCTRL~\cite{zhu2024toward} and ResAD~\cite{yao2024resad}.

Our contributions are threefold: \textbf{(i)} we identify \textbf{residual domain entanglement} as a bottleneck in normal-abnormal guided GAD; \textbf{(ii)} we propose \methname, combining RDM and CDMT to learn anomaly-relevant, domain-agnostic residuals; and \textbf{(iii)} we evaluate it on MVTec AD, VisA, and BraTS across NAGL and two adapted residual-based GAD baselines, showing consistent image-level gains and competitive localization performance under cross-domain transfer.

\section{Related Work}
\label{sec:relatedwork}

\noindent\textbf{{Generalist Anomaly Detection.}}
Visual anomaly detection is mostly single-domain, using memory banks~\cite{roth2022towards,defard2021padim}, reconstruction~\cite{you2022unified}, knowledge distillation~\cite{deng2022anomaly,salehi2021multiresolution}, normalizing flows~\cite{yu2021fastflow,gudovskiy2022cflow}, or meta-learning~\cite{aqeel2025towards}. Few- and zero-shot methods~\cite{huang2022registration,jeong2023winclip,li2024promptad,aqeel2026anomaly, aqeel2025contrastive} and generative augmentation~\cite{nazir2025diffusion} reduce supervision, but often rely on category-specific adaptation or normal-only references. Generalist Anomaly Detection (GAD) instead trains one model on source domains and transfers it to unseen targets without fine-tuning. InCTRL~\cite{zhu2024toward} learns in-context residuals from few-shot normal prompts, while ResAD~\cite{yao2024resad} models residual feature distributions to reduce class-specific variation. Both rely on normal references, and recent work mines additional normality from the query itself~\cite{wang2026hunting}. NAGL~\cite{wangnormal} introduced normal-abnormal guided GAD, exploiting abnormal references via residual mining and anomaly feature learning. We build on this setting but address a different limitation: normal-abnormal residuals can remain entangled with domain-specific information, limiting cross-domain transfer.

\noindent\textbf{{Domain-Invariant Representation Learning.}}
Domain adaptation theory links target error to source error and cross-domain divergence~\cite{ben2010theory}, motivating domain-adversarial training with gradient reversal~\cite{ganin2016domain} and domain generalization methods~\cite{muandet2013domain,li2018learning}. Disentanglement approaches separate invariant and domain-specific factors using adversarial~\cite{peng2019domain,liu2018unified} or variational~\cite{kingma2013auto} objectives, while meta-learning studies cross-domain episodes for few-shot generalization~\cite{finn2017model,snell2017prototypical,tseng2020cross,cai2020cross}. Unlike these works, \methname neither disentangles raw features nor assumes target-domain access during training. It disentangles normal-abnormal residuals, separating anomaly-relevant changes from domain-specific factors for unseen-domain GAD.

\section{Methodology}
\label{sec:method}

\begin{figure}[!t]
    \centering
    \includegraphics[width=\textwidth]{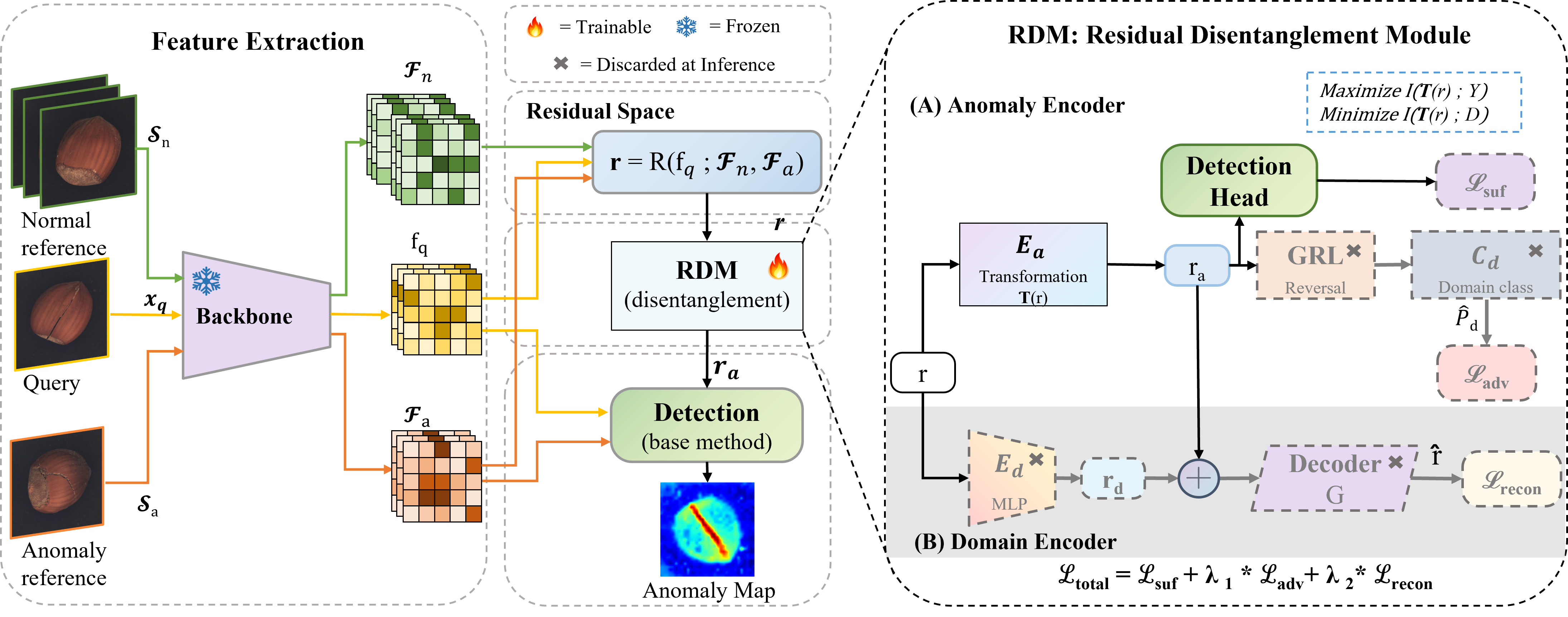}
     \caption{\textbf{Overview of DART.} \textit{(Left)} Given a query $\mathbf{x}_q$, normal references $\mathcal{S}_n$, and an abnormal reference $\mathcal{S}_a$, a frozen backbone produces features $\mathbf{f}_q$, $\mathcal{F}_n$, and $\mathcal{F}_a$. \textit{(Center)} The host residual operator computes $\mathbf{r} = \mathcal{R}(\mathbf{f}_q;\mathcal{F}_n, \mathcal{F}_a)$ in residual space; the proposed \textbf{Residual Disentanglement Module (RDM)} decomposes $\mathbf{r}$, and only its anomaly component $\mathbf{r}_a$ is passed, together with $\mathbf{f}_q$, $\mathcal{F}_n$, and $\mathcal{F}_a$, to the unchanged host detection head. \textit{(Right)} Inside RDM, the anomaly encoder $E_a$ (A) instantiates the transformation $T$ and extracts $\mathbf{r}_a$, while the domain encoder $E_d$ (B) extracts the domain-specific $\mathbf{r}_d$. Three losses shape the split: $\mathcal{L}_{\mathrm{suf}}$ keeps anomaly information in $\mathbf{r}_a$, $\mathcal{L}_{\mathrm{adv}}$ suppresses domain information via a GRL and domain classifier $C_d$, and $\mathcal{L}_{\mathrm{recon}}$ enforces that $G([\mathbf{r}_a;\mathbf{r}_d])$ ($\oplus$) reconstructs $\mathbf{r}$; $\lambda_1,\lambda_2$ denote $\lambda_{\mathrm{adv}},\lambda_{\mathrm{recon}}$. Components marked $\times$ ($E_d$, $G$, GRL, $C_d$) are discarded at inference, leaving only $E_a$ active.}
    \label{fig:pipeline}
\end{figure}


Generalist Anomaly Detection (GAD) learns a unified model from source domains $\mathcal{D}_S=\{D_1,\dots,D_K\}$ and applies it to unseen target domains $\mathcal{D}_T$ without parameter updates. We adopt the normal-abnormal guided setting~\cite{wangnormal}, where each episode contains a query image $\mathbf{x}_q$ and a reference set $\mathcal{S}=\{\mathcal{S}_n,\mathcal{S}_a\}$, with normal samples $\mathcal{S}_n=\{\mathbf{x}_{n,i}\}_{i=1}^{K_1}$ and abnormal samples $\mathcal{S}_a=\{\mathbf{x}_{a,j}\}_{j=1}^{K_2}$, $K_1\geq K_2$. During training, episodes are sampled from $\mathcal{D}_S$ and the model predicts an anomaly map for $\mathbf{x}_q$ conditioned on $\mathcal{S}$, using supervision $\mathbf{Y}_q$ that may contain image-level labels, pixel-level masks, or both. During inference, each target-domain query is paired with target-domain references, and no fine-tuning is performed. Thus, GAD is a conditional inference problem in which predictions depend on both the query and the reference set~\cite{wangnormal,zhu2024toward,yao2024resad}.

Residual-based GAD methods detect anomalies by measuring how query features deviate from reference features. Different methods instantiate this idea in different ways: InCTRL~\cite{zhu2024toward} uses discrepancy maps between query and normal prompts, ResAD~\cite{yao2024resad} models residual feature distributions, and NAGL~\cite{wangnormal} exploits residuals involving normal and abnormal references. We abstract these variants through a generic residual operator:
\begin{equation}
    \mathbf{r}=\mathcal{R}(\mathbf{f}_q;\mathcal{F}_n,\mathcal{F}_a),
    \label{eq:residual}
\end{equation}
where $\mathbf{f}_q=\phi(\mathbf{x}_q)$ is the query feature, $\mathcal{F}_n=\phi(\mathcal{S}_n)$ and $\mathcal{F}_a=\phi(\mathcal{S}_a)$ are normal and abnormal reference features, and $\mathcal{R}(\cdot)$ denotes the residual computation used by the host GAD method.

Our key observation is that residuals are not purely anomaly-specific: they also encode domain-dependent factors such as texture, geometry, and illumination. We call this \textbf{residual domain entanglement}. Under domain shift, these nuisance components can transfer poorly, causing spurious activations or missed detections in unseen target domains.

\subsection{Overview of \methname}
\label{sec:overview}

To reduce residual domain entanglement, we seek a transformation $T(\mathbf{r})=\mathbf{r}_a$ that preserves anomaly-relevant information while suppressing domain-specific variation. Let $Y$ denote the anomaly label and $D$ the domain identity. The transformed residual should be sufficient for anomaly prediction, while being as uninformative as possible about the domain:
\begin{equation}
    \max_T \ I(T(\mathbf{r});Y)-\beta I(T(\mathbf{r});D),
    \label{eq:mi_objective}
\end{equation}
where $I(\cdot;\cdot)$ denotes mutual information and $\beta$ controls the strength of domain invariance. The first term encourages predictive sufficiency, while the second penalizes domain information. Since enforcing invariance alone can remove useful anomaly cues, the representation must also preserve enough information to reconstruct the original residual.

This motivates \methname, a plug-in framework with two complementary components. The \textbf{Residual Disentanglement Module (RDM)} decomposes a raw residual into an anomaly-relevant component $\mathbf{r}_a$ and a domain-specific component $\mathbf{r}_d$, trained with sufficiency, adversarial invariance, and reconstruction losses. The \textbf{Cross-Domain Meta-Training (CDMT)} strategy forms episodes in which references and queries come from different source domains, making domain-specific cues unreliable and encouraging RDM to isolate transferable anomaly information. \methname is inserted after residual computation and before the downstream anomaly predictor. The scheme of the approach is in Fig.~\ref{fig:pipeline}.

\subsection{Residual Disentanglement Module}
\label{sec:rdm}

Given a raw residual $\mathbf{r} \in \mathbb{R}^{H \times W \times C}$ produced by the host GAD pipeline, RDM learns a decomposition of $\mathbf{r}$ into an anomaly component $\mathbf{r}_a$ and a complementary domain-specific component $\mathbf{r}_d$:
\begin{equation}
    (\mathbf{r}_a, \mathbf{r}_d) = \mathrm{RDM}(\mathbf{r}).
\end{equation}
Here, $\mathbf{r}_a$ is intended to retain the transferable information required for anomaly prediction, while $\mathbf{r}_d$ absorbs complementary information that is useful for reconstruction but should not drive detection. This decomposition is not enforced explicitly, but is encouraged through the joint training losses described below.

\noindent\textbf{Anomaly Encoder $E_a$.}
$E_a$ instantiates the transformation $T$ of Eq.~\eqref{eq:mi_objective}, $\mathbf{r}_a = E_a(\mathbf{r})$, as a transformer block with multi-head self-attention, feed-forward layers, residual connections, and layer normalization. Domain-specific appearance often manifests as broad spatial regularities, texture, material, illumination, while anomaly evidence is typically localized and sparse. Self-attention allows the model to compare local irregularities against global context, well suited to isolate transferable anomaly structure from domain-wide appearance.

\noindent\textbf{Domain Encoder $E_d$.}
$E_d$ is a lightweight two-layer MLP with GELU activation and layer normalization, $\mathbf{r}_d = E_d(\mathbf{r})$, whose role is to absorb residual information not required in $\mathbf{r}_a$ for domain-robust detection.

\noindent\textbf{Reconstruction Decoder $G$.}
To make the decomposition complete, i.e., to discard nothing from $\mathbf{r}$, $G$ reconstructs $\mathbf{r}$ from the channel-wise concatenation $[\mathbf{r}_a;\mathbf{r}_d]$, producing $\hat{\mathbf{r}} = G([\mathbf{r}_a;\mathbf{r}_d])$.

\noindent\textbf{Domain Classifier $C_d$.}
To explicitly suppress domain information in $\mathbf{r}_a$, we attach a spectrally normalized two-layer MLP to pooled invariant features and train it adversarially through a gradient reversal layer:
\begin{equation}
    \hat{\mathbf{p}}_d = C_d\!\big(\mathrm{GRL}_{\alpha}(\psi(\mathbf{r}_a))\big),
    \label{eq:domain_cls}
\end{equation}
where $\hat{\mathbf{p}}_d \in [0,1]^K$ is the predicted domain distribution and $\psi(\cdot)$ concatenates mean and max pooling over spatial locations. Mean pooling captures the dominant global domain signature, while max pooling exposes the most domain-discriminative local regions. Their combination provides a stronger probe for residual domain leakage.

\subsubsection{Training Objective}
\label{sec:training_objective}

The losses used in RDM are designed to instantiate two requirements: predictive sufficiency and domain invariance. A reconstruction term further enforces completeness of the decomposition.

\noindent\textbf{{Adversarial invariance loss.}}
To suppress domain information in $\mathbf{r}_a$, we train the anomaly branch to fool the domain classifier:
\begin{equation}
    \mathcal{L}_{\mathrm{adv}}
    =
    - \sum_{k=1}^{K}
    \mathbb{1}[d=k]
    \log \hat{p}_{d,k},
\end{equation}
where $d$ is the identity of the reference domain $D_i$, from which the residual inherits its appearance. Through gradient reversal, $C_d$ learns to predict the domain, while $E_a$ receives the opposite gradient and is pushed toward domain invariance. This acts as a practical surrogate for reducing the information about $D$ encoded in $\mathbf{r}_a$.

\noindent\textbf{{Sufficiency loss.}}
To ensure that $\mathbf{r}_a$ alone remains predictive of anomaly labels, we feed only $\mathbf{r}_a$ to the downstream detector:
\begin{equation}
    \mathcal{L}_{\mathrm{suf}}
    =
    \mathcal{L}_{\mathrm{det}}\!\big(\mathrm{Detect}(\mathbf{r}_a, \phi(\mathbf{x}_q)), \mathbf{Y}_q\big),
\end{equation}
where $\mathrm{Detect}(\cdot)$ denotes the prediction head of the underlying GAD method. This term encourages anomaly-relevant information to be concentrated in the anomaly branch rather than distributed across both branches.

\noindent\textbf{{Reconstruction loss.}}
To enforce completeness, we reconstruct $\mathbf{r}$ from $(\mathbf{r}_a, \mathbf{r}_d)$:
\begin{equation}
    \mathcal{L}_{\mathrm{recon}}
    =
    1 - \frac{1}{HW} \sum_{i=1}^{HW}
    \cos(\hat{\mathbf{r}}_i,\mathbf{r}_i).
\end{equation}
We use cosine similarity instead of $\ell_2$ loss because residuals can be noisy and vary substantially in scale. $\mathcal{L}_{\mathrm{recon}}$ constrains the pair $(\mathbf{r}_a,\mathbf{r}_d)$ rather than $\mathbf{r}_a$ alone; it is $\mathcal{L}_{\mathrm{suf}}$ that anchors anomaly information in $\mathbf{r}_a$ and prevents its collapse (Table~\ref{tab:loss_space}e). All branches are optimized jointly, while the relative influence of reconstruction is controlled by $\lambda_{\mathrm{recon}}$.

\noindent\textbf{{Total objective.}}
The final training loss is
\begin{equation}
    \mathcal{L}_{\mathrm{total}}
    =
    \mathcal{L}_{\mathrm{suf}}
    + \lambda_{\mathrm{adv}}  \mathcal{L}_{\mathrm{adv}}
    + \lambda_{\mathrm{recon}} \mathcal{L}_{\mathrm{recon}},
    \label{eq:loss}
\end{equation}
where $\lambda_{\mathrm{adv}} $ and $\lambda_{\mathrm{recon}}$ balance the strength of invariance and information preservation. From the perspective of the problem formulation, $\mathcal{L}_{\mathrm{suf}}$ enforces predictive sufficiency, $\mathcal{L}_{\mathrm{adv}}$ promotes domain invariance, and $\mathcal{L}_{\mathrm{recon}}$ enforces completeness of the decomposition.

\subsection{Cross-Domain Meta-Training}
\label{sec:cdmt}

RDM alone does not guarantee that $\mathbf{r}_a$ becomes domain-invariant. If training episodes are formed within a single domain, domain-specific cues in the residual remain predictive, and the model can achieve low training error without removing them. In the limiting case, the anomaly branch can simply preserve the full residual, i.e., $\mathbf{r}_a \approx \mathbf{r}$, while still achieving strong in-domain performance.

CDMT addresses this issue by constructing cross-domain episodes with probability $p_{\mathrm{cross}}$: the reference set $\mathcal{S}$ is sampled from domain $D_i$, while the query $\mathbf{x}_q$ is sampled from a different domain $D_j$, with $i \neq j$. Under this regime, domain-specific cues in the residual become unreliable for prediction, and the model is forced to rely on the transferable component of $\mathbf{r}$ instead. Combined with the adversarial invariance loss, CDMT provides the supervision needed to make the intended decomposition learnable in practice. In this way, RDM defines how residual information is separated, while CDMT provides the cross-domain signal that makes this separation meaningful.

\noindent\textbf{{Curriculum schedule.}}
Directly enabling cross-domain episodes and adversarial invariance from the first epoch is often unstable, because the model has not yet learned a stable residual decomposition. We therefore adopt a two-stage curriculum.

\textit{Phase 1, Warmup ($1$ to $T_w$):}
all episodes are same-domain, $p_{\mathrm{cross}}=0$, and adversarial training is disabled, $\alpha=0$. During this stage, the model learns a stable residual decomposition through $\mathcal{L}_{\mathrm{suf}}$ and $\mathcal{L}_{\mathrm{recon}}$.

\textit{Phase 2, Ramp-up ($T_w$ to $T$):}
the cross-domain sampling ratio $p_{\mathrm{cross}}$ increases linearly from $0$ to $p_{\max}$, and the gradient reversal strength follows
\begin{equation}
    \alpha(t)
    =
    \frac{2}{1 + \exp\!\left(-10 \cdot \frac{t-T_w}{T-T_w}\right)} - 1.
    \label{eq:grl_schedule}
\end{equation}
This gradual schedule prevents the adversarial objective from destabilizing the residual structure learned during warmup.

\noindent\textbf{{Plug-in Integration.}}
RDM operates on the residual representation after residual computation and before downstream detection. Given a base normal-abnormal GAD method, we first compute the raw residual $\mathbf{r}$ using the original pipeline, then apply RDM to obtain the anomaly component $\mathbf{r}_a$, and finally replace the original residual by $\mathbf{r}_a$ in the downstream anomaly detector. During training, the base objective is augmented as in Eq.~\eqref{eq:loss}, and episodes are constructed using CDMT. Since RDM preserves the residual interface, the backbone and the prediction head of the host method do not need to be redesigned.

\noindent\textbf{{Inference.}}
At inference time, the base GAD pipeline first computes the raw residual $\mathbf{r}$ from the target-domain references. The anomaly encoder then extracts $\mathbf{r}_a = E_a(\mathbf{r})$, which is passed to the downstream detector for anomaly prediction. The auxiliary training-only modules, namely the domain encoder $E_d$, the reconstruction decoder $G$, and the domain classifier $C_d$, are discarded. Therefore, the additional inference overhead over the base method is limited to the anomaly encoder $E_a$.

\section{Experiments}
\label{sec:exp}

\noindent\textbf{{Datasets and protocol.}}
We evaluate \methname in the GAD setting, where a model trained on source domains is deployed on unseen target domains without fine-tuning. We use MVTec AD~\cite{bergmann2019mvtec}, VisA~\cite{zou2022spot}, and BraTS~\cite{menze2014multimodal}, covering industrial and medical anomaly detection.
We adopt the same protocol of~\cite{wangnormal}: we use one dataset for training and another for testing, ensuring category-, dataset-, and acquisition-level isolation. Specifically, we consider VisA$\to$MVTec AD, MVTec AD$\to$VisA, and MVTec AD$\to$BraTS. At inference, references are sampled from the target-domain training split and no parameters are updated.

\noindent\textbf{{CDMT protocol \& reference selection.}}
In CDMT, a \emph{domain} is an object category, not a dataset: $\mathcal{D}_S$ holds 15 categories when MVTec AD is the source and 12 when VisA is, and the target dataset is held out entirely (App.~\ref{sec:supp:leakage}). Each episode draws references from a source category $D_i$ and the query from a category $D_j$ (Table~\ref{tab:episodes}); episodes are same-domain ($D_j{=}D_i$) during a $T_w{=}5$-epoch warmup, after which the cross-domain probability ramps linearly to $p_{\max}{=}0.3$ (Sec.~3.3). The domain label in $\mathcal{L}_{\mathrm{adv}}$ is the reference domain $D_i$, whose appearance the residual inherits; the query domain does not enter the adversarial loss. At inference, query and references come from the same held-out target category. The abnormal reference is a single image ($A^1$, $K_2{=}1$), sampled by the query's defect type on MVTec AD and uniformly on VisA.

\begin{table}[!h]
\centering
\small
\setlength{\tabcolsep}{6pt}
\caption{Representative CDMT episode triplets $(D_i, D_j, D_k)$. The target $D_k$ never appears as reference or query during training.}
\label{tab:episodes}
\begin{tabular}{llll}
\toprule
Target $D_k$ & Reference $D_i$ & Query $D_j$ & Type \\
\midrule
VisA--candle   & MVTec--bottle     & MVTec--bottle  & same-domain (warmup) \\
VisA--candle   & MVTec--bottle     & MVTec--zipper  & cross-domain \\
VisA--pcb1     & MVTec--transistor & MVTec--cable   & cross-domain \\
MVTec--bottle  & VisA--candle      & VisA--pcb2     & cross-domain \\
MVTec--wood    & VisA--macaroni1   & VisA--fryum    & cross-domain \\
BraTS--brain   & MVTec--pill       & MVTec--capsule & cross-domain \\
BraTS--brain   & MVTec--hazelnut   & MVTec--screw   & cross-domain \\
\bottomrule
\end{tabular}
\end{table}

\noindent\textbf{{Metrics.}}
For MVTec AD and VisA, we report image-level AUROC, AP, and F1-max, and pixel-level AUROC, PRO~\cite{bergmann2020uninformed}, and F1-max. For BraTS, we report image-level AUROC, pixel-level AUROC, and their mean.

\noindent\textbf{{Host methods \& baselines.}}
We integrate \methname into NAGL~\cite{wangnormal}, the current normal-abnormal guided GAD baseline, and into adapted versions of two residual-based GAD methods: InCTRL~\cite{zhu2024toward} and ResAD~\cite{yao2024resad}. This allows us to evaluate whether residual disentanglement improves both an existing normal-abnormal method and normal-only GAD pipelines extended with abnormal references.
We compare against two groups of methods. The first comprises established few-shot AD approaches that use \textit{normal references only}: SPADE~\cite{cohen2020sub}, WinCLIP~\cite{jeong2023winclip}, and PromptAD~\cite{li2024promptad}, along with the vanilla (unmodified) versions of our three host methods InCTRL, ResAD, and NAGL. The second group evaluates the effect of introducing abnormal references: for each host method we report \textbf{+Inv}, which naively incorporates the abnormal reference without disentanglement, and \textbf{+DART}, which applies the full RDM pipeline. Comparing vanilla $\to$ +Inv isolates the value of abnormal access alone, while comparing +Inv $\to$ +DART isolates the contribution of residual disentanglement.

\noindent\textbf{{\methname configuration.}}
\label{sec:implementation}
\methname operates on the raw residual produced by each host method and leaves the downstream detection head unchanged. RDM contains an anomaly encoder $E_a$, a domain encoder $E_d$, a reconstruction decoder $G$, and a domain classifier $C_d$ trained through gradient reversal. The objective combines anomaly sufficiency, adversarial domain invariance, and residual reconstruction. Training follows CDMT, with cross-domain episodes sampled from source domains only; target domains are never used for training, validation, or checkpoint selection. Host-method parameters remain frozen, and at inference only $E_a$ and the host detection head are retained. Implementation details and hyperparameters are provided in the supplementary material.

\begin{table*}[!t]
\centering
\caption{Comparison on MVTecAD and VisA datasets. $N^i$: $i$ normal references; $A^1$: one abnormal reference. \textbf{+Inv}: naive abnormal-reference integration without RDM. \textbf{Bold}/\underline{underline}: best/runner-up per setting block. Shaded rows are our \textbf{DART} variants.}
\label{tab:mvtec_visa}
\resizebox{\textwidth}{!}{%
\setlength{\tabcolsep}{3.5pt}
\renewcommand{\arraystretch}{1.0}
\begin{tabular}{ll cccccc cccccc}
\toprule
& & \multicolumn{6}{c}{\textbf{MVTecAD}} & \multicolumn{6}{c}{\textbf{VisA}} \\
\cmidrule(lr){3-8}\cmidrule(lr){9-14}
& & \multicolumn{3}{c}{Image-level} & \multicolumn{3}{c}{Pixel-level}
  & \multicolumn{3}{c}{Image-level} & \multicolumn{3}{c}{Pixel-level} \\
\cmidrule(lr){3-5}\cmidrule(lr){6-8}\cmidrule(lr){9-11}\cmidrule(lr){12-14}
Setting & Method
  & AUROC & AP & F1-max & AUROC & PRO & F1-max
  & AUROC & AP & F1-max & AUROC & PRO & F1-max \\
\midrule

\multirow{5}{*}{$N^1$}
  & SPADE~\cite{cohen2020sub}       & 81.0  & 90.6  & 90.3  & 91.2  & 83.9  & 42.4  & 79.5  & 82.0  & 80.7  & 95.6  & 84.1  & 35.5 \\
& WinCLIP~\cite{jeong2023winclip}   & 93.1  & 96.5  & 93.7  & 95.2  & 87.1  & 55.9  & 83.8  & 85.1  & 83.1  & 96.4  & 85.1  & \underline{41.3} \\
& PromptAD~\cite{li2024promptad}    & 94.6  & 97.1  & --    & 95.9  & 87.9  & --    & \underline{86.9} & \textbf{88.4} & --   & 96.7  & 85.1  & -- \\
\cmidrule(l){2-14}
& InCTRL~\cite{zhu2024toward}       & 94.99 & 97.33 & 95.07 & --    & --    & --    & 85.47 & 55.65 & 60.04 & --    & --    & -- \\
& ResAD~\cite{yao2024resad}         & 85.70 & 93.10 & 90.90 & 94.40 & 84.50 & 49.80 & 81.20 & 82.50 & 81.20 & 95.20 & 78.80 & 33.30 \\
\midrule

\multirow{3}{*}{$N^1\!+\!A^1$}
  & NAGL~\cite{wangnormal}          & \underline{95.54} & \underline{97.63} & \underline{95.40} & \textbf{96.53} & \underline{92.66} & \underline{58.03} & 86.23 & \underline{86.54} & \textbf{84.59} & \underline{97.20} & \underline{88.76} & \textbf{42.41} \\
  & InCTRL+Inv                      & 95.41 & 97.72 & 94.97 & --    & --    & --    & 79.87 & 55.37 & 57.95 & --    & --    & -- \\
  & ResAD+Inv                       & 82.70 & 91.10 & 89.90 & 93.20 & 81.00 & 46.60 & 74.70 & 76.70 & 78.20 & 94.00 & 72.80 & 25.30 \\
\cmidrule(l){2-14}
\rowcolor{dartrow}
  & \textbf{NAGL+DART}              & \textbf{95.93} & \textbf{97.95} & \textbf{95.76} & \underline{96.51} & \textbf{92.88} & \textbf{58.37} & \textbf{87.65} & \textbf{88.00} & \underline{84.25} & \textbf{97.21} & \textbf{88.77} & 41.25 \\
\rowcolor{dartrow}
\textit{Ours}
  & \textbf{InCTRL+DART}            & 95.55 & 97.74 & 95.21 & 94.52 & 85.23 & 51.16 & 86.06 & 60.10 & 63.31 & 95.44 & 69.15 & 27.24 \\
\rowcolor{dartrow}
  & \textbf{ResAD+DART}             & 85.90 & 93.10 & 90.30 & 94.10 & 84.00 & 50.30 & 81.70 & 84.30 & 81.50 & 95.40 & 78.20 & 28.50 \\
\midrule

\multirow{5}{*}{$N^2$}
  & SPADE~\cite{cohen2020sub}       & 82.9  & 91.7  & 91.1  & 92.0  & 85.7  & 44.5  & 80.7  & 82.3  & 81.7  & 96.2  & 85.7  & 40.5 \\
& WinCLIP~\cite{jeong2023winclip}   & 94.4  & 97.0  & 94.4  & 96.0  & 88.4  & 58.4  & 84.6  & 85.8  & 83.0  & 96.8  & 86.2  & \underline{43.5} \\
& PromptAD~\cite{li2024promptad}    & 95.7  & 97.9  & --    & 96.2  & 88.5  & --    & 88.3  & \textbf{90.0} & --   & 97.1  & 85.8  & -- \\
\cmidrule(l){2-14}
& InCTRL~\cite{zhu2024toward}       & 95.06 & 97.55 & 95.76 & --    & --    & --    & 87.96 & 58.67 & 63.61 & --    & --    & -- \\
& ResAD~\cite{yao2024resad}         & 87.60 & 94.10 & 91.70 & 95.30 & 86.30 & 51.60 & 83.20 & 84.20 & 82.20 & 95.90 & 80.40 & 37.40 \\
\midrule

\multirow{3}{*}{$N^2\!+\!A^1$}
  & NAGL~\cite{wangnormal}          & \underline{96.61} & \underline{98.11} & \underline{96.10} & \underline{96.63} & \textbf{93.13} & \underline{59.20} & \underline{87.90} & \underline{89.03} & \underline{85.15} & \textbf{97.34} & \textbf{89.25} & \textbf{44.06} \\
  & InCTRL+Inv                      & 95.73 & 97.88 & 95.33 & --    & --    & --    & 81.41 & 56.55 & 57.96 & --    & --    & -- \\
  & ResAD+Inv                       & 84.90 & 92.20 & 90.10 & 94.20 & 83.10 & 48.10 & 77.60 & 79.80 & 78.90 & 94.70 & 75.30 & 28.60 \\
\cmidrule(l){2-14}
\rowcolor{dartrow}
  & \textbf{NAGL+DART}              & \textbf{96.62} & \textbf{98.15} & \textbf{96.20} & \textbf{96.64} & \textbf{93.13} & \textbf{59.22} & \textbf{89.72} & \textbf{89.16} & \textbf{86.21} & \underline{97.29} & \underline{89.07} & 42.65 \\
\rowcolor{dartrow}
\textit{Ours}
  & \textbf{InCTRL+DART}            & 96.21 & 98.12 & 95.32 & 94.81 & 85.65 & 52.37 & 88.64 & 63.72 & 65.73 & 96.37 & 70.45 & 28.04 \\
\rowcolor{dartrow}
  & \textbf{ResAD+DART}             & 88.10 & 94.30 & 91.80 & 95.00 & 85.50 & 52.10 & 83.40 & 85.60 & 82.40 & 96.20 & 80.80 & 32.40 \\
\midrule

\multirow{5}{*}{$N^4$}
  & SPADE~\cite{cohen2020sub}       & 84.8  & 92.5  & 91.5  & 92.7  & 87.0  & 46.2  & 81.7  & 83.4  & 82.1  & 96.6  & 87.3  & 43.6 \\
& WinCLIP~\cite{jeong2023winclip}   & 95.2  & 97.3  & 94.7  & 96.2  & 89.0  & 59.5  & 87.3  & 88.8  & 84.2  & 97.2  & 87.6  & \textbf{47.0} \\
& PromptAD~\cite{li2024promptad}    & 96.6  & 98.5  & --    & 96.5  & 90.5  & --    & 89.1  & \underline{90.8} & --   & 97.4  & 86.2  & -- \\
\cmidrule(l){2-14}
& InCTRL~\cite{zhu2024toward}       & 95.41 & 97.44 & 95.95 & --    & --    & --    & 89.16 & 63.67 & 67.92 & --    & --    & -- \\
& ResAD~\cite{yao2024resad}         & 89.90 & 95.10 & 92.50 & 95.90 & 87.50 & 53.40 & 87.00 & 87.90 & 84.60 & 96.40 & 82.80 & 39.50 \\
\midrule

\multirow{3}{*}{$N^4\!+\!A^1$}
  & NAGL~\cite{wangnormal}          & \underline{97.08} & \underline{98.34} & \underline{96.37} & \underline{96.68} & \underline{93.29} & \underline{59.77} & \underline{90.30} & 90.70 & \underline{87.03} & \underline{97.44} & \underline{89.64} & \underline{44.86} \\
  & InCTRL+Inv                      & 94.19 & 97.43 & 94.46 & --    & --    & --    & 80.29 & 55.49 & 55.48 & --    & --    & -- \\
  & ResAD+Inv                       & 87.40 & 93.50 & 91.00 & 94.90 & 84.80 & 49.30 & 82.00 & 83.40 & 81.30 & 95.80 & 79.00 & 31.70 \\
\cmidrule(l){2-14}
\rowcolor{dartrow}
  & \textbf{NAGL+DART}              & \textbf{97.34} & \textbf{98.46} & \textbf{96.52} & \textbf{96.76} & \textbf{93.39} & \textbf{60.03} & \textbf{91.29} & \textbf{90.94} & \textbf{87.87} & \textbf{97.45} & \textbf{89.99} & 43.56 \\
\rowcolor{dartrow}
\textit{Ours}
  & \textbf{InCTRL+DART}            & 95.64 & 97.76 & 96.08 & 95.52 & 86.22 & 53.31 & 90.14 & 69.95 & 70.05 & 96.92 & 71.18 & 29.47 \\
\rowcolor{dartrow}
  & \textbf{ResAD+DART}             & 89.90 & 95.10 & 92.50 & 95.60 & 86.80 & 53.70 & 87.90 & 89.30 & 85.70 & 96.70 & 83.10 & 35.20 \\

\bottomrule
\end{tabular}%
}
\end{table*}

\section{Results}
\paragraph{State-of-the-art comparison.}
Table~\ref{tab:mvtec_visa} compares DART against existing methods on the MVTec AD and VisA benchmarks. NAGL+DART achieves the highest image-level AUROC across all shot settings on both datasets. On MVTec AD, NAGL+DART reaches 95.93\%, 96.62\%, and 97.34\% at $N^1$, $N^2$, and $N^4$ respectively, consistently outperforming all baselines including PromptAD (94.6\%, 95.7\%, 96.6\%) and vanilla NAGL. On VisA, where the domain gap from the MVTec-trained source is largest, the improvements are more pronounced: NAGL+DART achieves 89.72\% at $N^2{+}A^1$, a +1.82\% gain over vanilla NAGL (87.90\%), and 91.29\% at $N^4{+}A^1$ (+0.99\% over NAGL). This pattern, larger gains under larger domain shift, is consistent with our hypothesis that residual domain entanglement is the primary bottleneck, and that its impact scales with the dissimilarity between source and target distributions. Gains are also most pronounced at low-shot settings ($N^1$, $N^2$) where fewer target-domain references make cross-domain transfer most critical, and gradually narrow as additional references partially compensate for entanglement.

At the pixel level, NAGL+DART matches or improves upon vanilla NAGL on PRO and P-AUROC across most settings, while simultaneously achieving the best image-level metrics. This indicates that disentanglement does not sacrifice localization quality for classification gains, a common failure mode when modifying residual representations. The pixel-level F1-max shows occasional small drops (\emph{e.g.}, $-$1.16 at $N^1$ on VisA), which we attribute to F1's threshold sensitivity rather than to a systematic degradation, as both P-AUROC and PRO remain competitive.

\noindent\textbf{Disentanglement vs.\ naive abnormal access.}
The +Inv rows in Table~\ref{tab:mvtec_visa} separate \emph{access} to abnormal references from \emph{disentangling} them. Naive integration is unreliable and often harmful: ResAD+Inv loses $-$3.0 I-AUROC on MVTec AD and $-$6.5 on VisA at $N^1$ (PRO up to $-$6.0), and InCTRL+Inv swings from +0.42 at $N^1$ to $-$1.22 at $N^4$ without ever producing spatially coherent pixel-level maps. DART recovers and exceeds vanilla performance in the same settings, and the +Inv$\to$+DART gap widens with domain shift: +6.19 I-AUROC on VisA at $N^1$ (79.87\% vs.\ 86.06\%), growing to +9.85 at $N^4$ (80.29\% vs.\ 90.14\%). Residual disentanglement, not mere abnormal access, drives the improvement; the full per-category breakdown is in App.~\ref{sec:supp:disentanglement}.

\noindent\textbf{DART as a general plug-in.}
RDM operates on the residual $\mathbf{r}$ as a method-agnostic interface, requiring no modification to detection heads or method-specific hyperparameter tuning beyond $\lambda_{\mathrm{adv}}$ and $\lambda_{\mathrm{recon}}$. This design choice is validated by the breadth of host methods: NAGL mines attention-based residuals over ViT patch tokens, InCTRL computes multi-layer in-context residuals from CLIP features, and ResAD learns distributional deviations via normalizing flows on WideResNet features, three substantially different architectures, all of which improve under DART without any method-specific adaptation. A notable emergent capability is pixel-level localization in InCTRL, which originally reported no segmentation metrics: disentangling the residual produces spatially meaningful anomaly maps as a byproduct, achieving P-AUROC of 94.52--95.71\% on MVTec AD across shot settings.

\noindent\textbf{Efficiency.}
DART's overhead is modest and front-loaded: training adds 3.1--20.1M parameters and ${\approx}$1 GPU-hour per setting on a single RTX4090, incurred once on source domains with the host frozen. At inference only $E_a$ is retained (+0.82M for InCTRL, +0.89M for NAGL, +9.06M for ResAD) and operates on low-dimensional residuals, so runtime overhead is negligible (Table~\ref{tab:supp:efficiency}).

\subsection{Industrial-to-Medical Transfer}
Table~\ref{tab:brats} evaluates DART on the most challenging setting, MVTec AD $\to$ BraTS (industrial defects $\to$ brain-tumor MRI). DART improves all three host methods on this extreme domain shift in nearly every setting, with gains substantially larger than those on industrial benchmarks. InCTRL+DART achieves the most striking improvement: +13.28\% image-level AUROC at $N^1$ (65.20\% $\to$ 78.48\%) and +7.61\% at $N^4$ (69.89\% $\to$ 77.50\%), while ResAD+DART improves by +5.20\% at $N^2$ and NAGL+DART improves at both $N^2$ (+0.25\%) and $N^4$ (+0.99\%). The gains extend to localization: InCTRL+DART raises pixel-level AUROC from 83.65\% to 87.67\% at $N^1$ and from 90.06\% to 93.85\% at $N^4$. The single exception is NAGL+DART at $N^1$, where a pixel-level drop (93.00\% $\to$ 90.33\%) lowers the mean despite a slightly higher image-level AUROC; with two or more normal references it improves both. The magnified gains on BraTS support our core claim that the impact of residual domain entanglement scales with the severity of the domain shift.
\subsection{Ablation Study}
\label{sec:ablation}

We ablate DART's components using InCTRL as the host method at $k{=}2$ shots. All ablations use the same training protocol; only the specified component is modified.

\begin{table*}[t]
\centering
\caption{%
    \textbf{(a)}~Cross-domain evaluation on BraTS (MVTec AD $\to$ BraTS). I/P/M: image-/pixel-level AUROC and their mean. \textbf{Bold}/\underline{underline}: best/runner-up per $N^k{+}A^1$ block. \colorbox{dartrow}{\strut Shaded}: DART variants. \textbf{(b)}~Encoder architecture ablation on MVTec-AD and VisA. \textbf{(c)}~Disentanglement strategy on MVTec-AD and VisA. \textbf{(d)}~RDM--CDMT co-dependency. All ablations: InCTRL backbone, $k{=}2$.
}
\label{tab:brats_arch_strategy_codep}
\setlength{\tabcolsep}{3pt}
\renewcommand{\arraystretch}{1.05}
\begin{minipage}[t]{0.48\textwidth}
\centering
\small
\textbf{(a) BraTS state-of-the-art comparison}\\[4pt]
\label{tab:brats}
\scalebox{0.90}{%
\begin{tabular}{ll ccc}
\toprule
Setting & Method & I & P & M \\
\midrule
\multirow{2}{*}{$N^1$}
  & InCTRL~\cite{zhu2024toward}      & 65.20 & 83.65 & 74.43 \\
  & ResAD~\cite{yao2024resad}       & 72.30 & 93.40 & 82.85 \\
\midrule
$N^1\!+\!A^1$
  & NAGL~\cite{wangnormal}        & 76.32 & \underline{93.00} & \textbf{84.66} \\
\cmidrule(l){2-5}
\rowcolor{dartrow}
  & \textbf{NAGL+DART}   & \underline{76.36} & 90.33 & 83.35 \\
\rowcolor{dartrow}
  & \textbf{InCTRL+DART} & \textbf{78.48} & 87.67 & 83.08 \\
\rowcolor{dartrow}
  & \textbf{ResAD+DART}  & 75.70 & \textbf{93.60} & \underline{84.65} \\
\midrule
\multirow{4}{*}{$N^2$}
  & SPADE~\cite{cohen2020sub}       & 58.0  & 92.8  & 75.4 \\
  & WinCLIP~\cite{jeong2023winclip}     & 55.9  & 91.5  & 73.7 \\
\cmidrule(l){2-5}
  & InCTRL~\cite{zhu2024toward}      & 59.80 & 87.28 & 73.54 \\
  & ResAD~\cite{yao2024resad}       & 66.20 & 93.30 & 79.75 \\
\midrule
$N^2\!+\!A^1$
  & NAGL~\cite{wangnormal}        & \underline{76.85} & 89.57 & \underline{83.21} \\
\cmidrule(l){2-5}
\rowcolor{dartrow}
  & \textbf{NAGL+DART}   & \textbf{77.10} & 89.95             & \textbf{83.53} \\
\rowcolor{dartrow}
  & \textbf{InCTRL+DART} & 64.11          & \underline{92.20} & 78.16 \\
\rowcolor{dartrow}
  & \textbf{ResAD+DART}  & 71.40 & \textbf{93.80} & 82.60 \\
\midrule
\multirow{4}{*}{$N^4$}
  & SPADE~\cite{cohen2020sub}       & 66.3  & 94.8  & 80.6 \\
  & WinCLIP~\cite{jeong2023winclip}     & 67.3  & 93.2  & 80.3 \\
\cmidrule(l){2-5}
  & InCTRL~\cite{zhu2024toward}      & 69.89 & 90.06 & 79.98 \\
  & ResAD~\cite{yao2024resad}       & 69.50 & 93.40 & 81.45 \\
\midrule
$N^4\!+\!A^1$
  & NAGL~\cite{wangnormal}        & \underline{78.29} & 93.33 & \underline{85.81} \\
\cmidrule(l){2-5}
\rowcolor{dartrow}
  & \textbf{NAGL+DART}   & \textbf{79.28} & 92.54             & \textbf{85.91} \\
\rowcolor{dartrow}
  & \textbf{InCTRL+DART} & 77.50          & \underline{93.85} & 85.68 \\
\rowcolor{dartrow}
  & \textbf{ResAD+DART}  & 72.60 & \textbf{93.90} & 83.25 \\
\bottomrule
\end{tabular}}
\end{minipage}%
\hfill
\begin{minipage}[t]{0.48\textwidth}
\centering
\small
\textbf{(b) Architecture (MVTec-AD / VisA)}\\[4pt]
\label{tab:ablation_arch}
\scalebox{0.87}{%
\begin{tabular}{lcccc}
\toprule
\multirow{2}{*}{Variant}
    & \multicolumn{2}{c}{MVTec-AD}
    & \multicolumn{2}{c}{VisA} \\
\cmidrule(lr){2-3} \cmidrule(lr){4-5}
    & I-AUROC & I-AP & I-AUROC & I-AP \\
\midrule
Vanilla (InCTRL)     & 95.06 & 97.55 & 87.96 & 58.67 \\
\midrule
Linear projection    & 95.16 & 97.57 & 81.29 & 58.25 \\
w/o LayerNorm        & 94.30 & 97.31 & 84.65 & 59.18 \\
\midrule
\textbf{DART (full)} & \textbf{96.21} & \textbf{98.12} & \textbf{88.64} & \textbf{63.72} \\
\bottomrule
\end{tabular}}
\\[10pt]
\textbf{(c) Strategy (MVTec-AD / VisA)}\\[4pt]
\label{tab:ablation_strategy}
\scalebox{0.87}{%
\begin{tabular}{lcccc}
\toprule
\multirow{2}{*}{Method}
    & \multicolumn{2}{c}{MVTec-AD}
    & \multicolumn{2}{c}{VisA} \\
\cmidrule(lr){2-3} \cmidrule(lr){4-5}
    & I-AUROC & I-AP & I-AUROC & I-AP \\
\midrule
Vanilla              & 95.06 & 97.55 & 87.96 & 58.67 \\
\midrule
Adversarial-only     & 93.61 & 97.12 & 86.15 & 61.95 \\
Orthogonality-only   & 93.32 & 96.89 & 85.94 & 61.79 \\
\midrule
\textbf{DART (ours)} & \textbf{96.21} & \textbf{98.12} & \textbf{88.64} & \textbf{63.72} \\
\bottomrule
\end{tabular}}
\\[10pt]
\textbf{(d) RDM--CDMT co-dependency}\\[4pt]
\label{tab:ablation_codep}
\scalebox{0.87}{%
\begin{tabular}{ccccccc}
\toprule
\multirow{2}{*}{\textbf{RDM}} & \multirow{2}{*}{\textbf{CDMT}}
    & \multicolumn{2}{c}{\textbf{MVTec-AD}}
    & \multicolumn{2}{c}{\textbf{VisA}} \\
\cmidrule(lr){3-4} \cmidrule(lr){5-6}
    & & I-AUROC & $\Delta$
      & I-AUROC & $\Delta$ \\
\midrule
\xmark & \xmark & 95.06          & --              & 87.96           & --     \\
\xmark & \cmark & 95.06          & $+$0.00          & 87.96          & $+$0.00 \\
\cmark & \xmark & 95.54          & $+$0.48          & 88.17          & $+$0.21 \\
\cmark & \cmark & \textbf{96.21} & $+$\textbf{1.15} & \textbf{88.64} & $+$\textbf{0.68} \\
\bottomrule
\end{tabular}}
\end{minipage}
\end{table*}

\noindent\textbf{{Architecture (Table~\ref{tab:brats_arch_strategy_codep}b).}}
Replacing the MLP encoder with a linear projection causes a sharp drop on VisA ($-$7.35 I-AUROC) while leaving MVTec-AD nearly unchanged, demonstrating that non-linear capacity is essential specifically for cross-domain disentanglement where the domain gap is larger. Removing LayerNorm degrades both datasets ($-$0.75 MVTec-AD, $-$3.99 VisA), confirming its role in stabilizing the residual decomposition.

\noindent\textbf{{Disentanglement strategy (Table~\ref{tab:brats_arch_strategy_codep}c).}}
Both adversarial-only and orthogonality-only baselines degrade below vanilla on both datasets (up to $-$1.74 MVTec-AD, $-$2.02 VisA). Adversarial-only training suppresses domain information but lacks a constraint to preserve anomaly signal, allowing $E_a$ to collapse. Orthogonality-only training enforces geometric independence but permits anomaly-relevant directions to migrate into $\mathbf{r}_d$. DART's joint objective resolves both: $\mathcal{L}_{\text{suf}}$ prevents collapse, $\mathcal{L}_{\text{recon}}$ ensures completeness, and $\mathcal{L}_{\text{adv}}$ drives genuine domain suppression under these constraints. The +2.89 gap over the best single-objective baseline confirms that no proper subset achieves the same effect.

\noindent\textbf{{Co-dependency of RDM and CDMT (Table~\ref{tab:brats_arch_strategy_codep}d).}}
CDMT without RDM is inert: with the backbone frozen, cross-domain episodes have no trainable parameters to update, so results are identical to Vanilla ($\Delta{=}+0.00$ on both datasets). RDM without CDMT achieves only local disentanglement ($+$0.48 MVTec-AD, $+$0.21 VisA), learning to suppress within-domain variance but not generalising across domains. Full DART achieves $+1.15$ on MVTec-AD and $+0.68$ on VisA, with the gains of $+0.67$ and $+0.47$ over RDM alone attributable entirely to CDMT's cross-domain supervision. The co-dependency is directional and structural: no proper subset achieves cross-domain invariance.

\noindent\textbf{{Loss terms (Table~\ref{tab:loss_space}e).}}
Removing $\mathcal{L}_{\text{suf}}$ causes the largest degradation ($-$2.64 I-AUROC), confirming that anomaly sufficiency is the most critical constraint: without it, $E_a$ has no explicit objective to retain discriminative information in $\mathbf{r}_a$. Removing $\mathcal{L}_{\text{adv}}$, $\mathcal{L}_{\text{recon}}$, or CDMT individually produces smaller but consistent drops ($-$0.64 to $-$0.73), in line with their role as cross-domain regularizers whose benefit compounds under domain shift. The full model outperforms every partial variant, confirming that all components are jointly necessary. The same setting ($\lambda_{\mathrm{adv}}{=}0.1$, $\lambda_{\mathrm{recon}}{=}0.5$) is used for all hosts and datasets with no per-host or per-dataset tuning.

\noindent\textbf{{Residual vs.\ feature space (Table~\ref{tab:loss_space}f, Fig.~\ref{fig:sil}).}}
DART-F, which applies disentanglement in backbone feature space, \textit{degrades} performance ($-$0.54 vs.\ vanilla) while DART improves it (+1.15). Silhouette analysis of the same 640-d residuals (identical normalization, no detector in the loop) reveals why: DART-F \textit{amplifies} domain separation (0.246 $\to$ 0.611) because backbone features are domain-discriminative by pretraining design and a lightweight plug-in cannot override this geometry, whereas in residual space domain information is only incidental and DART suppresses it (0.246 $\to$ 0.192). Relabeling the same residuals by defect type (computed within each category and averaged, as MVTec AD defect types are category-specific), the silhouette rises from 0.08 to 0.25 under DART while DART-F stalls at 0.10: domain identity becomes a weaker axis of variation in $\mathbf{r}_a$ and defect identity a stronger one, \emph{demotion} rather than deletion of domain information, and we do not claim perfect invariance. Anomaly preservation is enforced structurally: the frozen host head reads only $\mathbf{r}_a$, so any direction it uses cannot be discarded without paying $\mathcal{L}_{\mathrm{suf}}$; removing $\mathcal{L}_{\mathrm{suf}}$ indeed yields this degenerate solution, invariance by discarding, falling below the unmodified host (Table~\ref{tab:loss_space}e).

\begin{table*}[t]
\centering
\setlength{\tabcolsep}{5pt}
\renewcommand{\arraystretch}{1.05}
\begin{minipage}[c]{0.40\textwidth}
\centering
\small
\captionof{table}{\textbf{(e)}~Loss term ablation; $\Delta$ vs.\ DART (full). \textbf{(f)}~Residual vs.\ feature space; $\Delta$ vs.\ Vanilla. MVTec-AD, InCTRL, $k{=}2$.}
\label{tab:loss_space}
\textbf{(e) Loss terms}\\[3pt]
\label{tab:ablation_loss}
\scalebox{0.80}{%
\begin{tabular}{lccc}
\toprule
Variant & I-AUROC & I-AP & $\Delta$ \\
\midrule
Vanilla (InCTRL)                       & 95.06 & 97.55 & $-$1.15 \\
\midrule
w/o $\mathcal{L}_{\text{suf}}$         & 93.57 & 96.90 & $-$2.64 \\
w/o $\mathcal{L}_{\text{recon}}$       & 95.48 & 97.82 & $-$0.73 \\
w/o $\mathcal{L}_{\text{adv}}$         & 95.57 & 97.86 & $-$0.64 \\
w/o CDMT                               & 95.54 & 97.84 & $-$0.67 \\
\midrule
\textbf{DART (full)}                   & \textbf{96.21} & \textbf{98.12} & -- \\
\bottomrule
\end{tabular}}
\\[8pt]
\textbf{(f) Disentanglement space}\\[3pt]
\label{tab:ablation_space}
\scalebox{0.80}{%
\begin{tabular}{lccc}
\toprule
Method & I-AUROC & I-AP & $\Delta$ \\
\midrule
Vanilla              & 95.06 & 97.55 & -- \\
DART-F (feature)     & 94.52 & 97.32 & $-$0.54 \\
\textbf{DART (ours)} & \textbf{96.21} & \textbf{98.12} & $+$1.15 \\
\bottomrule
\end{tabular}}
\end{minipage}%
\hfill
\begin{minipage}[c]{0.57\textwidth}
\centering
\includegraphics[width=\linewidth]{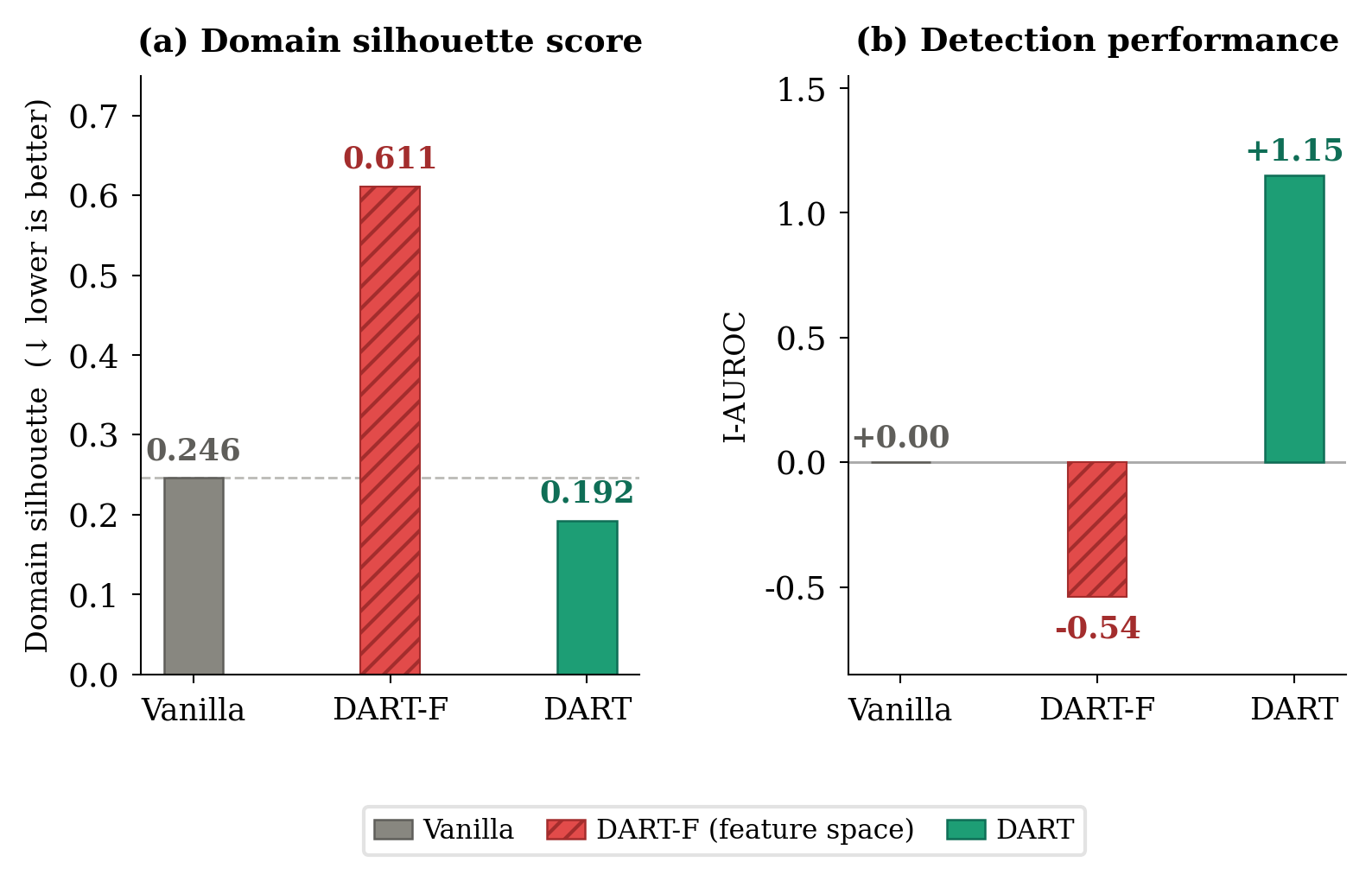}
\captionof{figure}{\textbf{Residual vs.\ feature space.} (a)~Domain silhouette (lower = less entanglement); (b)~I-AUROC gain over Vanilla (MVTec AD, $k{=}2$). DART-F amplifies domain separation; DART reduces it, improving detection.}
\label{fig:sil}
\end{minipage}
\end{table*}

\section{Conclusion}
In this work, we identified residual domain entanglement as a fundamental bottleneck limiting cross-domain transfer in normal-abnormal guided Generalist Anomaly Detection. To address this, we proposed DART, a plug-in framework comprising a Residual Disentanglement Module (RDM) and Cross-Domain Meta-Training (CDMT) that decomposes residuals into domain-agnostic anomaly components and domain-specific components. Extensive experiments across three host methods, three benchmarks, and three few-shot settings demonstrated that DART consistently improves cross-domain detection performance, with gains scaling with the severity of the domain shift (industrial-to-medical transfer). While adding 3--20M training-time parameters (0.8--9.1M at inference), DART delivers gains over vanilla (up to +1.82\% I-AUROC on industrial benchmarks and +13.28\% on medical transfer), whereas naive abnormal integration (+Inv) often degrades performance (e.g., $-$6.5\% I-AUROC for ResAD on VisA), making DART's net gain over +Inv as high as +9.85\%. Ablation studies confirmed that all components are jointly necessary and that disentanglement must operate in residual space. DART assumes at least one abnormal reference at test time, which zero-shot deployments may lack, and our evaluation is restricted to images; extending residual disentanglement to video and time-series anomalies is left to future work.



\clearpage


\begin{ack}
This work was supported by Fondazione Cariverona through the projects ``SmartSpindle5.0'' (Bando Ricerca e Sviluppo 2024/2025, Ref.~2025.0053, ID~54833, CUP~B33C25002220007) and ``FINAL LAP---FINALizzazione intelligente dei materiali LAPidei derivanti da taglio'' (Bando Ricerca e Sviluppo 2024). The authors declare no competing interests.
\end{ack}

{
\small
\bibliographystyle{ieeetr}
\bibliography{reference}
}

\newpage
\section*{NeurIPS Paper Checklist}

\begin{enumerate}

\item {\bf Claims}
    \item[] Question: Do the main claims made in the abstract and introduction accurately reflect the paper's contributions and scope?
    \item[] Answer: \answerYes{}
    \item[] Justification: The claims in the Abstract and Section~\ref{sec:introduction} are supported by the results in Table~\ref{tab:mvtec_visa}, Table~\ref{tab:brats} and the ablations in Section~\ref{sec:ablation}; the limitations are stated in the Conclusion.

\item {\bf Limitations}
    \item[] Question: Does the paper discuss the limitations of the work performed by the authors?
    \item[] Answer: \answerYes{}
    \item[] Justification: Limitations are stated at the end of the Conclusion (requirement of at least one abnormal reference at test time; image-only evaluation) and discussed further in Appendix~\ref{sec:supp:limit}.

\item {\bf Theory assumptions and proofs}
    \item[] Question: For each theoretical result, does the paper provide the full set of assumptions and a complete (and correct) proof?
    \item[] Answer: \answerNA{}
    \item[] Justification: This paper does not include theoretical results.

    \item {\bf Experimental result reproducibility}
    \item[] Question: Does the paper fully disclose all the information needed to reproduce the main experimental results of the paper to the extent that it affects the main claims and/or conclusions of the paper (regardless of whether the code and data are provided or not)?
    \item[] Answer: \answerYes{}
    \item[] Justification: Protocol, host methods, and the DART configuration are described in Section~\ref{sec:exp}; architectures, hyperparameters, the CDMT curriculum, reference-selection and leakage controls are given in Appendices~\ref{sec:implementation_appendix}--\ref{sec:supp:leakage}.

\item {\bf Open access to data and code}
    \item[] Question: Does the paper provide open access to the data and code, with sufficient instructions to faithfully reproduce the main experimental results, as described in supplemental material?
    \item[] Answer: \answerYes{}
    \item[] Justification: Code is made available at \url{https://github.com/aqeeelmirza/DART} (linked in the Abstract), with instructions to reproduce the main results. MVTec AD, VisA, and BraTS are public benchmarks, cited in Section~\ref{sec:exp}.

\item {\bf Experimental setting/details}
    \item[] Question: Does the paper specify all the training and test details (e.g., data splits, hyperparameters, how they were chosen, type of optimizer) necessary to understand the results?
    \item[] Answer: \answerYes{}
    \item[] Justification: Please see Section~\ref{sec:exp} and Appendix~\ref{sec:implementation_appendix} (optimizer, schedule, loss weights, curriculum, validation and checkpoint selection on source domains only).

\item {\bf Experiment statistical significance}
    \item[] Question: Does the paper report error bars suitably and correctly defined or other appropriate information about the statistical significance of the experiments?
    \item[] Answer: \answerYes{}
    \item[] Justification: NAGL and NAGL+DART results are averaged over three independent seeds controlling the few-shot reference sampling, with 1-sigma standard deviations below 0.8 I-AUROC and 0.5 P-AUROC in all settings; the reported gains exceed these margins. InCTRL and ResAD follow the single-run protocol of their original papers. See Appendix~\ref{sec:supp:stdev}.

\item {\bf Experiments compute resources}
    \item[] Question: For each experiment, does the paper provide sufficient information on the computer resources (type of compute workers, memory, time of execution) needed to reproduce the experiments?
    \item[] Answer: \answerYes{}
    \item[] Justification: Compute is reported in the Efficiency paragraph of the main text and in the Efficiency Analysis paragraph and Table~\ref{tab:supp:efficiency} of Appendix~\ref{sec:implementation_appendix}: a single NVIDIA RTX4090 (24\,GB), ${\approx}$1 GPU-hour per setting, and 100--150 GPU-hours in total.
    
\item {\bf Code of ethics}
    \item[] Question: Does the research conducted in the paper conform, in every respect, with the NeurIPS Code of Ethics \url{https://neurips.cc/public/EthicsGuidelines}?
    \item[] Answer: \answerYes{}
    \item[] Justification: We have read the NeurIPS Code of Ethics and confirm that our research complies with these ethical standards.

\item {\bf Broader impacts}
    \item[] Question: Does the paper discuss both potential positive societal impacts and negative societal impacts of the work performed?
    \item[] Answer: \answerYes{}
    \item[] Justification: Please see the Social Impacts paragraph in Appendix~\ref{sec:supp:limit}.
    
\item {\bf Safeguards}
    \item[] Question: Does the paper describe safeguards that have been put in place for responsible release of data or models that have a high risk for misuse (e.g., pre-trained language models, image generators, or scraped datasets)?
    \item[] Answer: \answerNA{}
    \item[] Justification: The released models are anomaly detectors for industrial and medical benchmark images and pose no such risks.

\item {\bf Licenses for existing assets}
    \item[] Question: Are the creators or original owners of assets (e.g., code, data, models), used in the paper, properly credited and are the license and terms of use explicitly mentioned and properly respected?
    \item[] Answer: \answerYes{}
    \item[] Justification: All datasets (MVTec AD, VisA, BraTS) and host methods (InCTRL, ResAD, NAGL) are publicly released by their authors under their respective licenses and are cited in the paper; we use them unmodified for research purposes.

\item {\bf New assets}
    \item[] Question: Are new assets introduced in the paper well documented and is the documentation provided alongside the assets?
    \item[] Answer: \answerYes{}
    \item[] Justification: The released code is documented in the repository (\url{https://github.com/aqeeelmirza/DART}): training and evaluation instructions, the evaluation protocol, and an MIT license.

\item {\bf Crowdsourcing and research with human subjects}
    \item[] Question: For crowdsourcing experiments and research with human subjects, does the paper include the full text of instructions given to participants and screenshots, if applicable, as well as details about compensation (if any)? 
    \item[] Answer: \answerNA{}
    \item[] Justification: Our paper does not involve crowdsourcing nor research with human subjects.

\item {\bf Institutional review board (IRB) approvals or equivalent for research with human subjects}
    \item[] Question: Does the paper describe potential risks incurred by study participants, whether such risks were disclosed to the subjects, and whether Institutional Review Board (IRB) approvals (or an equivalent approval/review based on the requirements of your country or institution) were obtained?
    \item[] Answer: \answerNA{}
    \item[] Justification: Our paper does not involve crowdsourcing nor research with human subjects.

\item {\bf Declaration of LLM usage}
    \item[] Question: Does the paper describe the usage of LLMs if it is an important, original, or non-standard component of the core methods in this research? Note that if the LLM is used only for writing, editing, or formatting purposes and does \emph{not} impact the core methodology, scientific rigor, or originality of the research, declaration is not required.
    \item[] Answer: \answerNA{}
    \item[] Justification: LLMs were used only for language editing; they are not a component of the method, experiments, or analysis.

\end{enumerate}

\newpage
\appendix
\section{Supplementary Material}
\label{sec:supp}

\subsection{Implementation Details}
\label{sec:implementation_appendix}

\paragraph{Host methods.}
We reproduce each host method following its original protocol to ensure that observed gains are attributable to DART rather than implementation differences.
\textit{InCTRL}~\cite{zhu2024toward} uses a frozen OpenCLIP ViT-B/16-plus-240 backbone at $240{\times}240$ resolution, computing multi-layer patch-level cosine distance maps and image-level residual vectors between query and normal prompts. The learnable adapter $\psi$ and classifier $\eta$ are trained with binary focal loss while the CLIP backbone remains frozen. Full architectural details are provided in \S\ref{sec:supp:inctrl}.
\textit{ResAD}~\cite{yao2024resad} employs a frozen WideResNet-50 with multi-layer feature aggregation. A Feature Constraintor is trained with the log-barrier one-class loss, a normalizing-flow Feature Distribution Estimator is trained with the negative log-likelihood objective, and a VQ codebook is constructed during training and frozen at test time. Full details in \S\ref{sec:supp:resad}.
\textit{NAGL}~\cite{wangnormal} uses a frozen DINOv2 ViT-S/14 backbone (21M parameters) with $M{=}25$ learnable proxies and trains the Residual Mining (RM) and Anomaly Feature Learning (AFL) attention modules with AdamW (lr $1{\times}10^{-5}$, decayed by $0.1$ at epochs 10 and 15) for 20 epochs of 500 episodes. Full details in \S\ref{sec:supp:nagl}.

Across all three host methods, input resolution follows the original paper (no additional augmentation), and the same set of normal and abnormal references is shared per few-shot setting $N^k{+}A^1$ (with $A^1$ throughout, and $k \in \{1,2,4\}$). NAGL results are averaged over three independent seeds, while InCTRL and ResAD follow the single-run protocol of their original papers (see \S\ref{sec:supp:stdev}).

\paragraph{DART hyperparameters.}
The Residual Disentanglement Module (RDM) consists of:
\begin{itemize}
    \item Anomaly encoder $E_a$: single transformer block (multi-head self-attention + FFN + LayerNorm), hidden dimension matching the host method's residual dimension.
    \item Domain encoder $E_d$: two-layer MLP with GELU activation and LayerNorm.
    \item Reconstruction decoder $G$: two-layer MLP taking concatenated $[\mathbf{r}_a; \mathbf{r}_d]$ as input.
    \item Domain classifier $C_d$: two-layer MLP with spectral normalization~\cite{miyato2018spectral}, preceded by a Gradient Reversal Layer (GRL)~\cite{ganin2016domain}.
\end{itemize}
Loss weights are set to $\lambda_{\mathrm{adv}} = 0.1$ and $\lambda_{\mathrm{recon}} = 0.5$ for all experiments. RDM is optimized with AdamW (lr $5{\times}10^{-5}$, weight decay $10^{-4}$) using a cosine learning rate schedule over 30 epochs, with batch size 16 and gradient clipping at norm 1.0. The host method's parameters remain entirely frozen during DART training.

\paragraph{Cross-Domain Meta-Training (CDMT) protocol.}
For each evaluation target $\mathcal{D}_T$, the entire target dataset is held out and CDMT episodes are sampled exclusively from the complementary source pool $\mathcal{D}_S$ (this protocol is summarized in Sec.~4 of the main text). Within each episode, with probability $p_{\mathrm{cross}}(t)$ the reference and query are sampled from \textit{different} source categories. Both $p_{\mathrm{cross}}$ and the GRL strength $\alpha(t)$ follow a two-phase curriculum:
\begin{itemize}
    \item \textbf{Phase~1 - Warmup} (epochs $1$ to $T_w{=}5$): all episodes are same-domain ($p_{\mathrm{cross}}{=}0$) and adversarial training is disabled ($\alpha{=}0$). Only $\mathcal{L}_{\mathrm{suf}} + \lambda_{\mathrm{recon}}\mathcal{L}_{\mathrm{recon}}$ are active, allowing the model to learn a stable residual decomposition.
    \item \textbf{Phase~2 - Ramp-up} (epochs $T_w{+}1$ to $T{=}30$): $p_{\mathrm{cross}}$ ramps linearly from 0 to $p_{\max}{=}0.3$, and $\alpha(t)$ follows the DANN schedule (Eq.~\ref{eq:grl_schedule} in the main text). The full objective $\mathcal{L}_{\mathrm{total}} = \mathcal{L}_{\mathrm{suf}} + \lambda_{\mathrm{adv}}\mathcal{L}_{\mathrm{adv}} + \lambda_{\mathrm{recon}}\mathcal{L}_{\mathrm{recon}}$ is active.
\end{itemize}

\paragraph{Domain label in adversarial training.}
The domain label $d$ used in $\mathcal{L}_{\mathrm{adv}}$ corresponds to the \textit{reference domain} $D_i$ from which the normal--abnormal pair is drawn. This is the domain whose identity we want to suppress in $\mathbf{r}_a$: the residual is computed from references in $D_i$, so any domain-specific information in $\mathbf{r}_a$ originates from $D_i$. The query domain $D_j$ does not enter the adversarial loss.

\paragraph{Adapting normal-only baselines for abnormal references.}
InCTRL and ResAD are originally designed for normal-only references. We adapt them for the $N^k{+}A^1$ setting as follows:
\begin{itemize}
    \item \textbf{InCTRL+DART:} The abnormal reference is passed through the frozen CLIP visual encoder alongside normal prompts. The residual between abnormal and normal token embeddings at each layer is computed analogously to the query-normal residual (Eq.~\ref{eq:supp:inctrl_patch}), producing an abnormal-normal residual tensor. This tensor is fed into RDM, which outputs $\mathbf{r}_a$. The disentangled residual is then fused with the existing query-normal residual via concatenation before the scoring head.
    \item \textbf{ResAD+DART:} The abnormal reference features are extracted from the frozen WideResNet backbone. Per-patch residuals between abnormal and normal features are computed via nearest-neighbor subtraction (Eq.~\ref{eq:supp:resad_residual}). These residuals are fed into RDM to produce $\mathbf{r}_a$, which is concatenated with the original query-normal residual features before the Feature Constraintor and normalizing flow.
    \item \textbf{+Inv baseline:} The same abnormal reference integration is performed, but without RDM, the raw abnormal-normal residual is concatenated directly. This isolates the contribution of disentanglement from the benefit of abnormal access.
\end{itemize}

\paragraph{Validation and model selection.}
Best checkpoints are selected by mean image-level AUROC on a held-out subset of source categories. No target-domain category, image, reference, label, or statistic is used during training, validation, hyperparameter tuning, or checkpoint selection.

\paragraph{Inference.}
At inference, the target domain $D_k \in \mathcal{D}_T$ is accessed for the first time. For each test query $q$: (i)~a few-shot support set is drawn from $D_k$; (ii)~the raw residual $\mathbf{r}$ is computed using the host method's pipeline; (iii)~$\mathbf{r}_a = E_a(\mathbf{r})$ is extracted; (iv)~the frozen detection head produces the anomaly score and segmentation map from $\mathbf{r}_a$. The domain encoder $E_d$, decoder $G$, and classifier $C_d$ are discarded. No parameters are updated on $\mathcal{D}_T$.


\paragraph{Efficiency Analysis.} Tab.~\ref{tab:supp:efficiency} summarizes the efficiency of each host method and the overhead introduced by DART, distinguishing training-time from inference-time cost. The full RDM adds 3.1M, 4.2M, and 20.1M parameters for InCTRL, NAGL, and ResAD respectively during training; at inference, only the anomaly encoder $E_a$ is retained (0.82M, 0.89M, and 9.06M), while the domain encoder, reconstruction decoder, and domain classifier are training-only components discarded after learning the residual decomposition. Training requires approximately 1 GPU-hour per setting (30 epochs on a single NVIDIA RTX4090 GPU) since the host backbone is fully frozen, and this cost is incurred only once: the learned module transfers to unseen target domains without any further optimization. Inference overhead is negligible as the RDM operates only on low-dimensional residual features. All results are reported for 1, 2 and 4 shot settings; total compute for all experiments (3 host methods $\times$ 3 datasets $\times$ 3 shot settings, with 3 seeds for NAGL) is approximately 100-150 GPU-hours on NVIDIA RTX4090 GPUs (24\,GB VRAM).

\begin{table}[h]
\centering
\caption{Efficiency comparison. DART overhead is split into training-time parameters (full RDM) and inference-time parameters (only $E_a$ retained); DART training time is ${\approx}$1 GPU-hour per setting for all hosts.}
\label{tab:supp:efficiency}
\setlength{\tabcolsep}{4pt}
\small
\begin{tabular}{l ccc cc}
\toprule
& \multicolumn{3}{c}{Host} & \multicolumn{2}{c}{DART Overhead (M)} \\
\cmidrule(lr){2-4}\cmidrule(lr){5-6}
Method & Params (M) & Train (H) & FPS & Train & Inference \\
\midrule
InCTRL~\cite{zhu2024toward}  & 117.5 & 0.7  & 1.2  & +3.1  & +0.82 \\
ResAD~\cite{yao2024resad}    & 59.2  & 20.6 & 7.8  & +20.1 & +9.06 \\
NAGL~\cite{wangnormal}       & 24.4  & 0.3  & 17.1 & +4.2  & +0.89 \\
\bottomrule
\end{tabular}
\end{table}

\subsection{Leakage Controls}
\label{sec:supp:leakage}

A central concern in cross-domain anomaly detection is information leakage between training and evaluation. We enforce three levels of isolation, all derived from the leave-one-dataset-out design.

\noindent\textbf{Category-level isolation.} The evaluation category never appears as a source domain during training, since the entire dataset it belongs to is excluded from $\mathcal{D}_S$.

\noindent\textbf{Dataset-level isolation.} Categories within the same dataset share acquisition pipeline, lighting, sensor resolution, and annotation conventions. Excluding individual test categories while retaining other categories from the same dataset would leak domain-specific appearance statistics into training. Our leave-one-dataset-out construction eliminates this.

\noindent\textbf{Support-set isolation.} The few-shot reference images used at test time are drawn from the standard evaluation split of $D_k$ and are not observed during training under any sampling scheme.

Together, these conditions guarantee that performance on $\mathcal{D}_T$ reflects genuine generalization from $\mathcal{D}_S$.

\subsection{Statistical Significance}
\label{sec:supp:stdev}

NAGL and NAGL+DART results are averaged over three independent seeds controlling the few-shot reference sampling, with standard deviations consistently below 0.8\% for I-AUROC and 0.5\% for P-AUROC across all datasets and shot settings; the observed DART improvements (e.g., +0.39\% I-AUROC on MVTec at $N^1$, +1.82\% on VisA at $N^2$) exceed these margins by a substantial factor. InCTRL and ResAD follow the single-run evaluation protocol of their original papers~\cite{zhu2024toward,yao2024resad}; for these backbones, robustness of DART gains is evidenced by their consistency across all shot settings and the majority of per-category results (Tables~\ref{tab:supp_inctrl_1shot}--\ref{tab:supp_nagl_visa_4shot}).


\subsection{Host Method Details}

\subsubsection{InCTRL}
\label{sec:supp:inctrl}

InCTRL~\cite{zhu2024toward} adapts CLIP to the few-shot anomaly detection setting by modeling the \textit{in-context residual} between a query image and normal sample prompts. Given a query image $x$ and $K$ normal prompts $P = \{p_1, \dots, p_K\}$, the frozen CLIP visual encoder $f_v(\cdot)$ produces patch token maps $\{T^l_x\}_{l=1}^n$ at each of its $n$ transformer blocks. A patch-level residual map is computed at each layer:
\begin{equation}
    M^l_x(i,j) = 1 - \langle T^l_x(i,j),\, h(T^l_x(i,j)\,|\,P)\rangle, \quad M_x = \frac{1}{n}\sum_{l=1}^{n} M^l_x
    \label{eq:supp:inctrl_patch}
\end{equation}
where $h(\cdot|P)$ retrieves the most similar patch embedding among all prompts. An image-level residual is obtained via a learnable adapter $\psi$: $F_x = \psi(f_v(x)) - \frac{1}{K}\sum_{k} \psi(f_v(p_k))$. A classifier $\eta$ scores this as $s_i(x) = \eta(F_x)$. A text-based score $s_a(x)$ is computed from CLIP text-image similarity. The final anomaly score combines all signals:
\begin{equation}
    s(x) = \phi\big(M_x \oplus s_i(x) \oplus s_a(x);\,\Theta_\phi\big) + \alpha \cdot \max(M_x)
\end{equation}

\subsubsection{ResAD}
\label{sec:supp:resad}

ResAD~\cite{yao2024resad} computes residual features by subtracting the nearest normal reference for each patch, making the representation class-invariant. For each input image $I_i$, a frozen WideResNet-50 produces multi-layer feature maps $\phi^l(I_i) \in \mathbb{R}^{H_l \times W_l \times C_l}$. The per-patch residual is:
\begin{equation}
    x^{l,r}_{h,w} = x^l_{h,w} - x^*_n, \qquad x^*_n = \arg\min_{x \in P_l}\|x - x^l_{h,w}\|_2
    \label{eq:supp:resad_residual}
\end{equation}
A Feature Constraintor $\psi(\cdot;\mathcal{W})$ projects normal residuals onto a hypersphere via a log-barrier one-class loss:
\begin{equation}
    \mathcal{L}_{\text{occ}} = -\frac{1}{N}\sum_{i=1}^{N} \log\sigma(-s_i)\cdot e^{s_i} + \frac{\lambda}{2}\sum_{k=1}^{K}\|\mathcal{W}_k\|_F^2
\end{equation}
where $s_i = \|\psi(x_i;\mathcal{W})\|_2 - R$. A normalizing-flow estimator models the constrained feature distribution; anomalies are flagged as low-likelihood outliers. A VQ codebook mitigates train-test distribution mismatch.

\subsubsection{NAGL}
\label{sec:supp:nagl}

NAGL~\cite{wangnormal} operates in the normal-abnormal guided setting, using residuals between abnormal and normal references as a transferable anomaly representation. Given query features $F^q$, normal features $F^n$, and abnormal features $F^a$ from a frozen ViT backbone, NAGL first computes a normal-guided score via nearest-neighbor patch matching. The Residual Mining (RM) module computes the abnormal--normal residual $\text{Res}(F^a, F^n) = F^a - F^n_*$ and uses learnable proxies $P$ with cross-attention and masked self-attention to produce residual proxies:
\begin{equation}
    \widetilde{P} = \mathrm{SA}\!\left(\mathrm{Softmax}\!\left(\tfrac{QK^\top}{\sqrt{d}} + M'\right) V\right), \quad V = W^V \,\text{Res}(F^a, F^n)
\end{equation}
The Anomaly Feature Learning (AFL) module transfers these proxies onto the query via cross-attention, producing anomaly proxies $\widehat{P}$ that highlight abnormal regions. The final score fuses normal-guided and abnormal-guided maps. Training uses focal + dice loss for segmentation and BCE for classification.

\subsection{Clarification: Which Residual Does RDM Operate On?}
\label{sec:supp:residual_clarification}

A potential source of confusion arises from the role of the abnormal reference $\mathcal{F}_a$ in residual computation. We clarify this here.

\paragraph{The residual $\mathbf{r}$.} As shown in Fig.~\ref{fig:pipeline}, the host residual operator computes
\begin{equation}
    \mathbf{r} = \mathcal{R}(\mathbf{f}_q;\, \mathcal{F}_n,\, \mathcal{F}_a),
\end{equation}
where $\mathbf{f}_q$ is the query feature and $\{\mathcal{F}_n, \mathcal{F}_a\}$  are the normal and abnormal reference features. This single residual $\mathbf{r}$ is the input to RDM. The exact form of $\mathcal{R}$ is method-specific: NAGL and ResAD compute the abnormal-normal residual via patch-level nearest-neighbour subtraction of $\mathcal{F}_a$ against $\mathcal{F}_n$, while InCTRL computes it as a patch-level cosine distance map and an image-level residual vector across transformer layers. In all cases, $\mathbf{r}$ encodes the deviation of the abnormal reference from the normal reference, and RDM receives this as input regardless of how the host method produces it, making DART a true plug-in framework.

\paragraph{The role of $\mathcal{F}_a$ in $\mathbf{r}$.} Within $\mathcal{R}$, $\mathcal{F}_n$ provides the normal baseline and $\mathcal{F}_a$ acts as a directional anchor, it encodes \emph{what anomaly looks like} in the reference domain. Together, they orient $\mathbf{r}$ so that its magnitude and direction reflect how the query deviates from normal \emph{in the direction of the reference anomaly}. This is precisely why $\mathbf{r}$ carries domain entanglement: both the transferable anomaly pattern and the domain-specific texture of $\mathcal{F}_a$ are embedded in $\mathbf{r}$.

\paragraph{What RDM does.} RDM receives $\mathbf{r}$ and decomposes it into a domain-invariant anomaly 
component $\mathbf{r}_a$ and a domain-specific component $\mathbf{r}_d$, such that $G([\mathbf{r}_a;\mathbf{r}_d]) \approx \mathbf{r}$. Only $\mathbf{r}_a$ is passed to the host detection head, discarding the domain texture that would otherwise cause false activations on unseen domains. This decomposition is the core contribution of DART and is what enables consistent cross-domain gains across all three host methods.

The information-theoretic objective $\max\; I(\mathbf{r}_a;\, Y)$ is meaningful because the anomaly label $Y$ of the query $\mathbf{x}_q$ varies across episodes, and $\mathbf{r}_a$ must retain sufficient anomaly information from the reference to correctly classify the query, while shedding the domain-specific signal in $\mathbf{r}_d$.

\subsection{Discussion}
\label{sec:supp:limit}
\paragraph{Limitations.} 
DART requires at least one abnormal reference sample at test time, which may not always be available in zero-shot scenarios; our experiments also focus solely on image data, and extension to video or time-series modalities remains future work.

\paragraph{Social Impacts.}
As a plug-in framework for generalist anomaly detection, DART does not raise specific ethical issues or adverse societal effects. All datasets used are publicly available. Qualitative illustrations are derived from industrial product imagery and medical imaging benchmarks, ensuring no violation of personal privacy.


\subsection{VisA Domain Entanglement Analysis}
\label{sec:supp:visa}

\begin{figure}[t]
    \centering
    \begin{subfigure}[b]{0.48\textwidth}
        \includegraphics[width=\textwidth]{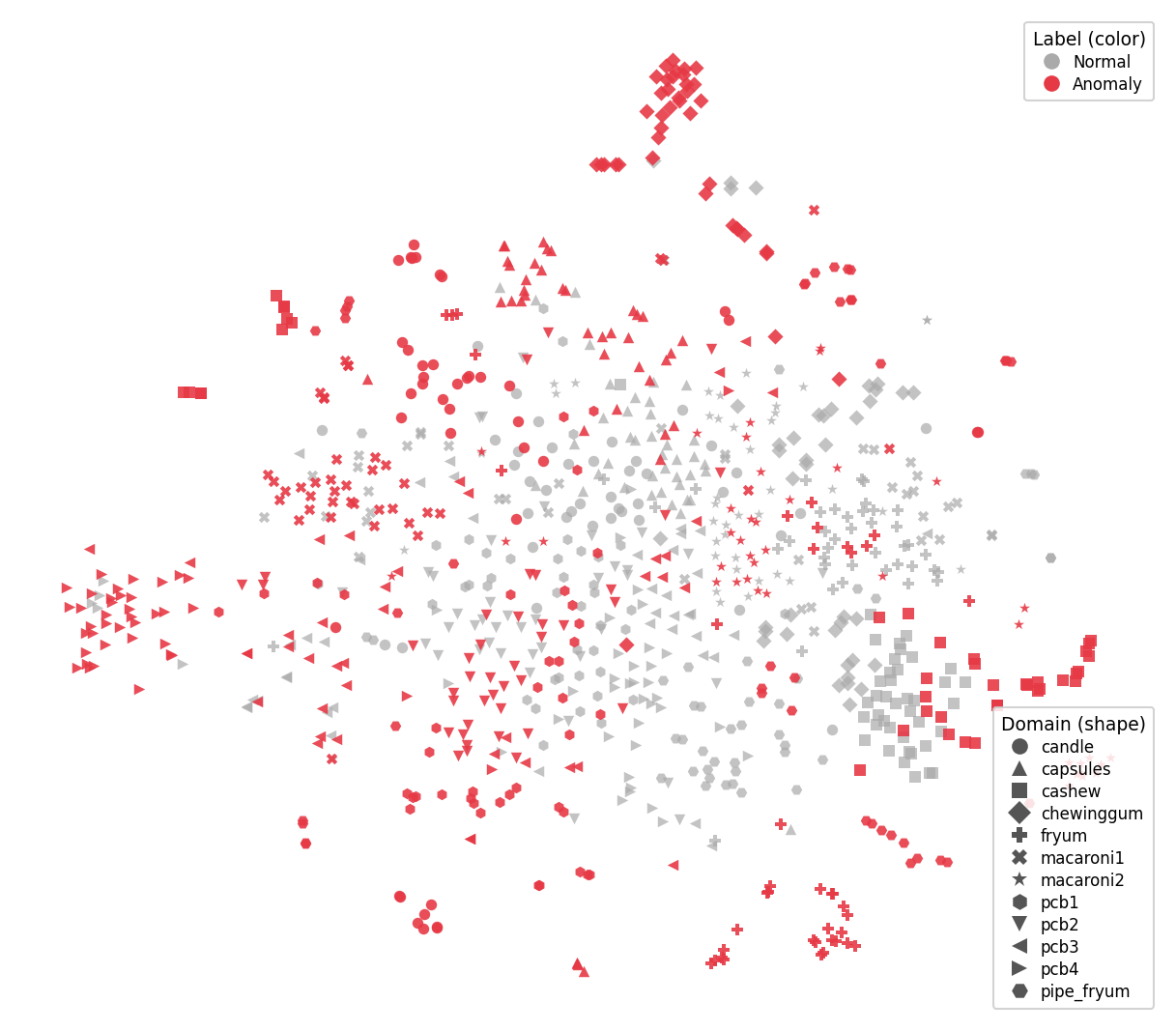}
        \caption{Without DART: clusters by domain identity}
        \label{fig:supp:tsne_vanilla}
    \end{subfigure}
    \hfill
    \begin{subfigure}[b]{0.48\textwidth}
        \includegraphics[width=\textwidth]{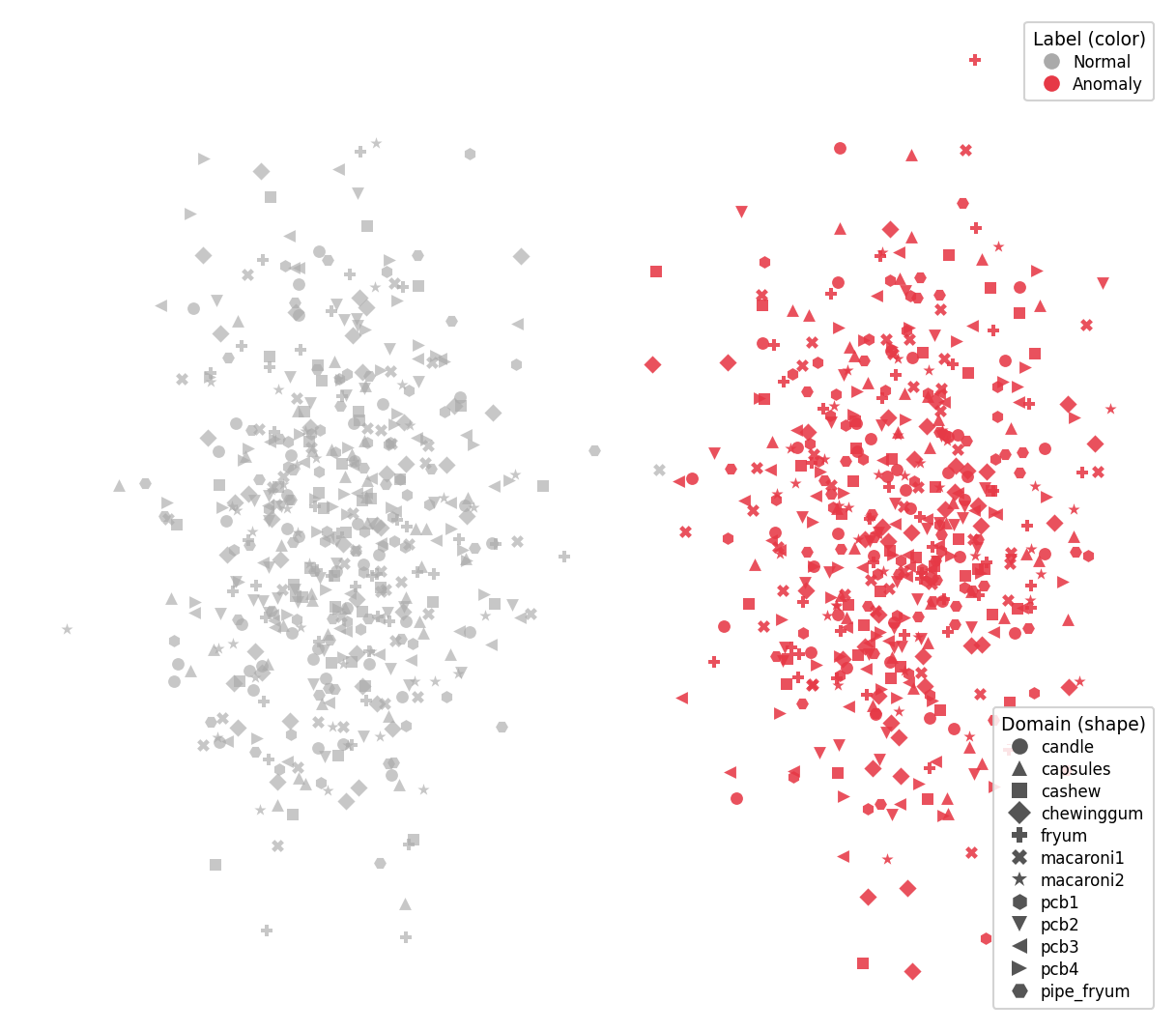}
        \caption{With DART: clusters by anomaly type}
        \label{fig:supp:tsne_dart}
    \end{subfigure}
    \caption{t-SNE visualization of token-level residuals on VisA at $k{=}2$. 
    \textbf{(a)}~Without DART, each category occupies a distinct region, indicating domain texture dominates. 
    \textbf{(b)}~With DART, defects of the same semantic type cluster together across domains.}
    \label{fig:visa_entanglement}
\end{figure}

To empirically validate that residual domain entanglement is the core bottleneck addressed by DART, we visualize the token-level residual space using t-SNE on VisA at $k{=}2$ shots (Figure~\ref{fig:visa_entanglement}). Without DART, residuals form tight clusters organized by \textit{domain identity} rather than anomaly semantics, confirming that the normal--abnormal difference vector carries substantial domain texture information. With DART, RDM strips the domain-specific component: residuals reorganize into clusters aligned with \textit{defect type}, grouping semantically similar anomalies across object categories. VisA exhibits particularly strong domain entanglement due to its high intra-class visual variance (structurally complex objects such as PCBs and food items), directly explaining why DART yields larger performance gains on VisA than on MVTec AD.

\subsection{Disentanglement Analysis: Vanilla vs.\ +Inv vs.\ +DART}
\label{sec:supp:disentanglement}

Table~\ref{tab:supp:disent} provides the full vanilla/+Inv/+DART breakdown for all three host methods on MVTec AD and VisA across all shot settings ($N^1$ through $N^4$), extending the mean-level +Inv comparison now included in Table~\ref{tab:mvtec_visa} of the main paper with all twelve metrics and per-metric directional markers.

\paragraph{Naive abnormal integration (+Inv) is consistently harmful.}
Across all three host methods, the +Inv baseline, which incorporates abnormal references without disentanglement, degrades performance relative to vanilla in the majority of settings. The damage is most severe for ResAD, where +Inv drops I-AUROC by $-$3.0\% on MVTec AD and $-$6.5\% on VisA at $N^1$, with pixel-level PRO degrading by up to $-$6.0 points. For InCTRL, +Inv is inconsistent: it provides marginal image-level gains at low shots ($N^1$: +0.42\%, $N^2$: +0.67\%) but degrades at higher shots ($N^4$: $-$1.22\%). Crucially, +Inv never produces pixel-level metrics for InCTRL, the entangled abnormal residuals generate spatially incoherent activations that prevent meaningful localization. These results confirm our core hypothesis: raw abnormal references carry domain-entangled information that corrupts rather than aids cross-domain detection.

\paragraph{DART consistently recovers and exceeds vanilla performance.}
In contrast to +Inv, DART improves over vanilla across all three host methods in the large majority of metrics and settings. On InCTRL, DART achieves +0.56\% to +1.15\% I-AUROC gains on MVTec AD across all shots, while simultaneously enabling pixel-level localization (P-AUROC 94.52--95.71\%) that is entirely absent in both vanilla and +Inv. On ResAD, DART recovers the performance destroyed by +Inv and exceeds vanilla (e.g., +0.50\% I-AUROC, +1.80\% I-AP on VisA at $N^2$). On NAGL, which already handles abnormal references natively, DART provides consistent image-level gains on both datasets (up to +1.82\% I-AUROC on VisA at $N^2$), with pixel-level metrics remaining competitive or improving.

\paragraph{The gap between +Inv and +DART widens with domain shift.}
On MVTec AD (where the VisA-trained model transfers to a moderately different industrial domain), the +Inv$\to$+DART gap is moderate. On VisA (where the MVTec-trained model faces a larger domain shift), the gap widens substantially: for example, InCTRL+Inv drops to 79.87\% I-AUROC at $N^1$ while InCTRL+DART reaches 86.06\%, a gap of +6.19 points. This scaling behavior directly validates that residual disentanglement becomes increasingly critical as the domain gap grows.


\begin{table*}[t]
\centering
\caption{Full disentanglement analysis on MVTec AD and VisA. Three variants per host method: vanilla (normal only), +Inv (naive abnormal addition without RDM), +DART (with RDM). {\color{green!60!black}$\uparrow$}/{\color{red}$\downarrow$}: vs.\ vanilla.}
\label{tab:supp:disent}
\resizebox{\textwidth}{!}{%
\begin{tabular}{ll cccccc cccccc}
\toprule
& & \multicolumn{6}{c}{\textbf{MVTec AD}} & \multicolumn{6}{c}{\textbf{VisA}} \\
\cmidrule(lr){3-8} \cmidrule(lr){9-14}
& & \multicolumn{3}{c}{Image-level} & \multicolumn{3}{c}{Pixel-level} & \multicolumn{3}{c}{Image-level} & \multicolumn{3}{c}{Pixel-level} \\
\cmidrule(lr){3-5} \cmidrule(lr){6-8} \cmidrule(lr){9-11} \cmidrule(lr){12-14}
Setting & Method & AUROC & AP & F1 & AUROC & PRO & F1 & AUROC & AP & F1 & AUROC & PRO & F1 \\
\midrule
\multirow{9}{*}{$N^1{+}A^1$}
 & NAGL        & 95.54 & 97.63 & 95.40 & 96.53 & 92.66 & 58.03 & 86.23 & 86.54 & 84.59 & 97.20 & 88.76 & 42.41 \\
 & NAGL+DART   & 95.93 {\color{green!60!black}$\uparrow$} & 97.95 {\color{green!60!black}$\uparrow$} & 95.76 {\color{green!60!black}$\uparrow$} & 96.51 {\color{red}$\downarrow$} & 92.88 {\color{green!60!black}$\uparrow$} & 58.37 {\color{green!60!black}$\uparrow$} & 87.65 {\color{green!60!black}$\uparrow$} & 88.00 {\color{green!60!black}$\uparrow$} & 84.25 {\color{red}$\downarrow$} & 97.21 {\color{green!60!black}$\uparrow$} & 88.77 {\color{green!60!black}$\uparrow$} & 41.25 {\color{red}$\downarrow$} \\
\cmidrule(lr){2-14}
 & InCTRL      & 94.99 & 97.33 & 95.07 & -- & -- & -- & 85.47 & 55.65 & 60.04 & -- & -- & -- \\
 & InCTRL+Inv  & 95.41 & 97.72 & 94.97 & -- & -- & -- & 79.87 & 55.37 & 57.95 & -- & -- & -- \\
 & InCTRL+DART & 95.55 {\color{green!60!black}$\uparrow$} & 97.74 {\color{green!60!black}$\uparrow$} & 95.21 {\color{green!60!black}$\uparrow$} & 94.52 & 85.23 & 51.16 & 86.06 {\color{green!60!black}$\uparrow$} & 60.10 {\color{green!60!black}$\uparrow$} & 63.31 {\color{green!60!black}$\uparrow$} & 95.44 & 69.15 & 27.24 \\
\cmidrule(lr){2-14}
 & ResAD       & 85.70 & 93.10 & 90.90 & 94.40 & 84.50 & 49.80 & 81.20 & 82.50 & 81.20 & 95.20 & 78.80 & 33.30 \\
 & ResAD+Inv   & 82.70 & 91.10 & 89.90 & 93.20 & 81.00 & 46.60 & 74.70 & 76.70 & 78.20 & 94.00 & 72.80 & 25.30 \\
 & ResAD+DART  & 85.90 {\color{green!60!black}$\uparrow$} & 93.10 & 90.30 {\color{red}$\downarrow$} & 94.10 {\color{red}$\downarrow$} & 84.00 {\color{red}$\downarrow$} & 50.30 {\color{green!60!black}$\uparrow$} & 81.70 {\color{green!60!black}$\uparrow$} & 84.30 {\color{green!60!black}$\uparrow$} & 81.50 {\color{green!60!black}$\uparrow$} & 95.40 {\color{green!60!black}$\uparrow$} & 78.20 {\color{red}$\downarrow$} & 28.50 {\color{red}$\downarrow$} \\
\midrule
\multirow{9}{*}{$N^2{+}A^1$}
 & NAGL        & 96.61 & 98.11 & 96.10 & 96.63 & 93.13 & 59.20 & 87.90 & 89.03 & 85.15 & 97.34 & 89.25 & 44.06 \\
 & NAGL+DART   & 96.62 {\color{green!60!black}$\uparrow$} & 98.15 {\color{green!60!black}$\uparrow$} & 96.20 {\color{green!60!black}$\uparrow$} & 96.64 {\color{green!60!black}$\uparrow$} & 93.13 & 59.22 {\color{green!60!black}$\uparrow$} & 89.72 {\color{green!60!black}$\uparrow$} & 89.16 {\color{green!60!black}$\uparrow$} & 86.21 {\color{green!60!black}$\uparrow$} & 97.29 {\color{red}$\downarrow$} & 89.07 {\color{red}$\downarrow$} & 42.65 {\color{red}$\downarrow$} \\
\cmidrule(lr){2-14}
 & InCTRL      & 95.06 & 97.55 & 95.76 & -- & -- & -- & 87.96 & 58.67 & 63.61 & -- & -- & -- \\
 & InCTRL+Inv  & 95.73 & 97.88 & 95.33 & -- & -- & -- & 81.41 & 56.55 & 57.96 & -- & -- & -- \\
 & InCTRL+DART & 96.21 {\color{green!60!black}$\uparrow$} & 98.12 {\color{green!60!black}$\uparrow$} & 95.32 {\color{red}$\downarrow$} & 94.81 & 85.65 & 52.37 & 88.64 {\color{green!60!black}$\uparrow$} & 63.72 {\color{green!60!black}$\uparrow$} & 65.73 {\color{green!60!black}$\uparrow$} & 96.37 & 70.45 & 28.04 \\
\cmidrule(lr){2-14}
 & ResAD       & 87.60 & 94.10 & 91.70 & 95.30 & 86.30 & 51.60 & 83.20 & 84.20 & 82.20 & 95.90 & 80.40 & 37.40 \\
 & ResAD+Inv   & 84.90 & 92.20 & 90.10 & 94.20 & 83.10 & 48.10 & 77.60 & 79.80 & 78.90 & 94.70 & 75.30 & 28.60 \\
 & ResAD+DART  & 88.10 {\color{green!60!black}$\uparrow$} & 94.30 {\color{green!60!black}$\uparrow$} & 91.80 {\color{green!60!black}$\uparrow$} & 95.00 {\color{red}$\downarrow$} & 85.50 {\color{red}$\downarrow$} & 52.10 {\color{green!60!black}$\uparrow$} & 83.40 {\color{green!60!black}$\uparrow$} & 85.60 {\color{green!60!black}$\uparrow$} & 82.40 {\color{green!60!black}$\uparrow$} & 96.20 {\color{green!60!black}$\uparrow$} & 80.80 {\color{green!60!black}$\uparrow$} & 32.40 {\color{red}$\downarrow$} \\
\midrule
\multirow{9}{*}{$N^4{+}A^1$}
 & NAGL        & 97.08 & 98.34 & 96.37 & 96.68 & 93.29 & 59.77 & 90.30 & 90.70 & 87.03 & 97.44 & 89.64 & 44.86 \\
 & NAGL+DART   & 97.34 {\color{green!60!black}$\uparrow$} & 98.46 {\color{green!60!black}$\uparrow$} & 96.52 {\color{green!60!black}$\uparrow$} & 96.76 {\color{green!60!black}$\uparrow$} & 93.39 {\color{green!60!black}$\uparrow$} & 60.03 {\color{green!60!black}$\uparrow$} & 91.29 {\color{green!60!black}$\uparrow$} & 90.94 {\color{green!60!black}$\uparrow$} & 87.87 {\color{green!60!black}$\uparrow$} & 97.45 {\color{green!60!black}$\uparrow$} & 89.99 {\color{green!60!black}$\uparrow$} & 43.56 {\color{red}$\downarrow$} \\
\cmidrule(lr){2-14}
 & InCTRL      & 95.41 & 97.44 & 95.95 & -- & -- & -- & 89.16 & 63.67 & 67.92 & -- & -- & -- \\
 & InCTRL+Inv  & 94.19 & 97.43 & 94.46 & -- & -- & -- & 80.29 & 55.49 & 55.48 & -- & -- & -- \\
 & InCTRL+DART & 95.64 {\color{green!60!black}$\uparrow$} & 97.76 {\color{green!60!black}$\uparrow$} & 96.08 {\color{green!60!black}$\uparrow$} & 95.52 & 86.22 & 53.31 & 90.14 {\color{green!60!black}$\uparrow$} & 69.95 {\color{green!60!black}$\uparrow$} & 70.05 {\color{green!60!black}$\uparrow$} & 96.92 & 71.18 & 29.47 \\
\cmidrule(lr){2-14}
 & ResAD       & 89.90 & 95.10 & 92.50 & 95.90 & 87.50 & 53.40 & 87.00 & 87.90 & 84.60 & 96.40 & 82.80 & 39.50 \\
 & ResAD+Inv   & 87.40 & 93.50 & 91.00 & 94.90 & 84.80 & 49.30 & 82.00 & 83.40 & 81.30 & 95.80 & 79.00 & 31.70 \\
 & ResAD+DART  & 89.90 & 95.10 & 92.50 & 95.60 {\color{red}$\downarrow$} & 86.80 {\color{red}$\downarrow$} & 53.70 {\color{green!60!black}$\uparrow$} & 87.90 {\color{green!60!black}$\uparrow$} & 89.30 {\color{green!60!black}$\uparrow$} & 85.70 {\color{green!60!black}$\uparrow$} & 96.70 {\color{green!60!black}$\uparrow$} & 83.10 {\color{green!60!black}$\uparrow$} & 35.20 {\color{red}$\downarrow$} \\
\bottomrule
\end{tabular}%
}
\end{table*}

\subsection{BraTS Disentanglement Analysis}
\label{sec:supp:brats}

Table~\ref{tab:supp:brats_disent} provides the full vanilla/+Inv/+DART breakdown on the MVTec AD $\to$ BraTS transfer, the most extreme domain shift in our evaluation (industrial defects $\to$ brain tumor MRI).

\paragraph{The industrial-to-medical gap amplifies both the problem and the solution.}
On BraTS, the +Inv baseline is again unreliable: InCTRL+Inv provides partial gains at $N^1$ (+3.77\% I-AUROC) but degrades sharply at $N^4$ ($-$3.58\%), while ResAD+Inv is consistently neutral or harmful ($-$2.20\% at $N^2$, $-$1.90\% at $N^4$). DART, in contrast, produces substantially larger gains than on industrial benchmarks. InCTRL+DART improves by +13.28\% I-AUROC at $N^1$ (65.20\% $\to$ 78.48\%) and +7.61\% at $N^4$ (69.89\% $\to$ 77.50\%), representing the largest absolute improvements across all our experiments. ResAD+DART improves by +5.20\% at $N^2$ and +3.10\% at $N^4$. NAGL+DART improves consistently at $N^2$ (+0.25\%) and $N^4$ (+0.99\%).

\paragraph{Why are BraTS gains so much larger?}
The industrial-to-medical domain shift is qualitatively different from inter-industrial transfer: the visual appearance, texture statistics, imaging modality, and anomaly semantics all change simultaneously. In this regime, domain-specific information in the residual is not merely unhelpful but actively misleading, industrial texture patterns have no correspondence in MRI data and generate pure noise in the anomaly map. Disentanglement removes this noise entirely, leaving only the domain-agnostic anomaly semantics (structural irregularity, boundary disruption, intensity deviation) that transfer across the modality boundary. The magnitude of improvement directly reflects the magnitude of domain entanglement removed.

\paragraph{NAGL+DART at $N^1$ on BraTS.}
The one setting where DART underperforms on BraTS is NAGL+DART at $N^1$, where the mean metric drops from 84.66 to 83.35. This is attributable to a pixel-level AUROC drop (93.00 $\to$ 90.33) while the image-level metric improves slightly (76.32 $\to$ 76.36). We hypothesize that with only a single normal reference, NAGL's RM module, which already performs its own residual processing, has insufficient signal for RDM to learn a stable decomposition on this extreme domain shift. At $N^2$ and above, where additional references stabilize the residual computation, DART consistently improves NAGL on BraTS.

\begin{table*}[t]
\centering
\caption{BraTS disentanglement analysis and notation reference.}
\label{tab:combined}
\setlength{\tabcolsep}{3pt}
\renewcommand{\arraystretch}{1.05}
\small

\begin{subtable}[t]{0.52\linewidth}
\centering
\caption{BraTS disentanglement analysis (MVTec AD $\to$ BraTS). I/P/M: image-/pixel-level AUROC and their mean. {\color{green!60!black}$\uparrow$}/{\color{red}$\downarrow$}: vs.\ vanilla.}
\label{tab:supp:brats_disent}
\begin{tabular}{ll ccc}
\toprule
Setting & Method & I & P & M \\
\midrule
\multirow{8}{*}{$N^1{+}A^1$}
  & NAGL        & 76.32 & 93.00 & 84.66 \\
  & NAGL+DART   & 76.36 {\color{green!60!black}$\uparrow$} & 90.33 {\color{red}$\downarrow$} & 83.35 {\color{red}$\downarrow$} \\
\cmidrule(lr){2-5}
  & InCTRL      & 65.20 & 83.65 & 74.43 \\
  & InCTRL+Inv  & 68.97 & 87.45 & 78.21 \\
  & InCTRL+DART & 78.48 {\color{green!60!black}$\uparrow$} & 87.67 {\color{green!60!black}$\uparrow$} & 83.08 {\color{green!60!black}$\uparrow$} \\
\cmidrule(lr){2-5}
  & ResAD       & 72.30 & 93.40 & 82.85 \\
  & ResAD+Inv   & 71.70 & 94.20 & 82.95 \\
  & ResAD+DART  & 75.70 {\color{green!60!black}$\uparrow$} & 93.60 {\color{green!60!black}$\uparrow$} & 84.65 {\color{green!60!black}$\uparrow$} \\
\midrule
\multirow{8}{*}{$N^2{+}A^1$}
  & NAGL        & 76.85 & 89.57 & 83.21 \\
  & NAGL+DART   & 77.10 {\color{green!60!black}$\uparrow$} & 89.95 {\color{green!60!black}$\uparrow$} & 83.53 {\color{green!60!black}$\uparrow$} \\
\cmidrule(lr){2-5}
  & InCTRL      & 59.80 & 87.28 & 73.54 \\
  & InCTRL+Inv  & 63.79 & 89.24 & 76.52 \\
  & InCTRL+DART & 64.11 {\color{green!60!black}$\uparrow$} & 92.20 {\color{green!60!black}$\uparrow$} & 78.16 {\color{green!60!black}$\uparrow$} \\
\cmidrule(lr){2-5}
  & ResAD       & 66.20 & 93.30 & 79.75 \\
  & ResAD+Inv   & 64.00 & 93.20 & 78.60 \\
  & ResAD+DART  & 71.40 {\color{green!60!black}$\uparrow$} & 93.80 {\color{green!60!black}$\uparrow$} & 82.60 {\color{green!60!black}$\uparrow$} \\
\midrule
\multirow{8}{*}{$N^4{+}A^1$}
  & NAGL        & 78.29 & 93.33 & 85.81 \\
  & NAGL+DART   & 79.28 {\color{green!60!black}$\uparrow$} & 92.54 {\color{red}$\downarrow$} & 85.91 {\color{green!60!black}$\uparrow$} \\
\cmidrule(lr){2-5}
  & InCTRL      & 69.89 & 90.06 & 79.98 \\
  & InCTRL+Inv  & 66.31 & 88.36 & 77.34 \\
  & InCTRL+DART & 77.50 {\color{green!60!black}$\uparrow$} & 93.85 {\color{green!60!black}$\uparrow$} & 85.68 {\color{green!60!black}$\uparrow$} \\
\cmidrule(lr){2-5}
  & ResAD       & 69.50 & 93.40 & 81.45 \\
  & ResAD+Inv   & 67.60 & 93.10 & 80.35 \\
  & ResAD+DART  & 72.60 {\color{green!60!black}$\uparrow$} & 93.90 {\color{green!60!black}$\uparrow$} & 83.25 {\color{green!60!black}$\uparrow$} \\
\bottomrule
\end{tabular}
\end{subtable}
\hfill
\begin{subtable}[t]{0.44\linewidth}
\centering
\caption{Notation reference.}
\label{tab:supp:notation}
\begin{tabular}{ll}
\toprule
Symbol & Meaning \\
\midrule
$\mathbf{r}$                   & Raw residual (before disentanglement) \\
$\mathbf{r}_a$                 & Domain-agnostic anomaly component \\
$\mathbf{r}_d$                 & Domain-specific component (discarded) \\
$E_a$                          & Anomaly encoder (transformer block) \\
$E_d$                          & Domain encoder (MLP) \\
$G$                            & Reconstruction decoder \\
$C_d$                          & Domain classifier with GRL \\
$\mathcal{L}_{\mathrm{suf}}$   & Anomaly sufficiency loss \\
$\mathcal{L}_{\mathrm{adv}}$   & Adversarial domain confusion loss \\
$\mathcal{L}_{\mathrm{recon}}$ & Reconstruction loss \\
$\lambda_{\mathrm{adv}}$       & Weight for $\mathcal{L}_{\mathrm{adv}}$ (= 0.1) \\
$\lambda_{\mathrm{recon}}$     & Weight for $\mathcal{L}_{\mathrm{recon}}$ (= 0.5) \\
$\alpha(t)$                    & GRL strength (DANN schedule) \\
$p_{\mathrm{cross}}(t)$        & Cross-domain episode probability \\
$T_w$                          & Warmup epochs (= 5) \\
$D_i$                          & Reference domain in CDMT episode \\
$D_j$                          & Query domain in CDMT episode \\
$\mathcal{D}_S$                & Source domain pool \\
$\mathcal{D}_T$                & Target domain (held out) \\
$k$                            & Normal references ($\in\{1,2,4\}$) \\
\bottomrule
\end{tabular}
\end{subtable}

\end{table*}

\subsection{Notation Reference}
\label{sec:supp:notation_sec}

For clarity, we provide a unified notation reference in Table~\ref{tab:supp:notation}. The listed symbols are used consistently throughout the paper and supplement.

\subsection{Additional Results}
\label{tab:supp:inctrl}

We provide per-category breakdowns for all three host methods (InCTRL, ResAD, and NAGL) across MVTecAD and VisA datasets for 1, 2, 4 shot settings. For each method, we report results for the vanilla baseline, the naive abnormal reference variant (+Inv), and our proposed DART variant (+DART). All metrics are reported as percentages.

\begin{table*}[p]
\centering
\caption{Per-category MVTecAD results for \textbf{InCTRL} variants, $N^1+A^1$ (1-shot) setting. Pixel-level metrics reported only for InCTRL+DART.}
\label{tab:supp_inctrl_1shot}
\resizebox{\textwidth}{!}{%
\begin{tabular}{l cccccc cccccc cccccc}
\toprule
Category & \multicolumn{6}{c}{\textbf{InCTRL}} & \multicolumn{6}{c}{\textbf{InCTRL+Inv}} & \multicolumn{6}{c}{\textbf{InCTRL+DART}} \\
\cmidrule(lr){2-7}\cmidrule(lr){8-13}\cmidrule(lr){14-19}
Category & I-AUC & AP & I-F1 & P-AUC & PRO & P-F1 & I-AUC & AP & I-F1 & P-AUC & PRO & P-F1 & I-AUC & AP & I-F1 & P-AUC & PRO & P-F1 \\
\midrule
bottle & 99.92 & 99.98 & 99.21 & - & - & - & 99.60 & 99.88 & 98.44 & - & - & - & 99.68 & 99.90 & 98.44 & 97.68 & 89.63 & 68.43 \\
cable & 94.13 & 96.72 & 90.40 & - & - & - & 89.51 & 94.53 & 86.90 & - & - & - & 90.69 & 94.94 & 86.86 & 94.32 & 83.49 & 59.66 \\
capsule & 84.08 & 94.23 & 92.51 & - & - & - & 93.26 & 98.54 & 94.64 & - & - & - & 92.58 & 98.13 & 95.07 & 92.54 & 81.35 & 32.51 \\
carpet & 99.96 & 99.99 & 99.44 & - & - & - & 99.96 & 99.99 & 99.44 & - & - & - & 99.96 & 99.99 & 99.44 & 99.32 & 97.18 & 70.91 \\
grid & 98.33 & 99.53 & 98.25 & - & - & - & 98.75 & 99.63 & 99.12 & - & - & - & 99.00 & 99.69 & 99.12 & 97.68 & 91.47 & 35.77 \\
hazelnut & 99.64 & 99.80 & 97.81 & - & - & - & 97.32 & 98.77 & 95.65 & - & - & - & 97.57 & 98.89 & 97.06 & 98.45 & 89.52 & 59.31 \\
leather & 100.00 & 100.00 & 100.00 & - & - & - & 100.00 & 100.00 & 100.00 & - & - & - & 100.00 & 100.00 & 100.00 & 99.31 & 96.95 & 52.06 \\
metal\_nut & 100.00 & 100.00 & 100.00 & - & - & - & 96.92 & 99.33 & 96.17 & - & - & - & 97.07 & 99.36 & 96.17 & 87.59 & 85.95 & 52.40 \\
pill & 94.19 & 98.85 & 96.06 & - & - & - & 90.40 & 97.40 & 95.07 & - & - & - & 90.78 & 97.54 & 95.41 & 93.90 & 93.35 & 57.96 \\
screw & 73.68 & 87.51 & 86.47 & - & - & - & 81.35 & 92.30 & 87.94 & - & - & - & 84.79 & 93.88 & 88.03 & 91.48 & 73.32 & 20.35 \\
tile & 99.86 & 99.94 & 98.82 & - & - & - & 98.74 & 99.61 & 98.18 & - & - & - & 98.81 & 99.64 & 98.80 & 95.62 & 86.20 & 68.30 \\
toothbrush & 96.11 & 98.54 & 93.33 & - & - & - & 95.28 & 98.30 & 93.33 & - & - & - & 91.67 & 96.18 & 93.33 & 97.39 & 78.99 & 49.77 \\
transistor & 86.88 & 85.31 & 76.71 & - & - & - & 91.92 & 88.04 & 82.22 & - & - & - & 92.38 & 88.33 & 83.33 & 87.10 & 64.62 & 40.87 \\
wood & 99.21 & 99.78 & 98.33 & - & - & - & 98.95 & 99.71 & 98.33 & - & - & - & 99.12 & 99.75 & 98.33 & 95.26 & 89.90 & 63.51 \\
zipper & 98.84 & 99.73 & 98.72 & - & - & - & 99.24 & 99.83 & 99.15 & - & - & - & 99.19 & 99.82 & 98.73 & 90.10 & 76.53 & 35.61 \\
\midrule
\textbf{mean} & 94.99 & 97.33 & 95.07 & - & - & - & 95.41 & 97.72 & 94.97 & - & - & - & 95.55 & 97.74 & 95.21 & 94.52 & 85.23 & 51.16 \\
\bottomrule
\end{tabular}%
}
\end{table*}

\begin{table*}[p]
\centering
\caption{Per-category MVTecAD results for \textbf{InCTRL} variants, $N^2+A^1$ (2-shot) setting. Pixel-level metrics reported only for InCTRL+DART.}
\label{tab:supp_inctrl_2shot}
\resizebox{\textwidth}{!}{%
\begin{tabular}{l cccccc cccccc cccccc}
\toprule
Category & \multicolumn{6}{c}{\textbf{InCTRL}} & \multicolumn{6}{c}{\textbf{InCTRL+Inv}} & \multicolumn{6}{c}{\textbf{InCTRL+DART}} \\
\cmidrule(lr){2-7}\cmidrule(lr){8-13}\cmidrule(lr){14-19}
Category & I-AUC & AP & I-F1 & P-AUC & PRO & P-F1 & I-AUC & AP & I-F1 & P-AUC & PRO & P-F1 & I-AUC & AP & I-F1 & P-AUC & PRO & P-F1 \\
\midrule
bottle & 99.84 & 99.95 & 99.20 & - & - & - & 99.76 & 99.93 & 98.44 & - & - & - & 99.68 & 99.90 & 98.44 & 97.47 & 89.01 & 68.12 \\
cable & 94.06 & 96.73 & 90.80 & - & - & - & 90.37 & 94.96 & 86.32 & - & - & - & 91.47 & 95.42 & 87.43 & 94.19 & 82.93 & 59.06 \\
capsule & 81.85 & 93.41 & 92.44 & - & - & - & 91.70 & 98.26 & 92.64 & - & - & - & 92.90 & 98.49 & 94.06 & 92.41 & 82.87 & 31.38 \\
carpet & 99.92 & 99.97 & 99.44 & - & - & - & 99.96 & 99.99 & 99.44 & - & - & - & 99.96 & 99.99 & 99.44 & 99.30 & 96.61 & 69.58 \\
grid & 98.33 & 99.53 & 98.25 & - & - & - & 98.83 & 99.65 & 99.12 & - & - & - & 99.16 & 99.74 & 99.12 & 97.86 & 92.33 & 36.19 \\
hazelnut & 99.89 & 99.94 & 99.29 & - & - & - & 97.61 & 98.90 & 97.06 & - & - & - & 97.82 & 99.00 & 97.06 & 98.69 & 90.28 & 63.91 \\
leather & 100.00 & 100.00 & 100.00 & - & - & - & 100.00 & 100.00 & 100.00 & - & - & - & 100.00 & 100.00 & 100.00 & 99.32 & 96.86 & 52.25 \\
metal\_nut & 100.00 & 100.00 & 100.00 & - & - & - & 98.53 & 99.69 & 97.87 & - & - & - & 98.68 & 99.72 & 97.87 & 90.51 & 88.10 & 60.69 \\
pill & 94.24 & 98.80 & 95.44 & - & - & - & 92.80 & 98.55 & 94.74 & - & - & - & 92.39 & 98.44 & 94.41 & 93.87 & 93.89 & 58.07 \\
screw & 71.70 & 86.53 & 86.69 & - & - & - & 80.63 & 92.01 & 87.69 & - & - & - & 84.65 & 93.79 & 87.90 & 90.71 & 71.16 & 19.50 \\
tile & 99.89 & 99.96 & 98.82 & - & - & - & 99.49 & 99.83 & 99.40 & - & - & - & 99.60 & 99.86 & 99.40 & 95.68 & 86.72 & 67.70 \\
toothbrush & 97.22 & 98.93 & 95.08 & - & - & - & 94.72 & 98.24 & 94.92 & - & - & - & 94.17 & 98.02 & 93.33 & 98.14 & 78.94 & 51.98 \\
transistor & 90.25 & 89.89 & 83.54 & - & - & - & 92.54 & 88.47 & 84.78 & - & - & - & 93.79 & 89.79 & 83.87 & 88.36 & 66.41 & 43.27 \\
wood & 99.30 & 99.80 & 99.16 & - & - & - & 99.21 & 99.78 & 98.33 & - & - & - & 99.21 & 99.78 & 98.33 & 95.14 & 89.80 & 60.78 \\
zipper & 99.45 & 99.86 & 98.31 & - & - & - & 99.76 & 99.94 & 99.15 & - & - & - & 99.68 & 99.92 & 99.15 & 90.52 & 78.86 & 43.02 \\
\midrule
\textbf{mean} & 95.06 & 97.55 & 95.76 & - & - & - & 95.73 & 97.88 & 95.33 & - & - & - & 96.21 & 98.12 & 95.32 & 94.81 & 85.65 & 52.37 \\
\bottomrule
\end{tabular}%
}
\end{table*}

\begin{table*}[p]
\centering
\caption{Per-category MVTecAD results for \textbf{InCTRL} variants, $N^4+A^1$ (4-shot) setting. Pixel-level metrics reported only for InCTRL+DART.}
\label{tab:supp_inctrl_4shot}
\resizebox{\textwidth}{!}{%
\begin{tabular}{l cccccc cccccc cccccc}
\toprule
Category & \multicolumn{6}{c}{\textbf{InCTRL}} & \multicolumn{6}{c}{\textbf{InCTRL+Inv}} & \multicolumn{6}{c}{\textbf{InCTRL+DART}} \\
\cmidrule(lr){2-7}\cmidrule(lr){8-13}\cmidrule(lr){14-19}
Category & I-AUC & AP & I-F1 & P-AUC & PRO & P-F1 & I-AUC & AP & I-F1 & P-AUC & PRO & P-F1 & I-AUC & AP & I-F1 & P-AUC & PRO & P-F1 \\
\midrule
bottle & 100.00 & 100.00 & 100.00 & - & - & - & 99.84 & 99.95 & 99.21 & - & - & - & 99.92 & 99.98 & 99.21 & 97.71 & 89.12 & 68.98 \\
cable & 94.55 & 97.09 & 91.71 & - & - & - & 90.03 & 94.83 & 87.72 & - & - & - & 93.02 & 95.32 & 91.64 & 94.67 & 83.46 & 60.02 \\
capsule & 81.09 & 89.53 & 93.58 & - & - & - & 76.43 & 94.61 & 90.46 & - & - & - & 82.14 & 95.60 & 92.46 & 91.88 & 84.46 & 36.29 \\
carpet & 99.96 & 99.99 & 99.44 & - & - & - & 100.00 & 100.00 & 100.00 & - & - & - & 100.00 & 100.00 & 100.00 & 99.24 & 96.42 & 67.11 \\
grid & 98.83 & 99.65 & 99.12 & - & - & - & 98.41 & 99.46 & 96.49 & - & - & - & 98.83 & 99.61 & 98.21 & 98.19 & 92.88 & 37.96 \\
hazelnut & 99.71 & 99.83 & 99.29 & - & - & - & 96.54 & 98.18 & 97.10 & - & - & - & 98.82 & 98.32 & 98.10 & 98.99 & 92.86 & 66.11 \\
leather & 100.00 & 100.00 & 100.00 & - & - & - & 100.00 & 100.00 & 100.00 & - & - & - & 100.00 & 100.00 & 100.00 & 99.33 & 96.89 & 52.48 \\
metal\_nut & 100.00 & 100.00 & 100.00 & - & - & - & 97.26 & 99.38 & 96.26 & - & - & - & 97.75 & 99.50 & 98.81 & 92.10 & 89.42 & 64.50 \\
pill & 95.42 & 99.08 & 96.86 & - & - & - & 91.84 & 98.44 & 93.33 & - & - & - & 89.83 & 97.32 & 95.43 & 94.29 & 93.67 & 57.96 \\
screw & 78.44 & 91.22 & 87.69 & - & - & - & 83.56 & 94.42 & 87.41 & - & - & - & 88.71 & 94.94 & 89.73 & 93.94 & 74.60 & 13.87 \\
tile & 100.00 & 100.00 & 100.00 & - & - & - & 99.28 & 99.77 & 99.40 & - & - & - & 99.35 & 99.79 & 99.40 & 95.64 & 85.54 & 67.31 \\
toothbrush & 95.83 & 98.37 & 94.92 & - & - & - & 90.00 & 96.40 & 89.29 & - & - & - & 95.50 & 97.36 & 95.10 & 98.05 & 77.13 & 52.38 \\
transistor & 89.12 & 87.27 & 80.00 & - & - & - & 91.62 & 86.48 & 82.76 & - & - & - & 92.96 & 89.08 & 84.72 & 91.18 & 68.45 & 49.01 \\
wood & 98.60 & 99.59 & 97.44 & - & - & - & 98.25 & 99.52 & 98.31 & - & - & - & 99.04 & 99.72 & 98.31 & 95.30 & 88.64 & 61.42 \\
zipper & 99.61 & 99.90 & 99.15 & - & - & - & 99.71 & 99.93 & 99.15 & - & - & - & 99.55 & 99.89 & 99.15 & 92.24 & 79.82 & 44.19 \\
\midrule
\textbf{mean} & 95.41 & 97.44 & 95.95 & - & - & - & 94.19 & 97.43 & 94.46 & - & - & - & 95.64 & 97.76 & 96.08 & 95.52 & 86.22 & 53.31 \\
\bottomrule
\end{tabular}%
}
\end{table*}

\begin{table*}[p]
\centering
\caption{Per-category \textbf{VisA} results for \textbf{InCTRL} variants, $N^1+A^1$ (1-shot) setting. Pixel-level metrics reported only for InCTRL+DART.}
\label{tab:supp_inctrl_visa_1shot}
\resizebox{\textwidth}{!}{%
\begin{tabular}{l cccccc cccccc cccccc}
\toprule
Category & \multicolumn{6}{c}{\textbf{InCTRL}} & \multicolumn{6}{c}{\textbf{InCTRL+Inv}} & \multicolumn{6}{c}{\textbf{InCTRL+DART}} \\
\cmidrule(lr){2-7}\cmidrule(lr){8-13}\cmidrule(lr){14-19}
Category & I-AUC & AP & I-F1 & P-AUC & PRO & P-F1 & I-AUC & AP & I-F1 & P-AUC & PRO & P-F1 & I-AUC & AP & I-F1 & P-AUC & PRO & P-F1 \\
\midrule
candle & 83.40 & 39.46 & 48.57 & - & - & - & 80.79 & 65.08 & 65.38 & - & - & - & 85.60 & 53.64 & 53.97 & 98.61 & 88.59 & 26.32 \\
capsules & 67.76 & 49.66 & 51.61 & - & - & - & 67.94 & 49.20 & 51.43 & - & - & - & 66.45 & 35.86 & 41.86 & 92.84 & 56.32 & 27.16 \\
cashew & 91.00 & 74.09 & 67.86 & - & - & - & 89.50 & 78.55 & 72.00 & - & - & - & 86.63 & 65.56 & 60.00 & 98.38 & 77.92 & 34.23 \\
chewinggum & 97.09 & 93.75 & 88.89 & - & - & - & 99.23 & 97.00 & 90.48 & - & - & - & 96.41 & 90.20 & 86.96 & 98.92 & 79.07 & 50.84 \\
fryum & 95.40 & 89.32 & 85.71 & - & - & - & 77.10 & 65.86 & 66.67 & - & - & - & 94.50 & 86.29 & 82.93 & 92.46 & 72.27 & 30.28 \\
macaroni1 & 83.12 & 48.23 & 54.90 & - & - & - & 82.19 & 41.06 & 45.83 & - & - & - & 84.17 & 63.13 & 64.86 & 96.35 & 74.84 & 21.63 \\
macaroni2 & 75.74 & 36.27 & 44.12 & - & - & - & 70.33 & 30.90 & 41.03 & - & - & - & 80.81 & 59.80 & 65.00 & 94.19 & 75.10 & 16.56 \\
pcb1 & 87.45 & 35.80 & 47.46 & - & - & - & 92.79 & 58.62 & 58.33 & - & - & - & 86.81 & 34.97 & 48.28 & 97.23 & 68.78 & 28.06 \\
pcb2 & 80.29 & 24.00 & 37.21 & - & - & - & 60.39 & 12.42 & 20.74 & - & - & - & 80.29 & 23.78 & 37.21 & 93.25 & 51.19 & 8.58 \\
pcb3 & 71.48 & 21.47 & 34.09 & - & - & - & 51.11 & 11.02 & 19.82 & - & - & - & 76.91 & 39.97 & 47.37 & 91.34 & 53.18 & 28.48 \\
pcb4 & 96.50 & 69.85 & 81.82 & - & - & - & 98.28 & 87.44 & 87.50 & - & - & - & 96.53 & 74.22 & 81.82 & 92.35 & 45.83 & 10.05 \\
pipe\_fryum & 96.38 & 85.87 & 78.26 & - & - & - & 88.76 & 67.24 & 76.19 & - & - & - & 97.57 & 93.77 & 89.47 & 99.39 & 86.67 & 44.68 \\
\midrule
\textbf{mean} & 85.47 & 55.65 & 60.04 & - & - & - & 79.87 & 55.37 & 57.95 & - & - & - & 86.06 & 60.10 & 63.31 & 95.44 & 69.15 & 27.24 \\
\bottomrule
\end{tabular}%
}
\end{table*}

\begin{table*}[p]
\centering
\caption{Per-category \textbf{VisA} results for \textbf{InCTRL} variants, $N^2+A^1$ (2-shot) setting. Pixel-level metrics reported only for InCTRL+DART.}
\label{tab:supp_inctrl_visa_2shot}
\resizebox{\textwidth}{!}{%
\begin{tabular}{l cccccc cccccc cccccc}
\toprule
Category & \multicolumn{6}{c}{\textbf{InCTRL}} & \multicolumn{6}{c}{\textbf{InCTRL+Inv}} & \multicolumn{6}{c}{\textbf{InCTRL+DART}} \\
\cmidrule(lr){2-7}\cmidrule(lr){8-13}\cmidrule(lr){14-19}
Category & I-AUC & AP & I-F1 & P-AUC & PRO & P-F1 & I-AUC & AP & I-F1 & P-AUC & PRO & P-F1 & I-AUC & AP & I-F1 & P-AUC & PRO & P-F1 \\
\midrule
candle & 81.85 & 29.84 & 47.50 & - & - & - & 89.35 & 73.72 & 72.73 & - & - & - & 87.44 & 55.53 & 58.06 & 98.41 & 88.24 & 25.93 \\
capsules & 71.18 & 51.13 & 50.00 & - & - & - & 71.84 & 50.37 & 51.43 & - & - & - & 70.00 & 35.82 & 42.86 & 94.73 & 58.37 & 26.11 \\
cashew & 91.71 & 76.04 & 73.47 & - & - & - & 89.75 & 78.32 & 73.47 & - & - & - & 91.54 & 80.59 & 72.34 & 98.34 & 77.06 & 34.73 \\
chewinggum & 97.41 & 94.40 & 90.91 & - & - & - & 99.27 & 97.18 & 90.91 & - & - & - & 97.05 & 92.27 & 87.50 & 98.81 & 77.89 & 50.97 \\
fryum & 97.65 & 91.89 & 87.80 & - & - & - & 80.90 & 69.93 & 70.59 & - & - & - & 95.15 & 87.86 & 87.18 & 94.48 & 76.12 & 33.65 \\
macaroni1 & 83.52 & 44.58 & 57.14 & - & - & - & 78.60 & 37.75 & 46.15 & - & - & - & 85.93 & 63.44 & 61.11 & 96.38 & 77.73 & 21.63 \\
macaroni2 & 83.40 & 39.57 & 50.00 & - & - & - & 75.26 & 27.90 & 34.78 & - & - & - & 88.29 & 66.41 & 65.00 & 96.23 & 76.13 & 15.01 \\
pcb1 & 90.90 & 46.36 & 52.46 & - & - & - & 93.86 & 61.64 & 62.22 & - & - & - & 90.33 & 44.18 & 50.75 & 97.88 & 68.13 & 29.69 \\
pcb2 & 80.76 & 27.96 & 41.86 & - & - & - & 57.05 & 11.12 & 19.29 & - & - & - & 80.79 & 27.90 & 41.86 & 93.47 & 50.86 & 9.54 \\
pcb3 & 86.21 & 53.72 & 60.00 & - & - & - & 54.86 & 15.93 & 20.00 & - & - & - & 82.70 & 49.80 & 55.32 & 94.11 & 58.13 & 29.04 \\
pcb4 & 96.40 & 67.66 & 78.95 & - & - & - & 98.28 & 86.23 & 83.08 & - & - & - & 96.38 & 67.59 & 78.95 & 94.07 & 50.22 & 12.91 \\
pipe\_fryum & 94.48 & 80.90 & 73.17 & - & - & - & 87.90 & 68.52 & 70.83 & - & - & - & 98.05 & 93.29 & 87.80 & 99.51 & 86.54 & 47.31 \\
\midrule
\textbf{mean} & 87.96 & 58.67 & 63.61 & - & - & - & 81.41 & 56.55 & 57.96 & - & - & - & 88.64 & 63.72 & 65.73 & 96.37 & 70.45 & 28.04 \\
\bottomrule
\end{tabular}%
}
\end{table*}

\begin{table*}[p]
\centering
\caption{Per-category \textbf{VisA} results for \textbf{InCTRL} variants, $N^4+A^1$ (4-shot) setting. Pixel-level metrics reported only for InCTRL+DART.}
\label{tab:supp_inctrl_visa_4shot}
\resizebox{\textwidth}{!}{%
\begin{tabular}{l cccccc cccccc cccccc}
\toprule
Category & \multicolumn{6}{c}{\textbf{InCTRL}} & \multicolumn{6}{c}{\textbf{InCTRL+Inv}} & \multicolumn{6}{c}{\textbf{InCTRL+DART}} \\
\cmidrule(lr){2-7}\cmidrule(lr){8-13}\cmidrule(lr){14-19}
Category & I-AUC & AP & I-F1 & P-AUC & PRO & P-F1 & I-AUC & AP & I-F1 & P-AUC & PRO & P-F1 & I-AUC & AP & I-F1 & P-AUC & PRO & P-F1 \\
\midrule
candle & 78.37 & 23.08 & 43.14 & - & - & - & 88.98 & 71.93 & 66.67 & - & - & - & 88.81 & 56.61 & 54.55 & 98.60 & 88.32 & 25.18 \\
capsules & 73.64 & 50.80 & 48.48 & - & - & - & 66.23 & 41.34 & 41.38 & - & - & - & 73.46 & 48.39 & 48.48 & 94.58 & 58.90 & 26.80 \\
cashew & 90.71 & 78.60 & 73.17 & - & - & - & 89.38 & 77.17 & 71.70 & - & - & - & 90.62 & 80.08 & 71.11 & 98.16 & 77.19 & 35.25 \\
chewinggum & 97.09 & 88.34 & 93.33 & - & - & - & 93.36 & 90.76 & 86.36 & - & - & - & 98.00 & 96.79 & 95.45 & 98.81 & 77.97 & 52.46 \\
fryum & 97.80 & 94.26 & 88.89 & - & - & - & 89.90 & 81.73 & 76.92 & - & - & - & 96.10 & 91.63 & 86.49 & 95.95 & 72.60 & 36.45 \\
macaroni1 & 92.55 & 67.87 & 66.67 & - & - & - & 61.43 & 23.53 & 23.40 & - & - & - & 92.57 & 67.80 & 66.67 & 96.92 & 80.02 & 19.12 \\
macaroni2 & 89.33 & 49.31 & 63.83 & - & - & - & 78.55 & 31.14 & 40.82 & - & - & - & 90.14 & 71.54 & 73.68 & 96.88 & 75.77 & 18.57 \\
pcb1 & 93.83 & 56.89 & 59.02 & - & - & - & 93.69 & 62.79 & 57.14 & - & - & - & 93.60 & 56.50 & 57.89 & 98.31 & 70.29 & 33.59 \\
pcb2 & 83.61 & 48.95 & 63.16 & - & - & - & 51.55 & 8.99 & 18.29 & - & - & - & 83.68 & 49.31 & 64.86 & 93.98 & 55.53 & 11.50 \\
pcb3 & 78.77 & 40.89 & 54.55 & - & - & - & 60.18 & 12.19 & 23.76 & - & - & - & 79.15 & 48.09 & 53.57 & 95.06 & 56.23 & 31.49 \\
pcb4 & 97.16 & 78.53 & 80.00 & - & - & - & 98.14 & 84.49 & 84.85 & - & - & - & 97.18 & 79.07 & 80.00 & 96.24 & 53.97 & 15.90 \\
pipe\_fryum & 97.10 & 86.54 & 80.77 & - & - & - & 92.14 & 79.88 & 74.42 & - & - & - & 98.33 & 93.58 & 87.80 & 99.53 & 87.32 & 47.34 \\
\midrule
\textbf{mean} & 89.16 & 63.67 & 67.92 & - & - & - & 80.29 & 55.49 & 55.48 & - & - & - & 90.14 & 69.95 & 70.05 & 96.92 & 71.18 & 29.47 \\
\bottomrule
\end{tabular}%
}
\end{table*}

\label{tab:supp:resad}

\begin{table*}[p]
\centering
\caption{Per-category MVTecAD results for \textbf{ResAD} variants, $N^1+A^1$ (1-shot) setting.}
\label{tab:supp_resad_vtm_1shot}
\resizebox{\textwidth}{!}{%
\begin{tabular}{l cccccc cccccc cccccc}
\toprule
 & \multicolumn{6}{c}{\textbf{ResAD}} & \multicolumn{6}{c}{\textbf{ResAD+Inv}} & \multicolumn{6}{c}{\textbf{ResAD+DART}} \\
\cmidrule(lr){2-7}\cmidrule(lr){8-13}\cmidrule(lr){14-19}
Category & I-AUC & AP & I-F1 & P-AUC & PRO & P-F1 & I-AUC & AP & I-F1 & P-AUC & PRO & P-F1 & I-AUC & AP & I-F1 & P-AUC & PRO & P-F1 \\
\midrule
bottle & 100.00 & 100.00 & 100.00 & 98.65 & 96.06 & 76.37 & 99.52 & 99.84 & 99.21 & 98.83 & 95.63 & 78.45 & 100.00 & 100.00 & 100.00 & 98.70 & 96.27 & 77.18 \\
cable & 79.20 & 87.47 & 82.24 & 93.06 & 73.38 & 34.87 & 79.87 & 87.83 & 80.21 & 90.52 & 69.03 & 27.24 & 81.09 & 88.36 & 81.31 & 93.25 & 75.29 & 37.91 \\
capsule & 78.30 & 94.44 & 93.56 & 97.80 & 91.61 & 44.26 & 77.58 & 94.56 & 91.30 & 97.72 & 89.86 & 45.86 & 81.65 & 95.40 & 92.37 & 97.83 & 92.01 & 45.06 \\
carpet & 99.72 & 99.91 & 98.86 & 99.28 & 96.57 & 65.32 & 95.87 & 98.77 & 95.46 & 98.52 & 93.93 & 59.03 & 99.32 & 99.80 & 97.75 & 99.25 & 96.58 & 66.91 \\
grid & 77.94 & 91.28 & 87.69 & 83.14 & 64.08 & 23.34 & 72.77 & 86.23 & 87.02 & 79.00 & 55.30 & 17.13 & 82.21 & 92.99 & 88.00 & 81.35 & 64.97 & 24.04 \\
hazelnut & 83.93 & 91.79 & 84.44 & 97.70 & 84.99 & 55.62 & 88.11 & 94.12 & 87.59 & 97.98 & 86.26 & 61.99 & 77.18 & 88.03 & 79.41 & 97.45 & 81.39 & 56.96 \\
leather & 100.00 & 100.00 & 100.00 & 99.35 & 97.98 & 49.21 & 100.00 & 100.00 & 100.00 & 99.40 & 98.15 & 51.46 & 99.93 & 99.98 & 99.45 & 99.40 & 97.90 & 53.33 \\
metal\_nut & 87.00 & 96.81 & 91.63 & 96.73 & 86.22 & 78.25 & 74.29 & 92.82 & 91.18 & 95.36 & 76.35 & 70.07 & 86.12 & 96.49 & 91.63 & 96.51 & 86.23 & 78.49 \\
pill & 91.84 & 98.43 & 95.41 & 97.26 & 95.81 & 65.07 & 87.37 & 97.28 & 93.43 & 97.26 & 96.19 & 65.36 & 91.93 & 98.48 & 94.37 & 97.47 & 95.89 & 66.33 \\
screw & 47.28 & 74.59 & 85.61 & 94.02 & 79.64 & 4.69 & 48.68 & 75.54 & 85.61 & 91.94 & 72.78 & 3.45 & 50.58 & 76.15 & 85.61 & 94.63 & 81.30 & 7.23 \\
tile & 99.21 & 99.70 & 97.01 & 95.57 & 85.72 & 63.39 & 98.34 & 99.41 & 97.01 & 93.02 & 81.65 & 58.55 & 99.24 & 99.71 & 97.01 & 95.79 & 86.27 & 67.61 \\
toothbrush & 74.72 & 89.86 & 86.96 & 97.13 & 74.27 & 52.40 & 68.61 & 86.39 & 85.71 & 95.93 & 70.27 & 37.84 & 70.83 & 87.62 & 85.71 & 96.00 & 71.48 & 42.09 \\
transistor & 78.00 & 76.92 & 68.24 & 74.91 & 62.62 & 33.11 & 64.46 & 59.53 & 63.74 & 72.37 & 59.60 & 31.33 & 79.71 & 77.57 & 70.10 & 73.46 & 61.92 & 31.10 \\
wood & 99.21 & 99.77 & 98.30 & 95.25 & 89.72 & 55.07 & 99.30 & 99.80 & 99.16 & 94.50 & 87.13 & 53.89 & 98.77 & 99.64 & 97.44 & 94.90 & 86.78 & 56.78 \\
zipper & 88.68 & 95.70 & 93.98 & 96.73 & 88.30 & 46.63 & 85.90 & 94.88 & 92.61 & 95.08 & 83.51 & 38.05 & 90.34 & 96.47 & 94.17 & 95.98 & 85.98 & 44.21 \\
\midrule
\textbf{mean} & 85.67 & 93.11 & 90.93 & 94.44 & 84.46 & 49.84 & 82.71 & 91.13 & 89.95 & 93.16 & 81.04 & 46.65 & 85.93 & 93.11 & 90.29 & 94.13 & 84.02 & 50.35 \\
\bottomrule
\end{tabular}%
}
\end{table*}

\begin{table*}[p]
\centering
\caption{Per-category MVTecAD results for \textbf{ResAD} variants, $N^2+A^1$ (2-shot) setting.}
\label{tab:supp_resad_vtm_2shot}
\resizebox{\textwidth}{!}{%
\begin{tabular}{l cccccc cccccc cccccc}
\toprule
 & \multicolumn{6}{c}{\textbf{ResAD}} & \multicolumn{6}{c}{\textbf{ResAD+Inv}} & \multicolumn{6}{c}{\textbf{ResAD+DART}} \\
\cmidrule(lr){2-7}\cmidrule(lr){8-13}\cmidrule(lr){14-19}
Category & I-AUC & AP & I-F1 & P-AUC & PRO & P-F1 & I-AUC & AP & I-F1 & P-AUC & PRO & P-F1 & I-AUC & AP & I-F1 & P-AUC & PRO & P-F1 \\
\midrule
bottle & 99.92 & 99.98 & 99.21 & 98.76 & 96.39 & 77.40 & 99.44 & 99.82 & 99.21 & 98.88 & 95.71 & 78.87 & 100.00 & 100.00 & 100.00 & 98.80 & 96.71 & 78.27 \\
cable & 80.38 & 88.10 & 83.18 & 93.97 & 76.69 & 38.62 & 81.99 & 88.89 & 81.67 & 91.99 & 73.10 & 29.94 & 85.98 & 91.27 & 83.65 & 93.97 & 77.97 & 40.42 \\
capsule & 83.21 & 95.77 & 92.98 & 98.33 & 93.15 & 44.60 & 85.04 & 96.46 & 92.51 & 98.30 & 92.25 & 48.25 & 86.28 & 96.68 & 93.21 & 98.35 & 93.37 & 45.49 \\
carpet & 99.76 & 99.93 & 98.86 & 99.27 & 96.46 & 65.03 & 97.39 & 99.24 & 96.59 & 98.69 & 94.67 & 59.33 & 99.24 & 99.78 & 98.30 & 99.24 & 96.54 & 66.59 \\
grid & 83.46 & 93.17 & 88.71 & 87.53 & 71.09 & 23.31 & 83.29 & 93.40 & 88.00 & 83.28 & 60.24 & 14.73 & 87.80 & 95.42 & 90.91 & 84.75 & 65.16 & 24.39 \\
hazelnut & 83.75 & 91.95 & 81.54 & 98.13 & 87.65 & 57.51 & 87.36 & 93.69 & 83.66 & 98.41 & 88.57 & 63.78 & 77.68 & 88.08 & 80.25 & 97.97 & 85.15 & 58.26 \\
leather & 100.00 & 100.00 & 100.00 & 99.37 & 98.01 & 49.42 & 100.00 & 100.00 & 100.00 & 99.41 & 98.14 & 50.77 & 100.00 & 100.00 & 100.00 & 99.45 & 98.15 & 53.41 \\
metal\_nut & 88.42 & 97.10 & 92.54 & 96.97 & 87.20 & 79.16 & 77.96 & 94.08 & 91.54 & 95.87 & 78.87 & 73.00 & 87.68 & 97.05 & 92.00 & 96.83 & 87.51 & 79.45 \\
pill & 92.74 & 98.60 & 95.41 & 97.51 & 96.20 & 65.67 & 89.47 & 97.81 & 94.16 & 97.60 & 96.47 & 66.40 & 92.06 & 98.50 & 94.70 & 97.67 & 96.17 & 66.82 \\
screw & 52.04 & 74.46 & 85.61 & 95.13 & 82.76 & 8.04 & 47.57 & 74.20 & 85.61 & 93.11 & 75.96 & 4.39 & 53.00 & 76.40 & 85.61 & 95.43 & 83.68 & 10.55 \\
tile & 99.53 & 99.82 & 97.67 & 95.64 & 85.93 & 63.67 & 98.63 & 99.47 & 96.47 & 93.51 & 83.17 & 59.44 & 99.53 & 99.82 & 98.20 & 95.85 & 86.59 & 67.22 \\
toothbrush & 73.06 & 88.98 & 88.23 & 97.19 & 74.64 & 54.25 & 69.17 & 86.37 & 88.23 & 95.79 & 69.97 & 37.43 & 71.11 & 87.81 & 88.23 & 96.10 & 71.91 & 44.96 \\
transistor & 86.50 & 86.57 & 76.71 & 79.28 & 64.98 & 37.45 & 64.79 & 63.04 & 59.79 & 75.91 & 62.39 & 33.82 & 88.12 & 86.09 & 78.65 & 77.96 & 64.21 & 35.57 \\
wood & 99.30 & 99.79 & 98.33 & 95.26 & 90.44 & 55.21 & 99.39 & 99.83 & 99.16 & 94.71 & 87.50 & 54.07 & 98.95 & 99.67 & 98.33 & 94.99 & 87.89 & 56.69 \\
zipper & 91.96 & 97.01 & 96.69 & 97.88 & 92.40 & 54.12 & 91.67 & 97.21 & 94.69 & 97.04 & 89.60 & 48.00 & 93.38 & 97.74 & 95.51 & 97.64 & 91.49 & 53.34 \\
\midrule
\textbf{mean} & 87.60 & 94.08 & 91.71 & 95.35 & 86.27 & 51.56 & 84.88 & 92.23 & 90.09 & 94.17 & 83.11 & 48.15 & 88.05 & 94.29 & 91.84 & 95.00 & 85.50 & 52.10 \\
\bottomrule
\end{tabular}%
}
\end{table*}

\begin{table*}[p]
\centering
\caption{Per-category MVTecAD results for \textbf{ResAD} variants, $N^4+A^1$ (4-shot) setting.}
\label{tab:supp_resad_vtm_4shot}
\resizebox{\textwidth}{!}{%
\begin{tabular}{l cccccc cccccc cccccc}
\toprule
 & \multicolumn{6}{c}{\textbf{ResAD}} & \multicolumn{6}{c}{\textbf{ResAD+Inv}} & \multicolumn{6}{c}{\textbf{ResAD+DART}} \\
\cmidrule(lr){2-7}\cmidrule(lr){8-13}\cmidrule(lr){14-19}
Category & I-AUC & AP & I-F1 & P-AUC & PRO & P-F1 & I-AUC & AP & I-F1 & P-AUC & PRO & P-F1 & I-AUC & AP & I-F1 & P-AUC & PRO & P-F1 \\
\midrule
bottle & 99.76 & 99.92 & 99.21 & 98.81 & 96.46 & 77.62 & 99.44 & 99.82 & 99.21 & 98.89 & 95.90 & 78.83 & 100.00 & 100.00 & 100.00 & 98.88 & 96.72 & 78.71 \\
cable & 84.58 & 90.54 & 85.18 & 95.04 & 79.48 & 45.04 & 82.74 & 89.37 & 80.79 & 93.17 & 75.25 & 32.99 & 86.73 & 91.88 & 83.17 & 95.03 & 80.84 & 45.98 \\
capsule & 89.51 & 97.40 & 94.22 & 98.53 & 93.45 & 47.20 & 86.84 & 96.94 & 94.32 & 98.58 & 92.90 & 48.54 & 91.62 & 97.97 & 94.74 & 98.54 & 93.75 & 47.40 \\
carpet & 99.72 & 99.91 & 98.32 & 99.23 & 96.22 & 64.06 & 97.35 & 99.21 & 96.59 & 98.48 & 93.82 & 57.59 & 99.28 & 99.79 & 98.30 & 99.20 & 96.23 & 65.59 \\
grid & 90.48 & 96.48 & 90.00 & 91.17 & 75.78 & 30.55 & 90.48 & 96.18 & 92.44 & 87.91 & 68.39 & 19.93 & 92.40 & 97.08 & 92.56 & 89.29 & 69.78 & 28.21 \\
hazelnut & 88.79 & 94.10 & 85.16 & 98.40 & 89.76 & 59.67 & 92.57 & 96.13 & 88.73 & 98.51 & 90.25 & 64.56 & 84.64 & 92.05 & 83.97 & 98.30 & 88.04 & 60.41 \\
leather & 100.00 & 100.00 & 100.00 & 99.37 & 98.16 & 49.22 & 100.00 & 100.00 & 100.00 & 99.42 & 98.09 & 50.93 & 100.00 & 100.00 & 100.00 & 99.45 & 98.13 & 53.49 \\
metal\_nut & 93.94 & 98.61 & 93.12 & 98.17 & 91.98 & 84.39 & 89.25 & 97.34 & 91.84 & 97.78 & 87.54 & 81.14 & 92.86 & 98.34 & 93.55 & 98.07 & 91.63 & 84.53 \\
pill & 91.65 & 98.40 & 94.81 & 97.58 & 96.41 & 66.25 & 89.66 & 97.87 & 94.12 & 97.75 & 96.65 & 67.37 & 92.31 & 98.56 & 95.04 & 97.74 & 96.35 & 67.41 \\
screw & 56.61 & 76.03 & 85.61 & 95.96 & 85.02 & 11.36 & 53.39 & 76.00 & 86.23 & 94.26 & 79.22 & 5.32 & 56.22 & 77.01 & 85.61 & 96.19 & 85.76 & 12.45 \\
tile & 99.60 & 99.84 & 98.25 & 95.86 & 86.35 & 64.31 & 98.49 & 99.42 & 97.01 & 93.36 & 82.96 & 58.53 & 99.68 & 99.87 & 98.81 & 96.13 & 86.76 & 67.99 \\
toothbrush & 73.06 & 88.94 & 88.23 & 96.95 & 74.16 & 52.11 & 68.06 & 85.55 & 88.23 & 95.45 & 69.49 & 34.57 & 70.83 & 87.45 & 88.23 & 95.76 & 71.46 & 44.06 \\
transistor & 88.67 & 88.58 & 81.08 & 79.92 & 66.03 & 37.83 & 68.63 & 69.89 & 61.36 & 77.10 & 63.60 & 34.57 & 88.96 & 88.54 & 81.08 & 78.63 & 65.34 & 36.33 \\
wood & 99.30 & 99.79 & 98.33 & 95.24 & 89.98 & 55.37 & 99.47 & 99.85 & 99.16 & 94.73 & 87.52 & 54.19 & 98.77 & 99.62 & 97.52 & 94.96 & 87.50 & 56.75 \\
zipper & 92.75 & 97.51 & 95.87 & 98.10 & 92.76 & 55.88 & 94.09 & 98.23 & 94.96 & 97.40 & 90.80 & 50.63 & 93.75 & 97.98 & 95.55 & 97.92 & 93.04 & 55.48 \\
\midrule
\textbf{mean} & 89.89 & 95.07 & 92.49 & 95.89 & 87.47 & 53.39 & 87.36 & 93.45 & 91.00 & 94.85 & 84.83 & 49.31 & 89.87 & 95.08 & 92.54 & 95.61 & 86.76 & 53.65 \\
\bottomrule
\end{tabular}%
}
\end{table*}

\begin{table*}[p]
\centering
\caption{Per-category VisA results for \textbf{ResAD} variants, $N^1+A^1$ (1-shot) setting.}
\label{tab:supp_resad_mtv_1shot}
\resizebox{\textwidth}{!}{%
\begin{tabular}{l cccccc cccccc cccccc}
\toprule
 & \multicolumn{6}{c}{\textbf{ResAD}} & \multicolumn{6}{c}{\textbf{ResAD+Inv}} & \multicolumn{6}{c}{\textbf{ResAD+DART}} \\
\cmidrule(lr){2-7}\cmidrule(lr){8-13}\cmidrule(lr){14-19}
Category & I-AUC & AP & I-F1 & P-AUC & PRO & P-F1 & I-AUC & AP & I-F1 & P-AUC & PRO & P-F1 & I-AUC & AP & I-F1 & P-AUC & PRO & P-F1 \\
\midrule
candle & 86.25 & 84.83 & 80.24 & 98.26 & 93.44 & 32.97 & 85.09 & 84.29 & 80.67 & 98.20 & 93.76 & 27.06 & 89.19 & 88.35 & 84.00 & 98.96 & 95.14 & 29.39 \\
capsules & 70.74 & 78.74 & 78.75 & 95.32 & 76.09 & 29.60 & 58.38 & 73.05 & 76.92 & 95.33 & 60.69 & 28.89 & 58.62 & 68.84 & 76.98 & 94.13 & 71.42 & 18.78 \\
cashew & 91.08 & 95.16 & 87.53 & 96.22 & 89.72 & 41.13 & 93.18 & 96.86 & 90.45 & 95.85 & 89.31 & 38.56 & 93.12 & 96.68 & 91.79 & 97.53 & 90.83 & 45.43 \\
chewinggum & 97.48 & 97.88 & 95.22 & 98.49 & 82.33 & 64.47 & 95.56 & 98.10 & 93.94 & 98.38 & 75.58 & 55.96 & 98.30 & 99.27 & 96.41 & 98.96 & 82.17 & 54.46 \\
fryum & 77.49 & 88.50 & 81.64 & 90.67 & 70.89 & 35.00 & 70.34 & 83.46 & 81.15 & 90.31 & 67.66 & 31.89 & 76.78 & 88.38 & 82.01 & 90.68 & 70.72 & 31.02 \\
macaroni1 & 79.49 & 82.04 & 73.86 & 97.84 & 85.67 & 29.36 & 64.87 & 68.08 & 68.49 & 95.09 & 71.58 & 12.32 & 76.49 & 80.76 & 70.74 & 97.87 & 82.44 & 20.71 \\
macaroni2 & 70.18 & 67.47 & 72.13 & 95.84 & 79.89 & 6.87 & 65.72 & 61.91 & 71.21 & 93.35 & 70.40 & 2.42 & 77.69 & 78.64 & 75.81 & 96.71 & 84.24 & 9.34 \\
pcb1 & 79.60 & 74.29 & 78.42 & 96.48 & 74.55 & 52.58 & 66.23 & 59.42 & 73.00 & 93.72 & 72.53 & 19.10 & 84.78 & 81.47 & 81.98 & 95.26 & 70.92 & 24.51 \\
pcb2 & 61.91 & 58.42 & 68.60 & 92.31 & 63.49 & 16.36 & 60.59 & 59.81 & 68.18 & 92.56 & 64.18 & 8.90 & 66.62 & 68.60 & 68.46 & 92.49 & 63.54 & 15.40 \\
pcb3 & 70.21 & 71.74 & 70.05 & 94.00 & 70.91 & 36.39 & 69.16 & 69.39 & 70.42 & 94.12 & 63.52 & 28.99 & 70.70 & 70.32 & 71.21 & 94.05 & 67.89 & 26.55 \\
pcb4 & 93.64 & 91.99 & 90.87 & 89.60 & 65.45 & 17.23 & 73.23 & 69.04 & 72.41 & 84.62 & 52.46 & 7.40 & 91.92 & 91.92 & 84.65 & 90.09 & 65.66 & 16.64 \\
pipe\_fryum & 96.42 & 97.46 & 94.73 & 96.87 & 92.28 & 47.71 & 93.62 & 97.08 & 91.28 & 96.61 & 91.87 & 42.57 & 96.70 & 98.48 & 94.42 & 97.87 & 93.82 & 49.79 \\
\midrule
\textbf{mean} & 81.21 & 82.38 & 81.00 & 95.16 & 78.73 & 34.14 & 74.66 & 76.71 & 78.18 & 94.01 & 72.80 & 25.34 & 81.74 & 84.31 & 81.54 & 95.38 & 78.23 & 28.50 \\
\bottomrule
\end{tabular}%
}
\end{table*}

\begin{table*}[p]
\centering
\caption{Per-category VisA results for \textbf{ResAD} variants, $N^2+A^1$ (2-shot) setting.}
\label{tab:supp_resad_mtv_2shot}
\resizebox{\textwidth}{!}{%
\begin{tabular}{l cccccc cccccc cccccc}
\toprule
 & \multicolumn{6}{c}{\textbf{ResAD}} & \multicolumn{6}{c}{\textbf{ResAD+Inv}} & \multicolumn{6}{c}{\textbf{ResAD+DART}} \\
\cmidrule(lr){2-7}\cmidrule(lr){8-13}\cmidrule(lr){14-19}
Category & I-AUC & AP & I-F1 & P-AUC & PRO & P-F1 & I-AUC & AP & I-F1 & P-AUC & PRO & P-F1 & I-AUC & AP & I-F1 & P-AUC & PRO & P-F1 \\
\midrule
candle & 84.83 & 82.11 & 81.59 & 98.35 & 93.35 & 31.24 & 83.63 & 82.18 & 78.22 & 98.23 & 93.59 & 25.66 & 86.53 & 86.02 & 81.38 & 98.83 & 94.93 & 27.32 \\
capsules & 73.19 & 82.16 & 79.27 & 97.18 & 80.27 & 49.36 & 67.37 & 77.79 & 77.22 & 96.24 & 62.53 & 39.99 & 66.88 & 73.55 & 79.34 & 97.18 & 76.53 & 36.21 \\
cashew & 91.83 & 95.37 & 88.25 & 96.48 & 90.40 & 42.10 & 94.08 & 97.47 & 91.01 & 96.47 & 90.91 & 41.06 & 95.18 & 97.85 & 93.14 & 97.76 & 91.65 & 47.64 \\
chewinggum & 97.53 & 97.81 & 95.54 & 98.62 & 81.91 & 64.10 & 98.36 & 99.24 & 95.10 & 98.44 & 76.59 & 57.14 & 98.68 & 99.40 & 96.52 & 98.92 & 82.60 & 53.90 \\
fryum & 82.05 & 90.31 & 83.23 & 90.92 & 71.48 & 36.17 & 68.66 & 84.15 & 81.33 & 90.65 & 70.59 & 33.21 & 72.38 & 85.81 & 80.69 & 91.11 & 71.99 & 33.82 \\
macaroni1 & 83.13 & 84.73 & 77.05 & 98.22 & 88.10 & 30.23 & 68.94 & 72.08 & 69.50 & 95.72 & 76.17 & 14.23 & 82.75 & 85.52 & 75.13 & 98.49 & 88.40 & 23.05 \\
macaroni2 & 65.87 & 67.10 & 67.28 & 95.95 & 79.54 & 8.27 & 66.46 & 63.82 & 70.69 & 93.10 & 69.62 & 1.74 & 73.64 & 73.33 & 72.73 & 96.94 & 84.31 & 8.91 \\
pcb1 & 79.47 & 73.99 & 80.55 & 96.72 & 74.59 & 51.98 & 66.65 & 59.89 & 74.42 & 94.29 & 73.24 & 20.36 & 83.14 & 80.27 & 83.48 & 95.44 & 71.35 & 25.70 \\
pcb2 & 67.93 & 64.74 & 68.35 & 93.74 & 67.37 & 30.19 & 65.50 & 63.26 & 69.89 & 94.22 & 66.66 & 16.87 & 73.68 & 74.49 & 70.31 & 94.12 & 69.90 & 29.74 \\
pcb3 & 82.60 & 82.54 & 77.86 & 95.13 & 76.69 & 39.70 & 76.28 & 78.25 & 73.02 & 95.71 & 73.09 & 37.83 & 75.73 & 77.26 & 71.32 & 95.15 & 73.46 & 34.20 \\
pcb4 & 93.95 & 92.50 & 93.29 & 91.56 & 68.76 & 18.71 & 82.17 & 82.09 & 76.73 & 86.97 & 58.38 & 12.29 & 95.29 & 94.54 & 91.24 & 92.43 & 70.87 & 19.70 \\
pipe\_fryum & 96.59 & 97.31 & 94.04 & 97.19 & 92.41 & 46.76 & 93.52 & 96.91 & 90.15 & 96.88 & 92.52 & 42.55 & 97.44 & 98.77 & 94.12 & 97.81 & 93.98 & 49.13 \\
\midrule
\textbf{mean} & 83.25 & 84.22 & 82.19 & 95.84 & 80.41 & 37.40 & 77.63 & 79.76 & 78.94 & 94.74 & 75.32 & 28.58 & 83.44 & 85.57 & 82.45 & 96.18 & 80.83 & 32.44 \\
\bottomrule
\end{tabular}%
}
\end{table*}

\begin{table*}[p]
\centering
\caption{Per-category VisA results for \textbf{ResAD} variants, $N^4+A^1$ (4-shot) setting.}
\label{tab:supp_resad_mtv_4shot}
\resizebox{\textwidth}{!}{%
\begin{tabular}{l cccccc cccccc cccccc}
\toprule
 & \multicolumn{6}{c}{\textbf{ResAD}} & \multicolumn{6}{c}{\textbf{ResAD+Inv}} & \multicolumn{6}{c}{\textbf{ResAD+DART}} \\
\cmidrule(lr){2-7}\cmidrule(lr){8-13}\cmidrule(lr){14-19}
Category & I-AUC & AP & I-F1 & P-AUC & PRO & P-F1 & I-AUC & AP & I-F1 & P-AUC & PRO & P-F1 & I-AUC & AP & I-F1 & P-AUC & PRO & P-F1 \\
\midrule
candle & 86.52 & 83.89 & 82.31 & 98.55 & 94.18 & 31.39 & 84.32 & 81.67 & 80.00 & 98.54 & 94.25 & 27.13 & 88.63 & 86.94 & 81.82 & 98.92 & 95.25 & 27.99 \\
capsules & 79.18 & 86.78 & 79.96 & 97.56 & 82.94 & 52.76 & 68.82 & 80.26 & 77.65 & 96.88 & 67.59 & 44.73 & 73.55 & 81.87 & 81.55 & 97.45 & 81.53 & 45.19 \\
cashew & 92.92 & 96.33 & 89.69 & 96.73 & 91.22 & 42.94 & 94.66 & 97.56 & 90.32 & 96.65 & 91.37 & 41.93 & 95.48 & 97.99 & 92.86 & 97.88 & 91.54 & 48.84 \\
chewinggum & 98.22 & 98.59 & 97.37 & 98.73 & 82.11 & 63.76 & 98.52 & 99.33 & 96.91 & 98.66 & 76.40 & 59.10 & 99.30 & 99.70 & 98.49 & 98.94 & 81.02 & 54.25 \\
fryum & 87.55 & 93.85 & 84.88 & 91.56 & 73.35 & 38.63 & 78.02 & 88.70 & 83.33 & 91.40 & 72.04 & 37.50 & 83.02 & 91.90 & 83.65 & 91.81 & 73.40 & 38.30 \\
macaroni1 & 92.58 & 92.95 & 87.22 & 98.75 & 92.04 & 30.33 & 83.33 & 83.57 & 78.01 & 97.58 & 85.97 & 19.25 & 91.55 & 92.52 & 84.06 & 99.09 & 92.54 & 23.59 \\
macaroni2 & 73.31 & 75.83 & 69.79 & 97.10 & 84.03 & 16.40 & 68.64 & 69.45 & 70.27 & 95.60 & 78.16 & 4.31 & 82.38 & 83.55 & 77.88 & 97.63 & 87.24 & 13.25 \\
pcb1 & 82.25 & 78.78 & 82.17 & 97.00 & 76.84 & 55.22 & 72.13 & 64.76 & 75.59 & 95.27 & 77.10 & 23.48 & 82.96 & 79.12 & 83.70 & 95.99 & 74.32 & 28.46 \\
pcb2 & 72.43 & 68.77 & 71.74 & 94.68 & 71.21 & 32.42 & 67.29 & 65.85 & 70.00 & 95.13 & 70.21 & 20.31 & 80.31 & 80.08 & 78.26 & 95.09 & 73.78 & 33.12 \\
pcb3 & 86.08 & 86.40 & 79.88 & 96.06 & 81.66 & 41.49 & 82.04 & 81.87 & 77.45 & 96.62 & 76.96 & 38.25 & 82.56 & 83.14 & 75.89 & 96.28 & 79.57 & 38.16 \\
pcb4 & 95.62 & 94.64 & 94.11 & 92.42 & 71.38 & 19.83 & 89.42 & 89.61 & 82.25 & 89.42 & 65.04 & 17.07 & 96.67 & 95.63 & 93.72 & 93.36 & 73.09 & 20.65 \\
pipe\_fryum & 97.82 & 98.31 & 96.43 & 97.74 & 93.49 & 49.08 & 96.90 & 98.38 & 94.12 & 97.59 & 93.42 & 46.93 & 98.12 & 98.97 & 96.12 & 98.23 & 94.14 & 51.10 \\
\midrule
\textbf{mean} & 87.04 & 87.93 & 84.63 & 96.41 & 82.87 & 39.52 & 82.01 & 83.42 & 81.33 & 95.78 & 79.04 & 31.67 & 87.88 & 89.28 & 85.67 & 96.72 & 83.12 & 35.24 \\
\bottomrule
\end{tabular}%
}
\end{table*}

\label{tab:supp:nagl}


\begin{table*}[p]
\centering
\caption{Per-category MVTecAD results for \textbf{NAGL} variants, $N^1+A^1$ (1-shot) setting.}
\label{tab:supp_nagl_mvtec_1shot}
\resizebox{\textwidth}{!}{%
\begin{tabular}{l cccccc cccccc}
\toprule
 & \multicolumn{6}{c}{\textbf{NAGL}} & \multicolumn{6}{c}{\textbf{NAGL+DART}} \\
\cmidrule(lr){2-7}\cmidrule(lr){8-13}
Category & I-AUC & AP & I-F1 & P-AUC & PRO & P-F1 & I-AUC & AP & I-F1 & P-AUC & PRO & P-F1 \\
\midrule
bottle & 98.78 & 99.52 & 98.70 & 98.31 & 95.61 & 76.32 & 98.89 & 99.57 & 98.95 & 98.20 & 95.58 & 76.16 \\
cable & 90.09 & 94.35 & 87.96 & 95.48 & 88.10 & 58.16 & 92.62 & 95.89 & 90.59 & 95.32 & 88.16 & 57.92 \\
capsule & 90.68 & 97.91 & 94.17 & 97.95 & 96.36 & 45.57 & 91.18 & 98.02 & 94.77 & 98.00 & 96.48 & 46.31 \\
carpet & 100.00 & 100.00 & 100.00 & 99.29 & 98.40 & 68.15 & 100.00 & 100.00 & 100.00 & 99.36 & 98.52 & 68.89 \\
grid & 99.81 & 99.93 & 99.41 & 99.23 & 96.68 & 48.29 & 99.75 & 99.91 & 99.12 & 99.29 & 96.92 & 49.61 \\
hazelnut & 99.45 & 99.70 & 98.07 & 99.47 & 97.27 & 78.01 & 99.33 & 99.64 & 97.89 & 99.48 & 97.00 & 78.30 \\
leather & 100.00 & 100.00 & 100.00 & 99.25 & 98.64 & 45.63 & 100.00 & 100.00 & 100.00 & 99.28 & 98.73 & 45.93 \\
metal\_nut & 99.30 & 99.84 & 98.04 & 93.70 & 91.51 & 62.39 & 99.19 & 99.82 & 97.85 & 93.74 & 91.81 & 62.66 \\
pill & 93.42 & 98.67 & 95.61 & 92.77 & 95.99 & 48.78 & 92.86 & 98.57 & 95.55 & 92.41 & 96.12 & 46.95 \\
screw & 77.06 & 91.26 & 87.40 & 97.57 & 91.36 & 40.18 & 78.06 & 91.89 & 87.25 & 97.71 & 91.76 & 41.62 \\
tile & 99.51 & 99.82 & 98.03 & 96.41 & 91.07 & 69.28 & 99.43 & 99.79 & 98.18 & 96.54 & 91.86 & 70.63 \\
toothbrush & 98.61 & 99.51 & 97.21 & 99.05 & 94.93 & 63.72 & 98.61 & 99.51 & 97.21 & 99.10 & 94.91 & 65.11 \\
transistor & 88.15 & 84.41 & 80.13 & 84.62 & 63.44 & 43.82 & 90.10 & 86.94 & 81.88 & 83.76 & 63.39 & 41.00 \\
wood & 99.71 & 99.91 & 98.89 & 97.04 & 96.73 & 67.95 & 99.85 & 99.95 & 99.17 & 97.22 & 97.00 & 68.15 \\
zipper & 98.52 & 99.62 & 97.36 & 97.83 & 93.82 & 54.15 & 99.00 & 99.74 & 97.90 & 98.26 & 94.95 & 56.27 \\
\midrule
\textbf{mean} & 95.54 & 97.63 & 95.40 & 96.53 & 92.66 & 58.03 & 95.93 & 97.95 & 95.76 & 96.51 & 92.88 & 58.37 \\
\bottomrule
\end{tabular}%
}
\end{table*}

\begin{table*}[p]
\centering
\caption{Per-category MVTecAD results for \textbf{NAGL} variants, $N^2+A^1$ (2-shot) setting.}
\label{tab:supp_nagl_mvtec_2shot}
\resizebox{\textwidth}{!}{%
\begin{tabular}{l cccccc cccccc}
\toprule
 & \multicolumn{6}{c}{\textbf{NAGL}} & \multicolumn{6}{c}{\textbf{NAGL+DART}} \\
\cmidrule(lr){2-7}\cmidrule(lr){8-13}
Category & I-AUC & AP & I-F1 & P-AUC & PRO & P-F1 & I-AUC & AP & I-F1 & P-AUC & PRO & P-F1 \\
\midrule
bottle & 98.99 & 99.57 & 99.48 & 98.22 & 95.81 & 76.47 & 98.99 & 99.58 & 99.48 & 98.17 & 95.72 & 76.36 \\
cable & 92.73 & 95.86 & 89.74 & 95.72 & 88.70 & 58.87 & 92.69 & 95.91 & 89.78 & 95.54 & 88.40 & 58.76 \\
capsule & 90.79 & 97.89 & 94.81 & 98.01 & 96.60 & 46.37 & 90.63 & 97.86 & 94.67 & 98.06 & 96.66 & 47.01 \\
carpet & 100.00 & 100.00 & 100.00 & 99.33 & 98.46 & 68.59 & 100.00 & 100.00 & 100.00 & 99.35 & 98.48 & 68.75 \\
grid & 99.81 & 99.93 & 99.42 & 99.31 & 97.00 & 49.44 & 99.83 & 99.94 & 99.42 & 99.32 & 97.02 & 49.57 \\
hazelnut & 99.96 & 99.98 & 99.52 & 99.61 & 97.86 & 80.50 & 99.98 & 99.99 & 99.53 & 99.63 & 97.84 & 80.70 \\
leather & 100.00 & 100.00 & 100.00 & 99.29 & 98.66 & 46.56 & 100.00 & 100.00 & 100.00 & 99.28 & 98.65 & 46.10 \\
metal\_nut & 99.49 & 99.89 & 98.56 & 94.35 & 92.59 & 65.31 & 99.41 & 99.87 & 98.56 & 94.31 & 92.46 & 65.09 \\
pill & 95.17 & 99.02 & 96.73 & 92.66 & 96.17 & 49.13 & 95.24 & 99.05 & 96.83 & 92.73 & 96.31 & 48.99 \\
screw & 85.81 & 95.14 & 87.94 & 98.11 & 92.83 & 46.41 & 85.48 & 94.93 & 88.38 & 98.13 & 92.90 & 46.47 \\
tile & 99.80 & 99.92 & 98.62 & 96.61 & 91.87 & 70.36 & 99.81 & 99.93 & 98.81 & 96.49 & 91.79 & 70.29 \\
toothbrush & 99.26 & 99.72 & 97.81 & 99.11 & 95.38 & 64.71 & 99.26 & 99.72 & 97.81 & 99.10 & 95.26 & 64.69 \\
transistor & 88.08 & 84.97 & 80.95 & 83.47 & 62.94 & 39.94 & 88.81 & 85.73 & 82.05 & 83.87 & 63.33 & 40.17 \\
wood & 99.91 & 99.97 & 99.44 & 97.38 & 96.96 & 68.80 & 99.85 & 99.95 & 99.17 & 97.30 & 96.98 & 68.66 \\
zipper & 99.35 & 99.84 & 98.45 & 98.30 & 95.05 & 56.61 & 99.29 & 99.82 & 98.45 & 98.34 & 95.13 & 56.64 \\
\midrule
\textbf{mean} & 96.61 & 98.11 & 96.10 & 96.63 & 93.13 & 59.20 & 96.62 & 98.15 & 96.20 & 96.64 & 93.13 & 59.22 \\
\bottomrule
\end{tabular}%
}
\end{table*}

\begin{table*}[p]
\centering
\caption{Per-category MVTecAD results for \textbf{NAGL} variants, $N^4+A^1$ (4-shot) setting.}
\label{tab:supp_nagl_mvtec_4shot}
\resizebox{\textwidth}{!}{%
\begin{tabular}{l cccccc cccccc}
\toprule
 & \multicolumn{6}{c}{\textbf{NAGL}} & \multicolumn{6}{c}{\textbf{NAGL+DART}} \\
\cmidrule(lr){2-7}\cmidrule(lr){8-13}
Category & I-AUC & AP & I-F1 & P-AUC & PRO & P-F1 & I-AUC & AP & I-F1 & P-AUC & PRO & P-F1 \\
\midrule
bottle & 99.74 & 99.91 & 99.21 & 98.27 & 95.77 & 76.65 & 99.79 & 99.93 & 99.21 & 98.17 & 95.61 & 76.37 \\
cable & 93.93 & 96.63 & 90.58 & 95.86 & 89.22 & 60.03 & 95.01 & 97.26 & 91.86 & 95.82 & 89.35 & 60.35 \\
capsule & 96.04 & 99.19 & 95.87 & 98.26 & 97.05 & 51.79 & 96.48 & 99.29 & 96.04 & 98.28 & 97.08 & 52.18 \\
carpet & 100.00 & 100.00 & 100.00 & 99.33 & 98.43 & 68.25 & 100.00 & 100.00 & 100.00 & 99.35 & 98.49 & 68.63 \\
grid & 99.89 & 99.96 & 99.71 & 99.33 & 97.06 & 49.99 & 99.83 & 99.94 & 99.42 & 99.33 & 97.07 & 50.14 \\
hazelnut & 99.99 & 99.99 & 99.76 & 99.63 & 97.79 & 80.78 & 99.95 & 99.97 & 99.52 & 99.63 & 97.62 & 80.93 \\
leather & 100.00 & 100.00 & 100.00 & 99.28 & 98.54 & 46.83 & 100.00 & 100.00 & 100.00 & 99.30 & 98.56 & 47.13 \\
metal\_nut & 99.53 & 99.89 & 98.73 & 95.12 & 93.48 & 68.69 & 99.56 & 99.90 & 98.91 & 95.20 & 93.47 & 69.12 \\
pill & 95.27 & 99.08 & 96.48 & 92.56 & 96.20 & 49.57 & 95.94 & 99.20 & 96.84 & 92.74 & 96.36 & 49.81 \\
screw & 83.34 & 94.14 & 87.66 & 98.20 & 93.22 & 45.53 & 84.08 & 94.45 & 87.50 & 98.23 & 93.34 & 45.85 \\
tile & 99.80 & 99.92 & 98.81 & 96.54 & 91.60 & 69.96 & 99.87 & 99.95 & 99.20 & 96.51 & 91.78 & 70.29 \\
toothbrush & 99.72 & 99.89 & 98.92 & 99.09 & 95.37 & 64.12 & 99.81 & 99.93 & 99.45 & 99.11 & 95.32 & 64.50 \\
transistor & 89.76 & 86.67 & 82.55 & 83.07 & 63.58 & 38.76 & 90.64 & 87.30 & 82.68 & 84.01 & 64.57 & 39.21 \\
wood & 99.77 & 99.93 & 98.88 & 97.30 & 96.89 & 68.66 & 99.80 & 99.94 & 98.89 & 97.32 & 96.97 & 68.77 \\
zipper & 99.40 & 99.85 & 98.32 & 98.35 & 95.18 & 56.94 & 99.33 & 99.83 & 98.33 & 98.40 & 95.30 & 57.21 \\
\midrule
\textbf{mean} & 97.08 & 98.34 & 96.37 & 96.68 & 93.29 & 59.77 & 97.34 & 98.46 & 96.52 & 96.76 & 93.39 & 60.03 \\
\bottomrule
\end{tabular}%
}
\end{table*}

\begin{table*}[p]
\centering
\caption{Per-category VisA results for \textbf{NAGL} variants, $N^1+A^1$ (1-shot) setting.}
\label{tab:supp_nagl_visa_1shot}
\resizebox{\textwidth}{!}{%
\begin{tabular}{l cccccc cccccc}
\toprule
 & \multicolumn{6}{c}{\textbf{NAGL}} & \multicolumn{6}{c}{\textbf{NAGL+DART}} \\
\cmidrule(lr){2-7}\cmidrule(lr){8-13}
Category & I-AUC & AP & I-F1 & P-AUC & PRO & P-F1 & I-AUC & AP & I-F1 & P-AUC & PRO & P-F1 \\
\midrule
candle & 92.99 & 92.97 & 86.12 & 98.68 & 93.40 & 40.61 & 91.41 & 91.76 & 84.16 & 98.83 & 93.41 & 39.48 \\
capsules & 97.80 & 98.54 & 95.19 & 97.73 & 81.69 & 48.64 & 95.67 & 97.30 & 93.20 & 97.16 & 80.92 & 36.30 \\
cashew & 90.84 & 94.84 & 89.81 & 98.77 & 94.69 & 54.18 & 86.08 & 93.42 & 85.19 & 98.88 & 95.25 & 57.43 \\
chewinggum & 97.48 & 98.94 & 95.29 & 99.26 & 94.98 & 69.02 & 96.68 & 98.69 & 95.92 & 99.03 & 93.88 & 68.96 \\
fryum & 95.96 & 98.21 & 93.53 & 96.48 & 83.27 & 45.74 & 94.12 & 97.23 & 90.64 & 95.40 & 82.05 & 39.33 \\
macaroni1 & 85.52 & 84.67 & 80.56 & 98.75 & 92.80 & 32.60 & 90.67 & 90.44 & 83.11 & 98.42 & 91.37 & 30.83 \\
macaroni2 & 55.48 & 52.26 & 66.91 & 97.48 & 86.93 & 9.73 & 66.35 & 62.42 & 69.89 & 98.06 & 89.54 & 18.36 \\
pcb1 & 94.87 & 92.26 & 90.64 & 98.94 & 92.87 & 60.63 & 90.79 & 87.31 & 87.32 & 98.89 & 93.16 & 58.60 \\
pcb2 & 86.00 & 83.67 & 81.55 & 95.96 & 89.12 & 36.96 & 82.81 & 80.24 & 79.13 & 96.20 & 89.32 & 33.03 \\
pcb3 & 81.56 & 79.04 & 77.39 & 93.61 & 86.82 & 42.14 & 81.90 & 81.50 & 74.18 & 93.81 & 86.31 & 40.06 \\
pcb4 & 60.70 & 65.22 & 66.67 & 92.05 & 71.68 & 20.38 & 80.18 & 78.06 & 76.02 & 92.97 & 73.06 & 21.87 \\
pipe\_fryum & 95.52 & 97.87 & 91.46 & 98.71 & 96.85 & 48.29 & 95.12 & 97.63 & 92.23 & 98.89 & 96.97 & 50.78 \\
\midrule
\textbf{mean} & 86.23 & 86.54 & 84.59 & 97.20 & 88.76 & 42.41 & 87.65 & 88.00 & 84.25 & 97.21 & 88.77 & 41.25 \\
\bottomrule
\end{tabular}%
}
\end{table*}

\begin{table*}[p]
\centering
\caption{Per-category VisA results for \textbf{NAGL} variants, $N^2+A^1$ (2-shot) setting.}
\label{tab:supp_nagl_visa_2shot}
\resizebox{\textwidth}{!}{%
\begin{tabular}{l cccccc cccccc}
\toprule
 & \multicolumn{6}{c}{\textbf{NAGL}} & \multicolumn{6}{c}{\textbf{NAGL+DART}} \\
\cmidrule(lr){2-7}\cmidrule(lr){8-13}
Category & I-AUC & AP & I-F1 & P-AUC & PRO & P-F1 & I-AUC & AP & I-F1 & P-AUC & PRO & P-F1 \\
\midrule
candle & 93.94 & 94.15 & 87.32 & 98.88 & 93.92 & 42.41 & 92.37 & 93.15 & 85.02 & 98.70 & 92.42 & 37.88 \\
capsules & 98.20 & 98.84 & 95.19 & 97.38 & 80.61 & 48.09 & 95.67 & 97.29 & 92.52 & 97.27 & 80.81 & 48.33 \\
cashew & 86.84 & 93.72 & 86.12 & 98.58 & 93.56 & 56.20 & 93.78 & 97.10 & 92.08 & 99.06 & 96.76 & 59.06 \\
chewinggum & 96.94 & 98.76 & 95.29 & 99.16 & 94.52 & 69.94 & 98.20 & 99.20 & 95.43 & 99.16 & 94.80 & 68.58 \\
fryum & 96.50 & 98.40 & 94.36 & 96.10 & 82.38 & 44.64 & 96.88 & 98.62 & 95.05 & 95.88 & 82.63 & 43.97 \\
macaroni1 & 90.14 & 90.10 & 84.02 & 98.95 & 93.50 & 36.34 & 88.16 & 85.97 & 82.69 & 98.32 & 91.02 & 22.58 \\
macaroni2 & 63.53 & 59.83 & 68.84 & 98.13 & 89.73 & 15.26 & 67.73 & 59.96 & 71.38 & 98.33 & 90.99 & 15.71 \\
pcb1 & 95.11 & 92.45 & 91.79 & 98.87 & 93.02 & 57.91 & 91.60 & 87.44 & 89.10 & 98.89 & 93.18 & 59.43 \\
pcb2 & 86.54 & 84.30 & 81.17 & 96.14 & 89.79 & 37.29 & 83.87 & 79.68 & 78.79 & 96.32 & 89.34 & 35.78 \\
pcb3 & 84.84 & 87.19 & 77.19 & 94.09 & 88.14 & 46.40 & 88.00 & 89.57 & 80.72 & 93.89 & 87.64 & 42.85 \\
pcb4 & 65.50 & 72.19 & 66.89 & 92.92 & 74.73 & 23.53 & 83.25 & 83.41 & 78.17 & 92.64 & 72.25 & 24.66 \\
pipe\_fryum & 96.70 & 98.42 & 93.60 & 98.86 & 97.05 & 50.75 & 97.14 & 98.57 & 93.53 & 98.99 & 96.97 & 52.94 \\
\midrule
\textbf{mean} & 87.90 & 89.03 & 85.15 & 97.34 & 89.25 & 44.06 & 89.72 & 89.16 & 86.21 & 97.29 & 89.07 & 42.65 \\
\bottomrule
\end{tabular}%
}
\end{table*}

\begin{table*}[p]
\centering
\caption{Per-category VisA results for \textbf{NAGL} variants, $N^4+A^1$ (4-shot) setting.}
\label{tab:supp_nagl_visa_4shot}
\resizebox{\textwidth}{!}{%
\begin{tabular}{l cccccc cccccc}
\toprule
 & \multicolumn{6}{c}{\textbf{NAGL}} & \multicolumn{6}{c}{\textbf{NAGL+DART}} \\
\cmidrule(lr){2-7}\cmidrule(lr){8-13}
Category & I-AUC & AP & I-F1 & P-AUC & PRO & P-F1 & I-AUC & AP & I-F1 & P-AUC & PRO & P-F1 \\
\midrule
candle & 92.75 & 93.07 & 87.08 & 98.84 & 93.62 & 42.21 & 92.50 & 93.27 & 85.31 & 98.81 & 93.10 & 39.42 \\
capsules & 98.22 & 98.94 & 94.47 & 97.39 & 80.21 & 51.93 & 95.28 & 97.00 & 92.52 & 98.02 & 83.05 & 49.48 \\
cashew & 92.26 & 96.57 & 87.91 & 98.63 & 93.76 & 58.16 & 91.92 & 96.27 & 89.45 & 98.89 & 96.90 & 57.77 \\
chewinggum & 97.28 & 98.89 & 95.88 & 99.17 & 94.67 & 69.84 & 97.34 & 98.90 & 94.95 & 99.12 & 94.51 & 67.82 \\
fryum & 96.90 & 98.59 & 94.85 & 96.11 & 82.62 & 45.26 & 97.32 & 98.79 & 95.43 & 95.94 & 83.50 & 44.18 \\
macaroni1 & 90.16 & 89.04 & 85.45 & 98.99 & 93.57 & 32.90 & 88.68 & 86.08 & 85.20 & 98.70 & 92.84 & 26.41 \\
macaroni2 & 65.58 & 60.63 & 70.92 & 98.27 & 90.42 & 19.02 & 71.68 & 65.44 & 72.95 & 98.58 & 92.01 & 21.10 \\
pcb1 & 95.94 & 93.22 & 93.88 & 98.80 & 92.79 & 56.71 & 93.09 & 89.06 & 91.35 & 98.85 & 93.26 & 58.39 \\
pcb2 & 87.54 & 85.36 & 82.67 & 96.19 & 89.76 & 37.74 & 87.23 & 83.93 & 81.98 & 96.55 & 89.92 & 37.59 \\
pcb3 & 87.84 & 90.24 & 81.28 & 94.29 & 88.99 & 46.60 & 90.68 & 91.62 & 83.96 & 93.73 & 88.69 & 41.86 \\
pcb4 & 83.43 & 86.01 & 77.21 & 93.71 & 77.99 & 26.06 & 92.30 & 92.06 & 86.96 & 93.28 & 74.96 & 27.46 \\
pipe\_fryum & 95.72 & 97.87 & 92.75 & 98.91 & 97.28 & 51.85 & 97.50 & 98.86 & 94.36 & 98.89 & 97.08 & 51.26 \\
\midrule
\textbf{mean} & 90.30 & 90.70 & 87.03 & 97.44 & 89.64 & 44.86 & 91.29 & 90.94 & 87.87 & 97.45 & 89.99 & 43.56 \\
\bottomrule
\end{tabular}%
}
\end{table*}

\end{document}